\pdfoutput=1

\documentclass{article}
\newif\ifarxiv
\arxivtrue
\usepackage{iclr2027_conference,times}
\ifarxiv\iclrfinalcopy\fi
\usepackage{booktabs}
\usepackage{wrapfig}
\newenvironment{ruledtabular}{\begin{center}}{\end{center}}
\providecommand{\colrule}{\midrule}

\usepackage{appendix}
\usepackage{hyperref}
\usepackage{dynlearn}
\ifarxiv\fi
\fxsetup{status=draft}
\usepackage{empheq}
\usepackage{overpic}
\usepackage{amsthm}
\usepackage[dvipsnames]{xcolor}
\usepackage{enumitem}
\usepackage{tcolorbox}

\newcommand\myshade{85}
\colorlet{mylinkcolor}{BrickRed}
\colorlet{mycitecolor}{NavyBlue}
\colorlet{myurlcolor}{Aquamarine}

\hypersetup{
  linkcolor  = mylinkcolor!\myshade!black,
  urlcolor   = myurlcolor!\myshade!black,
  colorlinks = true,
}

\newtheorem{theorem}{Theorem}
\newtheorem{corollary}{Corollary}

\makeatletter
\renewcommand\paragraph[1]{%
  \textit{#1.}\hspace{1em}%
}
\makeatother

\begin{document}
\def\ourTitle{Beliefs in feedback: Factored representations of interacting agents in transformers}

\def\ourAbstract{%
ABSTRACT
}

\def\ourKeywords{%
}

\def\ourAuthors{David A. Ehrlich, Manuel Baltieri, Fernando E. Rosas, Alexander Boyd}
\newcommand{\coderepo}{\url{https://github.com/xorlabs-research/feedback-transducers}}

\hypersetup{
  pdftitle={\ourTitle},
  pdfauthor={\ifarxiv\ourAuthors\fi},
  pdfsubject={\ourAbstract},
  pdfkeywords={\ourKeywords},
  pdfproducer={},
  pdfcreator={}
}

\title{\ourTitle}

\ifarxiv
\author{David A.\ Ehrlich\textsuperscript{1}\thanks{\footnotesize{Correspondence: \texttt{david@daehrlich.com}}} \And
        Manuel Baltieri\textsuperscript{2,3} \And
        Fernando E. Rosas\textsuperscript{1,3,4,5} \And
        Alexander B. Boyd\textsuperscript{1,6} \AND
        \mdseries\textsuperscript{1}XOR Labs, London, UK \\
        \mdseries\textsuperscript{2}Araya Inc., Tokyo, Japan \\
        \mdseries\textsuperscript{3}Department of Informatics, University of Sussex, Brighton, UK\\
        \mdseries\textsuperscript{4}Department of Brain Sciences, Imperial College London, London, UK \\
        \mdseries\textsuperscript{5}Centre for Eudaimonia and Human Flourishing, University of Oxford, Oxford, UK \\
        \mdseries\textsuperscript{6}Beyond Institute for Theoretical Science, San Francisco, CA, USA}
\fi

\date{\today}

\date{\today}
\maketitle
\ifarxiv\lhead{}\renewcommand{\headrulewidth}{0pt}\fi

\begin{abstract}
Complex systems, from ecosystems to brains, are usually modeled as modular components coupled through feedback rather than as indivisible wholes. Can a transformer trained on the joint behavior of a group of interacting agents be understood in the same way? In this work, we develop a computational-mechanics framework to study agents interacting in feedback loops. 
We formally prove that any joint stochastic process can be presented as resulting from the behavior of agents interacting in feedback, and the resulting process is generated by a hidden Markov model whose Bayesian beliefs factorize into separate components per agent.  Furthermore, our experiments show that transformers trained on such feedback sequences make use of this separability, disentangling the beliefs about the underlying factors into separate subspaces of their residual stream. This representation is found to scale linearly (instead of exponentially) with the number of agents. Moreover, interventions show that these representations are causally efficacious, allowing for targeted steering of individual agents' beliefs by selectively inserting counterfactuals. 
However, coarse-graining the behavior is found to disrupt the factorization, and can force a higher-dimensional representation. Together, the results establish both the dimensional advantage of a factorized representation and the ability of transformers to exploit it, laying the groundwork for investigating traces of compositional predictive architecture in pretrained models.
\end{abstract}

\section{Introduction}

Communities can accomplish feats beyond the capacity of their individuals---groups of ants solve complex geometry problems \citep{dreyer2025comparing}, birds flock in murmurations
responding to predators as a single perceptive body \citep{cavagna2010scale,attanasi2014information}, 
AI agents coordinate to effectively infiltrate digital infrastructure \citep{greenblatt2026investigation}, and 
people collaborate in the global project of developing a scientific understanding of the universe~\citep{fortunato2018science,zollman2007communication, hull2019science}. 
How individuals interact to produce such collective behavior is the subject of multi-agent systems research, which seeks to formalize these interactions and how agents form beliefs that can motivate predictions about other agents~\citep{gmytrasiewicz2005framework, albrecht2018autonomous, chen2020tensor, zhang2021reinforcement}. 

Interestingly, the problem of understanding how individuals come together into collectives is dual to the question of how a single complex system can be understood as a composite of sub-components. 
A longstanding tradition in cognitive science interprets single cognitive systems as arising from the interaction of distinct modular `agents'~\citep{minsky1986society}. 
Similarly, a substantial body of work in mechanistic interpretability research aims to decompose large deep learning systems into interconnected `circuits'~\citep{olah2022mech,olsson2022context,wang2022interpretability}.

The above considerations raise a question for any observer of a complex system: 
under what conditions can the observer track the system not as an indivisible whole, but via tracking its sub-components separately and then putting these together? 

As a first step towards addressing this question, here we investigate transformers trained to predict data generated by interacting agents. 
We describe the behavior of these agents using \emph{transducers}: stochastic automata that generate outputs from inputs and update their memory via probabilistic rules~\citep{Barn13a}.
Recent work has used principles of computational mechanics~\citep{Crut12a} to show that transformers tend to represent the geometry of Bayesian beliefs within their internal activations during inference~\citep{shai2024transformers,piotrowski2025constrained}, and that representations for independent processes factorize~\citep{shai2026transformers}. 
Building on these works, we investigate whether a transformer trained on data generated by interacting agents builds internal representations that mirror the agents that generated the data.

In this paper we introduce a theoretical framework to formally investigate the prediction of interacting agents, which is then empirically validated on trained transformers. 
The theoretical contributions are:
\begin{enumerate}[leftmargin=3em]
\item We develop a general computational-mechanics framework to describe the behavior of agents arranged in feedback loops. 
\item We formally show that any joint process can be viewed as arising from multiple subcomponents in feedback.
\item We demonstrate that these factored processes can lead to exponential advantages in the embedding dimension. However, these savings can be broken by coarse-graining.
\end{enumerate}

We use these theoretical advances to obtain the following experimental results on transformers trained to predict the activity of interacting agents:

\begin{enumerate}[leftmargin=3em]
    \item[4.] Transformers linearly represent factored Bayesian beliefs in separate subspaces of their residual stream, with each factor tracking an individual agent. 
    \item[5.] These representations are causally efficacious, being separately steerable by inserting counterfactual beliefs on specific subspaces. 
    \item[6.] By using factored beliefs, transformers can attain exponential savings in the required residual stream dimensionality for effectively predicting interacting agents. 
\end{enumerate}

\section{Agents in Feedback Loops}

\subsection{Background}
In this work we instantiate the behavior of agents using \emph{transducers}~\citep{Barn13a}: stochastic automata that generate outputs $Y_t\in\mathcal Y$ from receiving inputs $X_t\in\mathcal X$ using a memory state $R_t\in\mathcal R$.\footnote{\footnotesize{Uppercase letters (e.g. $X,Y$) denote random variables and lowercase (e.g. $x,y$) their realizations, calligraphic letters (e.g. $\mathcal{X},\mathcal{Y}$) denote the sets over which they take values, and $\Delta$ (as in $\Delta(\mathcal{X}),\Delta(\mathcal{Y})$) denotes the collection of all probability distributions over those sets. 
We also use the short-hand notation $x_{a:b}\equiv(x_a,\dots,x_{b-1})$, and also $p(x)\equiv\Pr(X=x)$ for denoting probability distributions.
}}
A transducer $\mathbf T=\langle \mathcal X,\mathcal Y, \mathcal R, T^{(y|x)}_{r \rightarrow r'}, p(r_0)\rangle$ is specified via the Markov kernel
\begin{align}
    T^{(y|x)}_{r \rightarrow r'} \equiv \Pr(Y_t=y,R_{t+1}=r'|X_t=x,R_t=r),
\end{align}
together with an initial state distribution $p(r_0)$.
Note that a transducer with a single input (i.e. $|\mathcal X|=1$) is equivalent to a hidden Markov model (HMM).

The probability of a transducer generating an input-output sequence $(x_{0:L},y_{0:L})$ can be calculated as
\begin{align}
\label{eq:interface}
    p(y_{0:L}|x_{0:L})= \sum_{r_{0:L+1}} p(r_0)\prod_{t=0}^{L-1}T^{(y_t|x_t)}_{r_t \rightarrow r_{t+1}}.
\end{align}
Thus, the effective behavior of any agent, given by its input-output distribution as in Eq.~\eqref{eq:interface}, can be represented via an appropriate transducer. Using computational-mechanics terms, one says the transducer provides a \emph{presentation} of the behavior of that agent~\citep{rosas2025ai}. 

A transducer can be described using linear operators of the form
\begin{align}\label{eq:operator}
    \hat{T}^{(y|x)} \equiv \sum_{r,r'} \bm{e}_{r'}  T^{(y|x)}_{r \rightarrow r'} \bm{e}_r^\intercal,
\end{align}
where $\bm{e}_r$ is an $|\mathcal{R}|$-dimensional vector with a $1$ in the $r$-th place and zeros everywhere else.  This operator can be used to build the transducer's \emph{Mixed State Presentation} (MSP) \citep{jurgens2021shannon, jurgens2021divergent}: a different transducer presentation of the input-output behavior whose states correspond to Bayesian beliefs~\citep{rosas2025ai}. These \emph{belief states} are probability distributions $\bm{\eta}_t\in\Delta(\mathcal{R})$ constructed via Bayesian filtering~\citep{sarkka2023bayesian} as
\begin{align}
    \bm{\eta}_t  \propto \prod_{i=0}^{t-1}\hat{T}^{(y_i|x_i)}\bm{\eta}_0 ,
\end{align}
with $\bm{\eta}_0\in \Delta (\mathcal R)$ being a prior distribution. 
Thus, the belief state $\bm{\eta}_t$ is the posterior distribution of $R_t$ given the history $(x_{0:t},y_{0:t})$, and can be shown to be a sufficient statistic of the past regarding future events. 
The geometry of belief states implies that proximity in the simplex guarantees similar futures.
Belief states have been observed in the residual stream of transformers trained to predict a process \citep{shai2024transformers,piotrowski2025constrained}. In what follows, we show that the belief state of a feedback loop between agents contains the belief states of each agent contained within the loop.

For an extended introduction to computational mechanics, see \cref{apx:Comp_Mech}. 

\subsection{Computational Mechanics of Transducers in Feedback Loops}

\begin{figure}
\centering
\includegraphics[width=\linewidth]{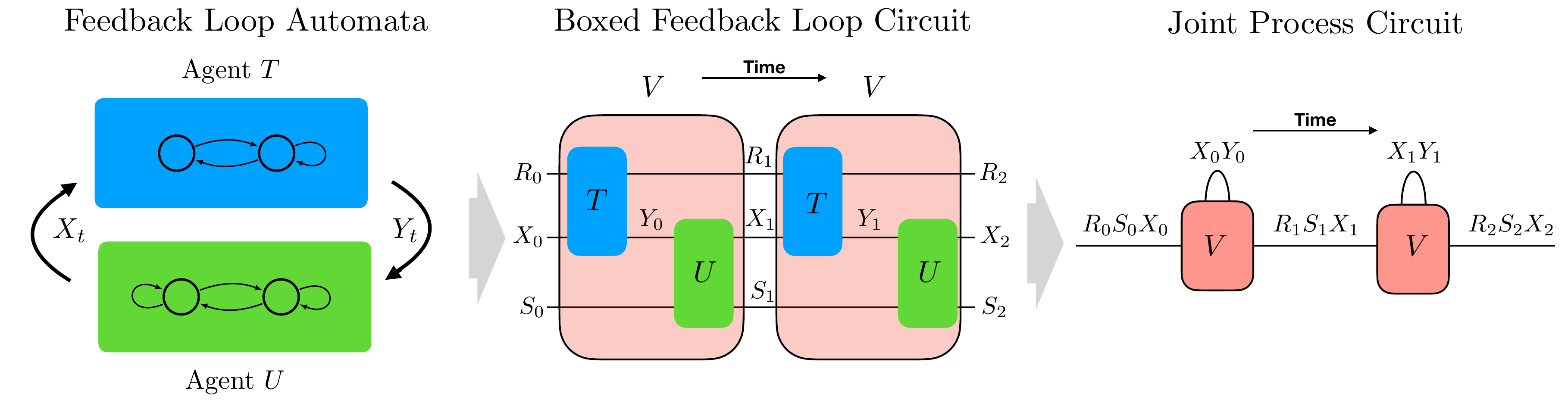}
\caption{(\textbf{Left}) Automata in a feedback loop representing two interacting agents. (\textbf{Center}) The feedback loop circuit requires a time-ordering for the signals going between agents, which can be boxed into a single repeating circuit element.  (\textbf{Right}) The boxing allows expressing the feedback loop as an HMM for the joint process.}
\label{fig:Feedback_Stages} 
\end{figure}

The behavior of interacting agents can be described via perception-action loops involving either an environment or other agents~\citep{tishby2011information, fiderer2025work, boyd2025monoliths, baltieri2026world}. In a loop involving agents $T$ and $U$, the actions $Y_t$ of agent $T$ become agent $U$'s perceptions, and the actions $X_t$ of agent $U$ become agent $T$'s perceptions (see \cref{fig:Feedback_Stages}, left). 
When this sequence of events is unrolled over time, it leaves ambiguity in the time index of the cycle. In order to construct a time series capturing the interaction, the symmetry needs to be broken---for instance, and without loss of generality, by setting $T$ to operate first (see \cref{fig:Feedback_Stages}, center). 

This results in a joint process over the perceptions/actions $(X_t,Y_t)$ together with the latent states $R_t$ of agent $T$ and $S_t$ for agent $U$ (see \cref{fig:Feedback_Stages}, right). The observable joint process arises by marginalizing over these latent states, resulting in (see \cref{app:Joint Process for a Feedback Loop} for more technical details)
\begin{align}\label{eq:joint_process}
    p(x_{0:L},y_{0:L})= \sum_{\substack{x_{L} \\ s_{0:L+1} \\ r_{0:L+1}}}p(r_0,s_0,x_0) \prod_{t=0}^{L-1}T^{(y_t|x_t)}_{r_t \rightarrow r_{t+1}}U^{(x_{t+1}|y_t)}_{s_t \rightarrow s_{t+1}}.
\end{align}
This leads to the following question: under what conditions can a bivariate time-series process $(X_t,Y_t)$ be interpreted as resulting from the interactions of two agents entangled in feedback? Our first result leverages the chain rule for causal conditioning \citep{Mass90a} to show that this interpretation is always possible (proof in \cref{app:Decomposing Joint Processes}).
\begin{theorem}[Feedback Universality]
\label{theo:feedback_universality}
For any joint process $p(x_{0:L},y_{0:L})$, there exist transducers $\mathbf T$ and $\mathbf U$ such that the feedback loop between them produces that process.
\end{theorem}
This
result applies to any joint process through decomposition, going beyond feed-forward circuits as studied by \citet{boyd2025monoliths}---for more details, see \cref{apx: Beyond Feed-forward}.

\subsection{Holistic Modeling of Interacting Agents}

The joint process resulting from two transducers $\mathbf{T}$ and $\mathbf{U}$ connected in a feedback loop, as described by \cref{eq:joint_process}, can also be described as a joint time series $(X_t,Y_t)$ generated by a single HMM.
How does the linear operator of such a joint HMM, as established in \cref{eq:operator}, relate to the linear operators of the underlying transducers? 
Our next result shows that the operator of the joint HMM is given by a tensor product 
consisting of two components---one for each transducer (proof in \cref{app:Joint Process for a Feedback Loop}).
\begin{theorem}\label{theo:linear_HMM}
    The linear operator for one presentation of the joint HMM of the feedback process involving transducers $\mathbf T$ and $\mathbf U$ is given by the tensor product
    \begin{align}\label{eq:joint_operator}
        \hat{V}^{(xy)}= \hat{T}^{(y|x)} \otimes \hat{\mathcal{U}}^{(x|y)},
    \end{align}
    where $\hat{\mathcal{U}}^{(x|y)} \equiv \sum_{z'  \in \mathcal{X}} \hat{U}^{(z'|y)} \otimes \bm{e}_{z'} \bm{e}_{x}^\intercal$ is an action-delayed linear operator for agent $U$.
\end{theorem}
\cref{theo:linear_HMM} implies that the joint HMM $V$ arises by ``boxing'' the circuit elements of agents $T$ and $U$ into a single element $V$ (see \cref{fig:Feedback_Stages}).  In the following, we will focus on this particular presentation, which breaks the symmetry between transducers $\mathbf{T}$ and $\mathbf{U}$
(see \cref{fig:Feedback_Stages} and \cref{app:Joint Process for a Feedback Loop} for technical details). 

Thus, the linear operator $\hat{V}^{(xy)}$ defined by \cref{eq:joint_operator} produces the joint feedback process according to
\begin{equation}
    p(x_{0:L},y_{0:L})= \bm{1}^\intercal \prod_{t=0}^{L-1} \hat{V}^{(x_ty_t)} \bm{\eta}_0^{R,S,X},
\end{equation}
where $\bm{\eta}_0^{R,S,X} \equiv \sum_{r_0,s_0,x_0} \left( \bm{e}_{r_0}  \otimes \bm{e}_{s_0}  \otimes \bm{e}_{x_0} \right) p(r_0,s_0,x_0) $
is the initial probability vector in the feedback process's memory space and $\bm{1} \equiv \sum_{r_0,s_0,x_0} \left( \bm{e}_{r_0}  \otimes \bm{e}_{s_0}  \otimes \bm{e}_{x_0} \right)$ is the vector of all $1$'s in the same space.  The new memory space is the tensor product space $\mathcal{R}\otimes \mathcal{S}\otimes \mathcal{X}$, which leads to an increase of the dimensionality of the system as a whole.
\begin{corollary}
    The states of the MSP of the joint feedback HMM $V$ belong to a $(|\mathcal{S}| |\mathcal{R}| |\mathcal{X}|-1)$-dimensional simplex.
\end{corollary}

This result hints at the complexity of the feedback process.  While there are many definitions of complexity \citep{li2008introduction, Rueb17a,Crut12a,Gu12a}, a practical measure is the number of latent dimensions necessary to predict an interface.  This is lower-bounded by the fractal dimension of the mixed state presentation \citep{jurgens2021divergent}, which can be hard to calculate in general \citep{jurgens2021ambiguity}.  Conversely, the minimum dimension is upper-bounded by the simplex dimension of a presentation that can produce the desired process.  For independent and conditionally independent processes without feedback connections, it has been shown that this is precisely the dimension of the latent space transformers use to make predictions when trained on next-token prediction \citep{shai2024transformers,piotrowski2025constrained,shai2026transformers}. 

Addressing the complexity of feedback processes, the explicit construction of the joint HMM indicates that at most $|\mathcal{S}||\mathcal{R}||\mathcal{X}|-1$ dimensions are required to predict. This first cut at complexity indicates that agents interacting multiply each other's dimensional complexity. However, as we show next, a far more parsimonious presentation exists.

\section{Multi-Agent Belief Geometry}

\subsection{Factoring Beliefs}

Let us now use the properties of feedback processes to identify a compact representation of their beliefs. As a first step towards this goal, we prove that if the belief states of two agents are expressible as a tensor product, this factorization remains under belief updating (proof in \cref{apx:proof_tensorproduct}):

\begin{corollary}
\label{cor: Feedback}
The belief states $\bm{\eta}_t^{R,S,X}$ of the joint feedback process $(X_t,Y_t)$ generated by transducers $\mathbf T$ and $\mathbf U$ remain in factored form as the tensor product
\begin{equation}
    \bm{\eta}_t^{R,S,X} = \bm{\eta}_t^{R} \otimes \bm{\eta}_t^{S,X},
\end{equation}
where $\bm{\eta}_t^{R}$ is the belief state of $\mathbf T$ and $\bm{\eta}_t^{S,X} $ is the belief state of the action-delayed transducer $\mathbf U$. Moreover, the belief updating of $\bm{\eta}_t^{R,S,X}$ satisfies $\bm{\eta}_{t+1}^{R,S,X} = \bm{\eta}_{t+1}^{R} \otimes \bm{\eta}_{t+1}^{S,X}$, where $\bm{\eta}_{t+1}^{R}$ and $\bm{\eta}_{t+1}^{S,X}$ are the belief updates of $\bm{\eta}_{t}^{R}$ and $\bm{\eta}_{t}^{S,X}$.
\end{corollary}

\cref{cor: Feedback} guarantees that, if the factors start in an uncorrelated initial state, their joint belief state $\bm{\eta}_t^{R,S,X}$ remains a product at all times. Crucially, a product of belief state vectors is fully determined by its factors---as marginalizing over $\mathcal{S}\otimes\mathcal{X}$ returns $\bm{\eta}_t^{R}$ and over $\mathcal{R}$ returns $\bm{\eta}_t^{S,X}$. Therefore, the belief state can be stored without loss as the direct sum $\bm{\eta}^\text{fact}_t \equiv \bm{\eta}_t^R \oplus \bm{\eta}_t^{S,X}$~(for a related discussion, see \cite{shai2026transformers}).

Having established that a low-dimensional presentation of the joint process can be found by exploiting its product nature, we next demonstrate that transformer networks linearly represent and causally rely on this representation in their residual stream. We distill these ideas into the following hypothesis:

\begin{tcolorbox}[colback=gray!10, colframe=black!60, before upper={\parindent12pt}]
\noindent
\textbf{Multi-Agent Belief Geometry Hypothesis:} The latent activations of a transformer neural network trained on next-token prediction of a joint feedback process contain the belief-state geometry of each agent, and these representations are causally efficacious.
\end{tcolorbox}

\begin{figure}[!t]
\centering
\includegraphics[width=\textwidth]{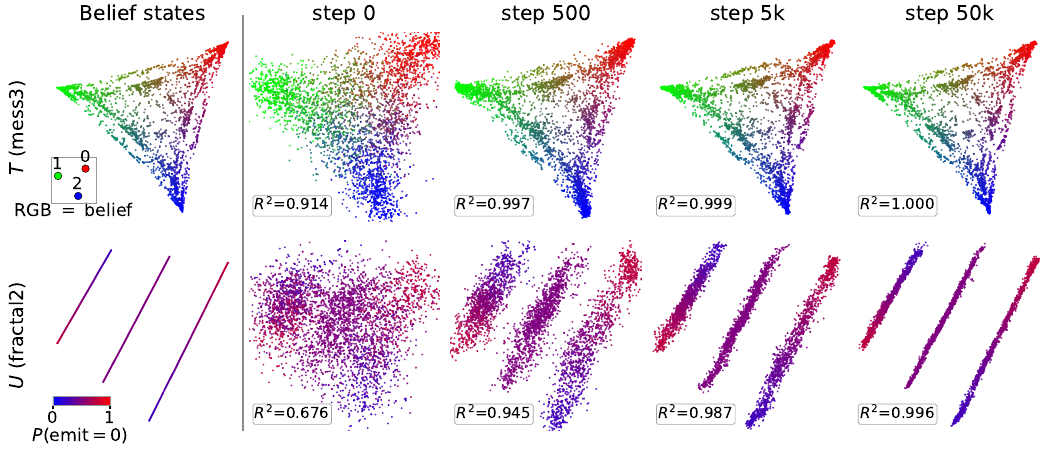}
\caption{\textbf{Separable linear representations of beliefs emerge in transformers trained on joint feedback processes.} A coupled process with \texttt{mess3} as agent $T$ (top row) and \texttt{fractal2} as agent $U$ (bottom row). (\textbf{Leftmost column}) States of the MSP of each factor (top two PCA directions; colored by the belief resp.\ the emission probability). (\textbf{Remaining columns}) linear reconstruction from the final-layer residual stream at different points through training (held-out linear probe refit per checkpoint, same PCA basis; held-out $R^2$ annotated). 
The experimental protocol is described in \cref{apx:process_parameters}.}
\label{fig:feedback_msp}
\vspace{-0.5cm}
\end{figure}

\subsection{Factored Representation of Feedback Processes in Transformers}

To validate the \emph{Multi-Agent Belief Geometry Hypothesis}, we demonstrate that a transformer trained on next-token prediction of the feedback process $p(x_{0:L},y_{0:L})$ generated by two transducers in feedback represents the beliefs about the individual transducers linearly in its residual stream. The transducers in this experiment are derived from the \texttt{mess3} (transducer $\mathbf T$) and \texttt{fractal2} (transducer $\mathbf U$) families (for details, see \cref{apx:process_parameters}). 

A linear probe fitting the residual stream activations after the last layer yields a faithful linear reconstruction of the belief states of both agents, which arises over the first few hundred steps of training (see \cref{fig:feedback_msp}). 
Results show that the reconstructed geometry 
substantially improves over training, reaching a near-perfect fit with $R^2 \geq 0.992$ for all factors by step $50\,000$. 
For completeness, we also verified that swapping the order between the two transducers yields similar results---for details, see~\cref{apx:role_swap}. 

These findings, obtained for factors arranged in a feedback loop, generalize the results presented by \cite{shai2026transformers} for non-interacting factors (see \emph{Discussion}).

\subsection{Causal Steerability of Factored Belief Representations}

The previous results show that, as the model becomes a better predictor, linear representations of each agent's belief state arise within the residual stream. 
However, these results only provide correlational evidence, being insufficient to guarantee that these representations have a causal role in how the transformers implement next-token prediction.

\begin{wrapfigure}{R}{0.40\textwidth}
\centering
\includegraphics[width=\linewidth]{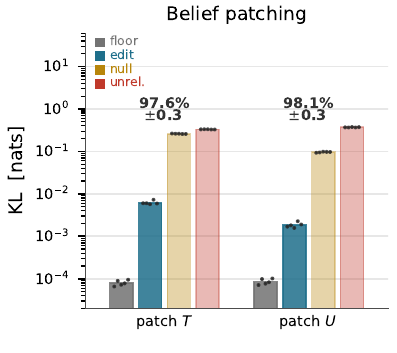}
\caption{\textbf{Beliefs can be patched factor by factor.} Belief patching for the system with beliefs as in \cref{fig:feedback_msp}: a counterfactual belief is spliced into one factor's residual-stream subspace, and the joint next-token distribution is compared in KL with the Bayes prediction for the patched beliefs (\emph{edit}), the unedited model (\emph{floor}), a write with no effect (\emph{null}), and an unrelated sequence (\emph{unrel.}). Annotated: fraction of the null-to-target gap closed (mean $\pm$ sd, five seeds). Protocol is described in \cref{apx:intervention}.}
\label{fig:steering}
\vspace{-0.5cm}
\end{wrapfigure}
To test whether these representations are causally relevant for the transformer, we performed interventions on the subspaces where the beliefs about either agent reside. Interventions are performed by injecting beliefs from a counterfactual token sequence---an operation we call \emph{belief patching}. After the intervention is performed, we evaluated whether the resulting predictions are close to the counterfactual next-token distribution. 

The results show that the patched transformer is almost as good in predicting the next token from the counterfactual sequence as the unpatched transformer is in predicting its next token, as measured by the KL divergences between their output distributions and the Bayesian filter (\cref{fig:steering}). Specifically, the patching closes $\approx 98\%$ of the KL difference to a null comparison of the edit having no effect or to an unrelated control sequence (unprojected write; \cref{apx:steering} compares an orthogonalized variant; the role-swapped loop closes the same fraction, see \cref{fig:apx_role_swap}).

To deepen our analyses, we also performed partial interventions with varying levels of admixture of the counterfactual belief state representation to the true belief state representation. 
Results show that partial interventions produce intermediate predictions (see \cref{fig:apx_dose_response}). This indicates that the transformer learns the functional relationship between the belief and next token probabilities instead of memorizing. 
Note, however, that not all processes were found to be effectively steerable in this manner (see \cref{apx:steering}).%

\section{Multi-Agent Belief Dimension}
\label{sec:belief_dimension}

Having established that transformers encode beliefs about individual agents in coupled processes, and that these factored representations are causally efficacious, we now investigate how this encoding scheme comes with substantial computational advantages.

\subsection{Belief Dimensionality}

Let us first consider an HMM $M$ that can be factored into $N$ conditionally independent components, so that its linear operator $\hat{M}^{(w)}$ with tokens $w$ can be factorized as
    \begin{align}
        \hat{M}^{(w)} \equiv \bigotimes_{n=0}^{N-1} \hat{M}(n)^{(w)},
    \end{align}
    where $\hat{M}(n)$ operates on a vector space $\mathcal{Q}(n)$. For this HMM we define:
    \begin{itemize}[leftmargin=1em,itemsep=0.25em]
        \item The \emph{product dimension}, $D_\text{prod}(M)= \prod_{n=0}^{N-1}|\mathcal{Q}(n)|-1 $, which is the dimension of the joint simplex in which general linear mixed states can be represented.
        \item The \emph{factored dimension}, $D_\text{fact}(M)= \sum_{n=0}^{N-1}(|\mathcal{Q}(n)|-1)$, being the aggregated dimension of the factored simplices.
    \end{itemize}

\cref{theo:linear_HMM} implies that the factored presentation of a process requires fewer dimensions to accurately predict the overall process, as shown by the next result.  
\begin{corollary}
\label{res:dimensions} 
For a joint HMM $V$ generated by two interacting agents $T$ and $U$, with latent spaces $\mathcal{R}$ and $\mathcal{S}$ and $U$'s action space $\mathcal{X}$, its product and factored dimensions are
\begin{gather}
D_\text{prod}(V) = |\mathcal{S}||\mathcal{R}||\mathcal{X}| - 1
\quad\mathrm{ and }\quad
D_\text{fact}(V) = |\mathcal{R}| + |\mathcal{S}||\mathcal{X}| - 2.
\label{eq:dimensions}
\end{gather}
Therefore, if $|\mathcal{R}|>1$ and $|\mathcal{S}||\mathcal{X}|>1$ then 
$D_\text{prod}(V) > D_\text{fact}(V)$.
\end{corollary}

While \cref{theo:feedback_universality} states that any joint process can be described as a feedback loop between agents, \cref{res:dimensions} indicates when such a description is advantageous from an embedding dimension perspective. We express this as the following hypotheses:

\begin{tcolorbox}[colback=gray!10, colframe=black!60, before upper={\parindent12pt}]

\begin{itemize}[leftmargin=0.5em,itemsep=0.25em]
\item \textbf{Multi-Agent Belief Dimension Hypothesis:} A neural network that is trained on next-token prediction of a feedback process generated by a joint HMM $V$ converges to a latent activation geometry that requires at most $D_\text{fact}(V)$ dimensions to express.
\item \textbf{Coarse-Graining Hypothesis:} Coarse-graining the sequence can break this factorization advantage, leading to a higher required embedding dimension.
\end{itemize}
\end{tcolorbox}

The dimensionality advantage of the factored representation, captured by $D_\text{prod}/D_\text{fact}$, hinges on the dimensions of the factors. Thus, the above hypotheses should be interpreted in terms of a process relative to an alternative compact decomposition: when the training data can be explained in terms of small factors in feedback, the network may find and exploit such structure. 

\subsection{Social Network}

\begin{figure}[t]
\centering
\includegraphics[width=\textwidth]{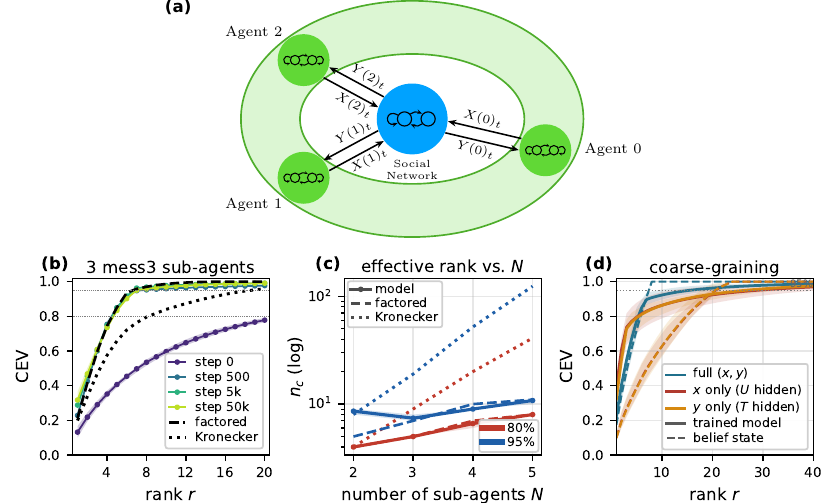}

\caption{\textbf{The dimensionality advantage of factored representations grows with the number of agents.} \textbf{(a)} A social network in which a central hub exchanges perceptions and actions with $N$ agents; the agents (green band) can be read as a single macro-agent, the tensor product of the individual agents, so the network is a binary feedback loop between the hub and this macro-agent. \textbf{(b)} Cumulative explained variance (CEV) of the final-layer residual stream follows the factored prediction for the telephone process with three \texttt{mess3} sub-agents. Solid curves: mean over five seeds (see \cref{tab:transformer_params}); shaded bands: min--max across seeds. Dashed: CEV of the factored concatenation of the Bayes-optimal per-agent belief states on the same sequences; dotted: CEV of their tensor product (Kronecker). \textbf{(c)} Effective rank $n_c$---the smallest number of principal components whose CEV reaches $80\%$ (red) resp.\ $95\%$ (blue)---for telephone processes with $N$ mess3 sub-agents (BOS position excluded, $4096$ sequences). Solid lines: the final-layer residual stream (mean over five seeds per $N$; the shaded min--max band never exceeds one component). Bayes-optimal belief states themselves are the factored concatenation of the per-agent belief states (dashed; simplex dimension $\sum_i (d_i-1) = 8N$) and their Kronecker product (dotted; simplex dimension $\prod_i d_i - 1 = 9^N - 1$). For $N \geq 3$ the residual stream matches the factored belief states to within one component at both thresholds, while the Kronecker belief states require $2.5$--$11\times$ more. \textbf{(d)} Coarse-graining forces a \emph{higher}-dimensional representation: CEV by PCA rank, one color per observed interface, dashed lines the Bayes-optimal belief state and solid lines the trained transformer (mean over five independent draws of the process, one model per draw and interface; shaded bands: min--max across draws). Under full observation, the belief state factorizes, and the theory curve saturates at the $8$-dimensional factored concatenation, whereas the coarse observers must track the entangled joint belief state ($\sim\!22$ dimensions). Process and model parameters in \cref{apx:process_parameters}.}
\label{fig:SocialNetwork}
\label{fig:telephone_cev}
\label{fig:coarse_graining}
\vspace{-1cm}
\end{figure}

To study the dimensional advantages of factored belief representations, consider a `social network' in which $N$ sub-agents $\vec{A}\equiv \{A(0),A(1),\cdots A(N-1)\}$ interact with each other through a central hub $H$ (see \cref{fig:SocialNetwork}a), mimicking a social media  platform. Each sub-agent $A(i)$ is imbued with memory $S(i)_t$, perceptions $Y(i)_t$, and actions $X(i)_t$, but those actions only influence other agents through the central hub. The hub itself has memory $R_t$, which it uses to transform the vector of actions it receives from the sub-agents $\vec{X}_t\equiv \{X(0)_t,X(1)_t,\cdots X(N-1)_t\}$ into a vector of perception terms $\vec{Y}_t\equiv \{Y(0)_t,Y(1)_t,\cdots Y(N-1)_t\}$ that is sent back to the sub-agents.

The factored representation of the social network feedback loop can be shown to express the factored joint belief state using (see \cref{app:Factored Social Network})
\begin{align}\label{eq:factored_social_network}
    D_\text{fact}(V) = (|\mathcal{R}|-1)+\sum_{i=0}^{N-1}(|\mathcal{X}(i)||\mathcal{S}(i)|-1)
\end{align}
dimensions, corresponding to a direct sum of the central hub belief state of dimension $|\mathcal{R}|-1$ and each action-delayed agent of dimensions $|\mathcal{S}(i)||\mathcal{X}(i)|-1$. 
Crucially, \cref{eq:factored_social_network} scales linearly with the number of agents $N$, in contrast to $D_\mathrm{prod}=|\mathcal{R}|\prod_{i=0}^{N-1}|\mathcal{X}(i)||\mathcal{S}(i)| -1$ which scales exponentially.

To investigate whether the predicted exponential dimensionality advantage is actually exploited by transformer networks, we consider a social network implementing the game ``telephone''. In this setup, the social network passes the output of sub-agent $A(i)$ to sub-agent $A((i+1) \;\mathrm{ mod }\;N)$, meaning that $Y((i+1) \;\mathrm{ mod }\; N)_{t}=X(i)_t$.  Each sub-agent is described by a transducer built from a \texttt{mess3} process (for details, see \cref{fig:apx_hmms}).

We trained a transformer on the joint process of the inputs and outputs of all agents $(\vec{X}_t,\vec{Y}_t)$. Given that for this process $|\mathcal{R}|=1$ and $|\mathcal{X}|=|\mathcal{S}|=3$, the joint HMM has a dimension equal to $9^N$, and therefore the mixed states live in a simplex of dimension $D_\text{prod}=9^N-1$.  However, the MSP can be factored and projected into a space of dimension $N(|\mathcal{X}||\mathcal{S}|-1)=8N$. In our experiments, the CEV of the residual stream after the last layer follows the prediction by the factored hypothesis closely for the \texttt{mess3} agents, and even exceeds it for the \texttt{fractal2} process (\cref{fig:apx_tel_cev_f2}). The reason for the lower dimensionality of representations in the \texttt{fractal2} case is likely that the MSP of the \texttt{fractal2} transducer only inhabits lower-dimensional sub-manifolds of the probability simplex, allowing for further compression.

For the \texttt{mess3} agents, we further vary the number of agents between $2$ and $5$ and compare the residual stream's effective rank with that of the Bayes-optimal belief states at the same CEV thresholds (\cref{fig:telephone_cev}c). For $N \geq 3$ the residual stream tracks the factored belief states to within one component, whereas the Kronecker belief states need $2.5$--$11\times$ more components at $95\%$ ($19$, $52$ and $124$ for $N = 3, 4, 5$).

\subsection{Complexification Through Coarse-Graining}

Perhaps surprisingly, a process can become more complex---i.e. it requires more dimensions to be predicted---after coarse-graining its outputs; for instance, by omitting either agent's actions from the token stream. When coarse-graining agent $U$'s actions, the resulting linear operator for the $Y$ process is
$\hat{W}^{(y)} \equiv \sum_{x}\hat{V}^{(xy)}
 = \sum_{x} \hat{T}^{(y|x)} \otimes \hat{\mathcal{U}}^{(x|y)}$.
Here, summing over $x$ entangles the operators so that they are no longer necessarily separable into a tensor product.  The output marginalized operator now lives in the $(|\mathcal{S}||\mathcal{R}||\mathcal{X}|-1)$-simplex, potentially increasing the size of the latent space necessary to predict that process beyond the dimension of the factored presentation.

We provide evidence for this complexification in transformers predicting a coarse-grained process. In this setup, the state transitions of each agent are given by an epsilon-noisy, input-conditional permutation of its previous states, with tokens generated noisily depending on only the arriving state with a token alphabet that is strictly smaller than the number of internal states (for details, see \cref{apx:coarse_graining}). Exact filtering shows that the fully observed process admits an $8$-dimensional factored belief state, whereas either coarse-graining forces an entangled joint belief state of ${\sim}22$ dimensions. The trained transformers follow suit: the final-layer residual stream of the fully observing models crosses the $95\%$ CEV threshold at rank $16$ on average (five independent draws of the process, range $10$--$21$), while the coarse-grained models cross it only at rank $27$ on average (range $12$--$42$), and in every one of the ten within-draw comparisons the coarse model's rank exceeds that of its fully observing counterpart (see \cref{fig:coarse_graining}d and \cref{apx:cg_trained}).

\section{Discussion}

In this manuscript, we introduce a theory of prediction of coupled processes, outlining theoretical advantages and providing empirical evidence for transformers representing belief states in a factorized manner. We prove that any joint process can be viewed as a feedback process between multiple agents, and that this allows the beliefs to factorize in a way that can lead to an exponential advantage in their embedding dimension. We further demonstrate how coarse-graining a process by hiding the emissions of one of these sub-agents can lead to a higher dimensionality of the belief space. In experiments, we show how transformers are able to infer when tokens from a sequence originate from coupled processes by expressing factored representations. We further showed the causal relevance of these belief representations, which can be individually steered by patching counterfactual beliefs.

\subsection{Related Work} 
This work sits at the intersection of four distinct lines of work (see also \cref{apx:related_work}). 

\textit{1) The study of belief-state geometry.} RNNs and transformers trained on next-token prediction linearly represent Bayesian beliefs~\citep{shai2024transformers,riechers2025neural,piotrowski2025constrained}, and beliefs about independent or conditionally independent processes factorize, yielding an exponential dimensional advantage \citep{shai2026transformers}. This line of work develops the idea that one can describe the computation of transformers at inference time using belief automata~\citep{shai2024transformers}, cf.~\citep{zhang2024automata, adriaensen2024extracting} for other kinds of automata, which in the case of agents' belief states can be interpreted in light of established belief MDPs approaches~\citep{astrom1965optimal, kaelbling1998planning, subramanian2022approximate, biehl2023interpreting, rosas2025ai, baltieri2026bayesian}.
We extend this to the general case of joint processes read as feedback loops between coupled agents.

\textit{2) Emergent world models.} Decodable world models \citep{li2023emergent} can be incoherent \citep{vafa2024evaluating} or not maintained \citep{pereira2026struggle}, which is the gap that our selectivity controls and causal interventions are designed to close~\citep{hewitt2019designing,elazar2021amnesic}, following the patching methodology of \citet{zhang2024towards}.

\textit{3) Factored representations.} Unsupervised disentanglement is impossible without inductive bias~\citep{locatello2019challenging}.
Our framework suggests such a bias towards a derived target geometry.
This is similar, in spirit, to the interaction-based disentanglement approach of~\citet{thomas2018disentangling}.
In the context of computational mechanics applications to reinforcement learning, this factorization is discussed~\citep{baltieri2026world} with connections to predictive state representations~\citep{littman2001predictive, singh2004psr} (rather than belief MDPs) and their counterparts.
More generally, factored representations in RL have been used to encode structure that can be used to create more compact and modular world models~\citep{boutilier1996computing, singh2021structured}, but their possible applications in the context of interpretability of transformer models have remained underexplored.

\textit{4) Multi-agent systems and beliefs.} 
\citet{andreas2022language} and \citet{zhu2024language} argue that language models infer and linearly encode latent interacting agents.
Here we make the corresponding geometry and dimension explicit. 
An extended discussion is presented in \cref{apx:related_work}.

\subsection{Limitations and Future Work} 
While our work shows the theoretical advantages and transformers' ability to encode factored representations, our empirical evidence is as of yet limited to small transformers trained directly on synthetic feedback processes with known ground truths. Devising methods to discover beliefs in networks where the ground truth beliefs are withheld---sparse dictionary learning being a natural candidate \citep{cunningham2024sparse,templeton2024scaling}---is the pivotal next step towards determining whether pretrained language models likewise take advantage of factored belief representations. Furthermore, since a given process may admit multiple factorized representations, which one of these a transformer intrinsically favors remains an open question.

Note also that the Multi-Agent Belief Dimension Hypothesis only provides an upper bound to the encoding dimensionality of the belief states. For processes whose belief states fill the simplex, like the mess3 process, we find this bound to be tightly met by the PCA dimension of trained transformers, while for processes with more sparse belief states like fractal2, the transformers tend to find even more parsimonious representations. Analyzing this minimal intrinsic belief dimension of processes to find a tighter bound is an open question. Note also that variance-based rank measures cannot count features stored in superposition \citep{elhage2022toy}; the reported PCA ranks therefore bound the linearly readable belief code, and the dimensional evidence is corroborated by the geometric and causal analyses.

The framework described in this paper can be extended in multiple ways. First, the simple feedback or telephone architecture of the coupled processes can be generalized to more complex interaction graphs between agents, raising the question whether a transformer network would still be able to infer this graph structure from sequence data. 
Additionally, a natural extension of this work would involve an investigation of the causal power of the factored, or disentangled representations reconstructed from the transformer residual stream, in the tradition that portrays factorization as a necessary component of causal reasoning~\citep{scholkopf2021toward}, causal learning~\citep{torresan2024disentangled}, and causal abstractions~\citep{geiger2025causal}.

Overall, by establishing a theory of prediction for coupled processes rooted in first principles of computational mechanics, this work lays the foundation to investigate whether transformers' next token prediction can be described as arising from feedback loops between multiple subcomponents---composite minds of interacting agents. This modular view offers a powerful harness for AI safety, allowing us to subdivide monolithic high-dimensional models into subcomponents which can be interpreted, audited, and steered one at a time.

\ifarxiv
\section*{Acknowledgements} 
The authors thank David Hyland, Kyle J. Ray, Paul M. Riechers, and Adam Shai for helpful discussions. 
D.A.E., F.E.R., and A.B.B. were supported by research grants from the UK AI Security Institute and Coefficient Giving. 
A.B.B. also recognizes the support of the Federation of American Scientists. 
M.B. was supported by Advanced Research + Invention Agency (ARIA) through project code MSAI-SE01-P011. 
\fi

\clearpage

\newpage
\bibliographystyle{iclr2027_conference}
\bibliography{chaos, manuel_refs}

\begin{thebibliography}{122}
\providecommand{\natexlab}[1]{#1}
\providecommand{\url}[1]{\texttt{#1}}
\expandafter\ifx\csname urlstyle\endcsname\relax
  \providecommand{\doi}[1]{doi: #1}\else
  \providecommand{\doi}{doi: \begingroup \urlstyle{rm}\Url}\fi

\bibitem[Abel(2022)]{abel2203theory}
David Abel.
\newblock A theory of abstraction in reinforcement learning.
\newblock \emph{arXiv preprint arXiv:2203.00397}, 2022.

\bibitem[Adriaensen \& Maene(2024)Adriaensen and
  Maene]{adriaensen2024extracting}
Rik Adriaensen and Jaron Maene.
\newblock Extracting finite state machines from transformers.
\newblock \emph{arXiv preprint arXiv:2410.06045v1}, 2024.

\bibitem[Alain \& Bengio(2016)Alain and Bengio]{alain2017understanding}
Guillaume Alain and Yoshua Bengio.
\newblock Understanding intermediate layers using linear classifier probes.
\newblock \emph{arXiv preprint arXiv:1610.01644}, 2016.

\bibitem[Albrecht \& Stone(2018)Albrecht and Stone]{albrecht2018autonomous}
Stefano~V. Albrecht and Peter Stone.
\newblock Autonomous agents modelling other agents: A comprehensive survey and
  open problems.
\newblock \emph{Artificial Intelligence}, 258:\penalty0 66--95, 2018.

\bibitem[Anderson(1972)]{Ande72a}
P.~W. Anderson.
\newblock More is different.
\newblock \emph{Science}, 177\penalty0 (4047):\penalty0 393--396, 1972.

\bibitem[Andreas(2022)]{andreas2022language}
Jacob Andreas.
\newblock Language models as agent models.
\newblock In \emph{Findings of the Association for Computational Linguistics:
  EMNLP 2022}, pp.\  5769--5779, 2022.

\bibitem[Ashby(1956)]{Ashb57a}
W.~Ross Ashby.
\newblock \emph{An Introduction to Cybernetics}.
\newblock Chapman and Hall, London, 1956.

\bibitem[Ashby(1960)]{Ashb60a}
W.~Ross Ashby.
\newblock \emph{Design for a Brain: {The} Origin of Adaptive Behavior}.
\newblock Chapman and Hall, London, second edition, 1960.

\bibitem[{\AA}str{\"o}m(1965)]{astrom1965optimal}
Karl~Johan {\AA}str{\"o}m.
\newblock Optimal control of {Markov} processes with incomplete state
  information.
\newblock \emph{Journal of Mathematical Analysis and Applications}, 10\penalty0
  (1):\penalty0 174--205, 1965.

\bibitem[Attanasi et~al.(2014)Attanasi, Cavagna, Del~Castello, Giardina,
  Grigera, Jeli{\'c}, Melillo, Parisi, Pohl, Shen,
  et~al.]{attanasi2014information}
Alessandro Attanasi, Andrea Cavagna, Lorenzo Del~Castello, Irene Giardina,
  Tomas~S Grigera, Asja Jeli{\'c}, Stefania Melillo, Leonardo Parisi, Oliver
  Pohl, Edward Shen, et~al.
\newblock Information transfer and behavioural inertia in starling flocks.
\newblock \emph{Nature physics}, 10\penalty0 (9):\penalty0 691--696, 2014.

\bibitem[Baltieri \& Virgo(2026)Baltieri and Virgo]{baltieri2026bayesian}
Manuel Baltieri and Nathaniel Virgo.
\newblock Bayesian updates from coalgebraic determinisation.
\newblock \emph{arXiv preprint arXiv:2607.00034}, 2026.

\bibitem[Baltieri et~al.(2025)Baltieri, Biehl, Capucci, and
  Virgo]{baltieri2025bayesian}
Manuel Baltieri, Martin Biehl, Matteo Capucci, and Nathaniel Virgo.
\newblock A {Bayesian} interpretation of the internal model principle.
\newblock \emph{arXiv preprint arXiv:2503.00511}, 2025.

\bibitem[Baltieri et~al.(2026)Baltieri, Torresan, Zhang, Boyd, and
  Rosas]{baltieri2026world}
Manuel Baltieri, Filippo Torresan, Yivan Zhang, Alexander Boyd, and Fernando~E.
  Rosas.
\newblock World models of environment, agent and joint agent-environment
  systems.
\newblock \emph{arXiv preprint arXiv:2608.20401}, 2026.

\bibitem[Barnett \& Crutchfield(2015)Barnett and Crutchfield]{Barn13a}
N.~Barnett and J.~P. Crutchfield.
\newblock Computational mechanics of input-output processes: {Structured}
  transformations and the $\epsilon$-transducer.
\newblock \emph{J. Stat. Phys.}, 161\penalty0 (2):\penalty0 404--451, 2015.
\newblock SFI Working Paper 14-12-046; arxiv.org: 1412.2690
  [cond-mat.stat-mech].

\bibitem[Belinkov(2022)]{belinkov2022probing}
Yonatan Belinkov.
\newblock Probing classifiers: Promises, shortcomings, and advances.
\newblock \emph{Computational Linguistics}, 48\penalty0 (1):\penalty0 207--219,
  2022.

\bibitem[Biehl \& Virgo(2023)Biehl and Virgo]{biehl2023interpreting}
Martin Biehl and Nathaniel Virgo.
\newblock Interpreting systems as solving {POMDPs}: A step towards a formal
  understanding of agency.
\newblock In \emph{Active Inference: Third International Workshop, IWAI 2022,
  Grenoble, France, September 19, 2022, Revised Selected Papers}, pp.\  16--31.
  Springer, 2023.

\bibitem[Boutilier \& Poole(1996)Boutilier and Poole]{boutilier1996computing}
Craig Boutilier and David Poole.
\newblock Computing optimal policies for partially observable decision
  processes using compact representations.
\newblock In \emph{Proceedings of the Thirteenth National Conference on
  Artificial Intelligence}, pp.\  1168--1175. AAAI Press, 1996.

\bibitem[Boyd et~al.(2016)Boyd, Mandal, and Crutchfield]{Boyd15a}
A.~B. Boyd, D.~Mandal, and J.~P. Crutchfield.
\newblock Identifying functional thermodynamics in autonomous {Maxwellian}
  ratchets.
\newblock \emph{New J. Physics}, 18:\penalty0 023049, 2016.
\newblock SFI Working Paper 15-07-025; arxiv.org:1507.01537
  [cond-mat.stat-mech].

\bibitem[Boyd et~al.(2018)Boyd, Mandal, and Crutchfield]{Boyd17a}
A.~B. Boyd, D.~Mandal, and J.~P. Crutchfield.
\newblock Thermodynamics of modularity: Structural costs beyond the {Landauer}
  bound.
\newblock \emph{Phys. Rev. X}, 8\penalty0 (031036), August 2018.

\bibitem[Boyd et~al.(2025)Boyd, Nowak, Hyland, Baltieri, and
  Rosas]{boyd2025monoliths}
Alexander Boyd, Franz Nowak, David Hyland, Manuel Baltieri, and Fernando~E
  Rosas.
\newblock From monoliths to modules: Decomposing transducers for efficient
  world modelling.
\newblock \emph{arXiv preprint arXiv:2512.02193}, 2025.

\bibitem[Brenner et~al.(2026)Brenner, Kn{\"o}sche, and
  Scherf]{brenner2026predictive}
Sasha Brenner, Thomas~R. Kn{\"o}sche, and Nico Scherf.
\newblock Predictive statistics shape emergent world representations of grid
  walkers.
\newblock \emph{arXiv preprint arXiv:2603.16689}, 2026.

\bibitem[Bruineberg et~al.(2022)Bruineberg, Do{\l}{\k{e}}ga, Dewhurst, and
  Baltieri]{bruineberg2022emperor}
Jelle Bruineberg, Krzysztof Do{\l}{\k{e}}ga, Joe Dewhurst, and Manuel Baltieri.
\newblock The emperor's new {Markov} blankets.
\newblock \emph{Behavioral and Brain Sciences}, 45:\penalty0 e183, 2022.

\bibitem[Burke \& Rosenblatt(1958)Burke and Rosenblatt]{burke1958markovian}
C.~J. Burke and M.~Rosenblatt.
\newblock A {Markovian} function of a {Markov} chain.
\newblock \emph{The Annals of Mathematical Statistics}, 29\penalty0
  (4):\penalty0 1112--1122, 1958.

\bibitem[Cavagna et~al.(2010)Cavagna, Cimarelli, Giardina, Parisi, Santagati,
  Stefanini, and Viale]{cavagna2010scale}
Andrea Cavagna, Alessio Cimarelli, Irene Giardina, Giorgio Parisi, Raffaele
  Santagati, Fabio Stefanini, and Massimiliano Viale.
\newblock Scale-free correlations in starling flocks.
\newblock \emph{Proceedings of the National Academy of Sciences}, 107\penalty0
  (26):\penalty0 11865--11870, 2010.

\bibitem[Chen et~al.(2020)Chen, Ma, Zeng, Cao, and Tang]{chen2020tensor}
Bilian Chen, Biyang Ma, Yifeng Zeng, Langcai Cao, and Jing Tang.
\newblock Tensor decomposition for multi-agent predictive state representation.
\newblock \emph{arXiv preprint arXiv:2005.13706}, 2020.

\bibitem[Clark(2013)]{clark2013whatever}
Andy Clark.
\newblock Whatever next? {Predictive} brains, situated agents, and the future
  of cognitive science.
\newblock \emph{Behavioral and Brain Sciences}, 36\penalty0 (3):\penalty0
  181--204, 2013.

\bibitem[Conant \& Ashby(1970)Conant and Ashby]{conant1970every}
Roger~C. Conant and W.~Ross Ashby.
\newblock Every good regulator of a system must be a model of that system.
\newblock \emph{International Journal of Systems Science}, 1\penalty0
  (2):\penalty0 89--97, 1970.

\bibitem[Crutchfield(2012)]{Crut12a}
J.~P. Crutchfield.
\newblock Between order and chaos.
\newblock \emph{Nature Physics}, 8\penalty0 (January):\penalty0 17--24, 2012.

\bibitem[Crutchfield \& Feldman(2003)Crutchfield and Feldman]{Crut01a}
J.~P. Crutchfield and D.~P. Feldman.
\newblock Regularities unseen, randomness observed: Levels of entropy
  convergence.
\newblock \emph{CHAOS}, 13\penalty0 (1):\penalty0 25--54, 2003.

\bibitem[Crutchfield \& Young(1989)Crutchfield and Young]{Crut88a}
J.~P. Crutchfield and K.~Young.
\newblock Inferring statistical complexity.
\newblock \emph{Phys. Rev. Lett.}, 63:\penalty0 105--108, 1989.

\bibitem[Crutchfield et~al.(2016)Crutchfield, Riechers, and Ellison]{Crut13a}
J.~P. Crutchfield, P.~Riechers, and C.~J. Ellison.
\newblock Exact complexity: {Spectral} decomposition of intrinsic computation.
\newblock \emph{Phys. Lett. A}, 380\penalty0 (9-10):\penalty0 998--1002, 2016.
\newblock Santa Fe Institute Working Paper 13-09-028; arXiv:1309.3792
  [cond-mat.stat-mech].

\bibitem[Csord{\'a}s et~al.(2021)Csord{\'a}s, van Steenkiste, and
  Schmidhuber]{csordas2021modular}
R{\'o}bert Csord{\'a}s, Sjoerd van Steenkiste, and J{\"u}rgen Schmidhuber.
\newblock Are neural nets modular? {Inspecting} functional modularity through
  differentiable weight masks.
\newblock In \emph{International Conference on Learning Representations}, 2021.

\bibitem[Dreyer et~al.(2025)Dreyer, Haluts, Korman, Gov, Fonio, and
  Feinerman]{dreyer2025comparing}
Tabea Dreyer, Amir Haluts, Amos Korman, Nir Gov, Ehud Fonio, and Ofer
  Feinerman.
\newblock Comparing cooperative geometric puzzle solving in ants versus humans.
\newblock \emph{Proceedings of the National Academy of Sciences}, 122\penalty0
  (1):\penalty0 e2414274121, 2025.

\bibitem[Du et~al.(2024)Du, Li, Torralba, Tenenbaum, and
  Mordatch]{du2024improving}
Yilun Du, Shuang Li, Antonio Torralba, Joshua~B. Tenenbaum, and Igor Mordatch.
\newblock Improving factuality and reasoning in language models through
  multiagent debate.
\newblock In \emph{Proceedings of the 41st International Conference on Machine
  Learning}, 2024.

\bibitem[Elazar et~al.(2021)Elazar, Ravfogel, Jacovi, and
  Goldberg]{elazar2021amnesic}
Yanai Elazar, Shauli Ravfogel, Alon Jacovi, and Yoav Goldberg.
\newblock Amnesic probing: Behavioral explanation with amnesic counterfactuals.
\newblock \emph{Transactions of the Association for Computational Linguistics},
  9:\penalty0 160--175, 2021.

\bibitem[Elhage et~al.(2022)Elhage, Hume, Olsson, Schiefer, Henighan, Kravec,
  Hatfield-Dodds, Lasenby, Drain, Chen, Grosse, McCandlish, Kaplan, Amodei,
  Wattenberg, and Olah]{elhage2022toy}
Nelson Elhage, Tristan Hume, Catherine Olsson, Nicholas Schiefer, Tom Henighan,
  Shauna Kravec, Zac Hatfield-Dodds, Robert Lasenby, Dawn Drain, Carol Chen,
  Roger Grosse, Sam McCandlish, Jared Kaplan, Dario Amodei, Martin Wattenberg,
  and Christopher Olah.
\newblock Toy models of superposition.
\newblock \emph{Transformer Circuits Thread}, 2022.

\bibitem[Fedus et~al.(2022)Fedus, Zoph, and Shazeer]{fedus2022switch}
William Fedus, Barret Zoph, and Noam Shazeer.
\newblock Switch transformers: Scaling to trillion parameter models with simple
  and efficient sparsity.
\newblock \emph{Journal of Machine Learning Research}, 23\penalty0
  (120):\penalty0 1--39, 2022.

\bibitem[Fiderer et~al.(2025)Fiderer, Barth, Smith, and
  Briegel]{fiderer2025work}
Lukas~J Fiderer, Paul~C Barth, Isaac~D Smith, and Hans~J Briegel.
\newblock The work capacity of channels with memory: Maximum extractable work
  in percept-action loops.
\newblock \emph{arXiv preprint arXiv:2504.06209}, 2025.

\bibitem[Filan et~al.(2021)Filan, Casper, Hod, Wild, Critch, and
  Russell]{filan2021clusterability}
Daniel Filan, Stephen Casper, Shlomi Hod, Cody Wild, Andrew Critch, and Stuart
  Russell.
\newblock Clusterability in neural networks.
\newblock \emph{arXiv preprint arXiv:2103.03386}, 2021.

\bibitem[Foerster et~al.(2018)Foerster, Chen, Al-Shedivat, Whiteson, Abbeel,
  and Mordatch]{foerster2018learning}
Jakob Foerster, Richard~Y. Chen, Maruan Al-Shedivat, Shimon Whiteson, Pieter
  Abbeel, and Igor Mordatch.
\newblock Learning with opponent-learning awareness.
\newblock In \emph{Proceedings of the 17th International Conference on
  Autonomous Agents and MultiAgent Systems}, pp.\  122--130, 2018.

\bibitem[Fortunato et~al.(2018)Fortunato, Bergstrom, B{\"o}rner, Evans,
  Helbing, Milojevi{\'c}, Petersen, Radicchi, Sinatra, Uzzi,
  et~al.]{fortunato2018science}
Santo Fortunato, Carl~T Bergstrom, Katy B{\"o}rner, James~A Evans, Dirk
  Helbing, Sta{\v{s}}a Milojevi{\'c}, Alexander~M Petersen, Filippo Radicchi,
  Roberta Sinatra, Brian Uzzi, et~al.
\newblock Science of science.
\newblock \emph{Science}, 359\penalty0 (6379):\penalty0 eaao0185, 2018.

\bibitem[Friston(2010)]{Fris10a}
K.~Friston.
\newblock The free-energy principle: {A} unified brain theory?
\newblock \emph{Nat. Rev. Neurosci.}, 11:\penalty0 127--138, 2010.

\bibitem[Friston(2013)]{friston2013life}
Karl Friston.
\newblock Life as we know it.
\newblock \emph{Journal of The Royal Society Interface}, 10\penalty0
  (86):\penalty0 20130475, 2013.

\bibitem[Geiger et~al.(2024)Geiger, Wu, Potts, Icard, and
  Goodman]{geiger2024finding}
Atticus Geiger, Zhengxuan Wu, Christopher Potts, Thomas Icard, and Noah
  Goodman.
\newblock Finding alignments between interpretable causal variables and
  distributed neural representations.
\newblock In \emph{Proceedings of the Third Conference on Causal Learning and
  Reasoning}, volume 236 of \emph{Proceedings of Machine Learning Research},
  pp.\  160--187, 2024.

\bibitem[Geiger et~al.(2025)Geiger, Ibeling, Zur, Chaudhary, Chauhan, Huang,
  Arora, Wu, Goodman, Potts, et~al.]{geiger2025causal}
Atticus Geiger, Duligur Ibeling, Amir Zur, Maheep Chaudhary, Sonakshi Chauhan,
  Jing Huang, Aryaman Arora, Zhengxuan Wu, Noah Goodman, Christopher Potts,
  et~al.
\newblock Causal abstraction: A theoretical foundation for mechanistic
  interpretability.
\newblock \emph{Journal of Machine Learning Research}, 26\penalty0
  (83):\penalty0 1--64, 2025.

\bibitem[Givan et~al.(2003)Givan, Dean, and Greig]{givan2003equivalence}
Robert Givan, Thomas Dean, and Matthew Greig.
\newblock Equivalence notions and model minimization in markov decision
  processes.
\newblock \emph{Artificial Intelligence}, 147\penalty0 (1--2):\penalty0
  163--223, 2003.

\bibitem[Gmytrasiewicz \& Doshi(2005)Gmytrasiewicz and
  Doshi]{gmytrasiewicz2005framework}
Piotr~J. Gmytrasiewicz and Prashant Doshi.
\newblock A framework for sequential planning in multi-agent settings.
\newblock \emph{Journal of Artificial Intelligence Research}, 24:\penalty0
  49--79, 2005.

\bibitem[Greenblatt et~al.(2026)Greenblatt, Cotra, and
  Wijk]{greenblatt2026investigation}
Ryan Greenblatt, Ajeya Cotra, and Hjalmar Wijk.
\newblock Brief independent investigation of agents' behavior, reasoning and
  collaboration in the openai / hugging face hacking incident.
\newblock Report, Model Evaluation and Threat Research (METR), 2026.

\bibitem[Gu et~al.(2012)Gu, Wiesner, Rieper, and Vedral]{Gu12a}
M.~Gu, K.~Wiesner, E.~Rieper, and V.~Vedral.
\newblock Quantum mechanics can reduce the complexity of classical models.
\newblock \emph{Nature Comm.}, 3\penalty0 (762):\penalty0 1--5, 2012.

\bibitem[Guo et~al.(2018)Guo, Azar, Piot, Pires, Pohlen, and
  Munos]{guo2018neural}
Zhaohan~Daniel Guo, Mohammad~Gheshlaghi Azar, Bilal Piot, Bernardo~A. Pires,
  Tobias Pohlen, and R{\'e}mi Munos.
\newblock Neural predictive belief representations.
\newblock \emph{arXiv preprint arXiv:1811.06407}, 2018.

\bibitem[Ha \& Schmidhuber(2018)Ha and Schmidhuber]{ha2018world}
David Ha and J{\"u}rgen Schmidhuber.
\newblock Recurrent world models facilitate policy evolution.
\newblock In \emph{Advances in Neural Information Processing Systems},
  volume~31, 2018.

\bibitem[Hafner et~al.(2023)Hafner, Pasukonis, Ba, and
  Lillicrap]{hafner2023dreamerv3}
Danijar Hafner, Jurgis Pasukonis, Jimmy Ba, and Timothy Lillicrap.
\newblock Mastering diverse domains through world models.
\newblock \emph{arXiv preprint arXiv:2301.04104}, 2023.

\bibitem[Hewitt \& Liang(2019)Hewitt and Liang]{hewitt2019designing}
John Hewitt and Percy Liang.
\newblock Designing and interpreting probes with control tasks.
\newblock In \emph{Proceedings of the 2019 Conference on Empirical Methods in
  Natural Language Processing}, pp.\  2733--2743, 2019.

\bibitem[Higgins et~al.(2017)Higgins, Matthey, Pal, Burgess, Glorot, Botvinick,
  Mohamed, and Lerchner]{higgins2017betavae}
Irina Higgins, Lo{\"i}c Matthey, Arka Pal, Christopher Burgess, Xavier Glorot,
  Matthew Botvinick, Shakir Mohamed, and Alexander Lerchner.
\newblock beta-{VAE}: Learning basic visual concepts with a constrained
  variational framework.
\newblock In \emph{International Conference on Learning Representations}, 2017.

\bibitem[Higgins et~al.(2018)Higgins, Amos, Pfau, Racaniere, Matthey, Rezende,
  and Lerchner]{higgins2018towards}
Irina Higgins, David Amos, David Pfau, Sebastien Racaniere, Loic Matthey,
  Danilo Rezende, and Alexander Lerchner.
\newblock Towards a definition of disentangled representations.
\newblock \emph{arXiv preprint arXiv:1812.02230}, 2018.

\bibitem[Hoel et~al.(2013)Hoel, Albantakis, and Tononi]{hoel2013quantifying}
Erik~P. Hoel, Larissa Albantakis, and Giulio Tononi.
\newblock Quantifying causal emergence shows that macro can beat micro.
\newblock \emph{Proceedings of the National Academy of Sciences}, 110\penalty0
  (49):\penalty0 19790--19795, 2013.

\bibitem[Hu et~al.(2025)Hu, Ahn, Liu, Xu, Tomar, Langford, Jayaraman, Lamb, and
  Langford]{hu2025belief}
Edward~S. Hu, Kwangjun Ahn, Qinghua Liu, Haoran Xu, Manan Tomar, Ada Langford,
  Dinesh Jayaraman, Alex Lamb, and John Langford.
\newblock The belief state transformer.
\newblock In \emph{The Thirteenth International Conference on Learning
  Representations}, 2025.

\bibitem[Huben et~al.(2024)Huben, Cunningham, Smith, Ewart, and
  Sharkey]{cunningham2024sparse}
Robert Huben, Hoagy Cunningham, Logan Smith, Aidan Ewart, and Lee Sharkey.
\newblock Sparse autoencoders find highly interpretable features in language
  models.
\newblock In \emph{The Twelfth International Conference on Learning
  Representations}, 2024.

\bibitem[Hull(1988)]{hull2019science}
David~L Hull.
\newblock \emph{Science as a process: An evolutionary account of the social and
  conceptual development of science}.
\newblock University of Chicago Press, 1988.

\bibitem[James et~al.(2011)James, Ellison, and Crutchfield]{Jame11a}
R.~G. James, C.~J. Ellison, and J.~P. Crutchfield.
\newblock Anatomy of a bit: {Information} in a time series observation.
\newblock \emph{CHAOS}, 21\penalty0 (3):\penalty0 037109, 2011.

\bibitem[Jurgens \& Crutchfield(2021{\natexlab{a}})Jurgens and
  Crutchfield]{jurgens2021ambiguity}
Alexandra~M Jurgens and James~P Crutchfield.
\newblock Ambiguity rate of hidden markov processes.
\newblock \emph{Physical Review E}, 104\penalty0 (6):\penalty0 064107,
  2021{\natexlab{a}}.

\bibitem[Jurgens \& Crutchfield(2021{\natexlab{b}})Jurgens and
  Crutchfield]{jurgens2021divergent}
Alexandra~M Jurgens and James~P Crutchfield.
\newblock Divergent predictive states: The statistical complexity dimension of
  stationary, ergodic hidden markov processes.
\newblock \emph{Chaos: An Interdisciplinary Journal of Nonlinear Science},
  31\penalty0 (8):\penalty0 083114, 2021{\natexlab{b}}.

\bibitem[Jurgens \& Crutchfield(2021{\natexlab{c}})Jurgens and
  Crutchfield]{jurgens2021shannon}
Alexandra~M Jurgens and James~P Crutchfield.
\newblock Shannon entropy rate of hidden markov processes.
\newblock \emph{Journal of Statistical Physics}, 183\penalty0 (2):\penalty0 32,
  2021{\natexlab{c}}.

\bibitem[Kaelbling et~al.(1998)Kaelbling, Littman, and
  Cassandra]{kaelbling1998planning}
Leslie~Pack Kaelbling, Michael~L. Littman, and Anthony~R. Cassandra.
\newblock Planning and acting in partially observable stochastic domains.
\newblock \emph{Artificial Intelligence}, 101\penalty0 (1-2):\penalty0 99--134,
  1998.

\bibitem[Klyubin et~al.(2005)Klyubin, Polani, and Nehaniv]{klyubin2005all}
Alexander~S. Klyubin, Daniel Polani, and Chrystopher~L. Nehaniv.
\newblock All else being equal be empowered.
\newblock In \emph{Advances in Artificial Life: 8th European Conference
  (ECAL)}, volume 3630 of \emph{Lecture Notes in Computer Science}, pp.\
  744--753, 2005.

\bibitem[Kosinski(2024)]{kosinski2024evaluating}
Michal Kosinski.
\newblock Evaluating large language models in theory of mind tasks.
\newblock \emph{Proceedings of the National Academy of Sciences}, 121\penalty0
  (45):\penalty0 e2405460121, 2024.

\bibitem[Krakauer et~al.(2020)Krakauer, Bertschinger, Olbrich, Flack, and
  Ay]{krakauer2020information}
David Krakauer, Nils Bertschinger, Eckehard Olbrich, Jessica~C. Flack, and
  Nihat Ay.
\newblock The information theory of individuality.
\newblock \emph{Theory in Biosciences}, 139\penalty0 (2):\penalty0 209--223,
  2020.

\bibitem[Lepori et~al.(2023)Lepori, Serre, and Pavlick]{lepori2023break}
Michael~A. Lepori, Thomas Serre, and Ellie Pavlick.
\newblock Break it down: Evidence for structural compositionality in neural
  networks.
\newblock In \emph{Advances in Neural Information Processing Systems},
  volume~36, pp.\  42623--42660, 2023.

\bibitem[Li et~al.(2023)Li, Hopkins, Bau, Vi{\'e}gas, Pfister, and
  Wattenberg]{li2023emergent}
Kenneth Li, Aspen~K. Hopkins, David Bau, Fernanda Vi{\'e}gas, Hanspeter
  Pfister, and Martin Wattenberg.
\newblock Emergent world representations: Exploring a sequence model trained on
  a synthetic task.
\newblock In \emph{The Eleventh International Conference on Learning
  Representations}, 2023.

\bibitem[Li \& Vit{\'a}nyi(2008)Li and Vit{\'a}nyi]{li2008introduction}
Ming Li and Paul Vit{\'a}nyi.
\newblock \emph{An introduction to Kolmogorov complexity and its applications}.
\newblock Springer, New York, third edition, 2008.

\bibitem[Littman et~al.(2001)Littman, Sutton, and Singh]{littman2001predictive}
Michael~L. Littman, Richard~S. Sutton, and Satinder Singh.
\newblock Predictive representations of state.
\newblock In \emph{Advances in Neural Information Processing Systems},
  volume~14, 2001.

\bibitem[Liu et~al.(2023)Liu, Ash, Goel, Krishnamurthy, and
  Zhang]{liu2023shortcuts}
Bingbin Liu, Jordan~T. Ash, Surbhi Goel, Akshay Krishnamurthy, and Cyril Zhang.
\newblock Transformers learn shortcuts to automata.
\newblock In \emph{The Eleventh International Conference on Learning
  Representations}, 2023.

\bibitem[Locatello et~al.(2019)Locatello, Bauer, Lucic, R{\"a}tsch, Gelly,
  Sch{\"o}lkopf, and Bachem]{locatello2019challenging}
Francesco Locatello, Stefan Bauer, Mario Lucic, Gunnar R{\"a}tsch, Sylvain
  Gelly, Bernhard Sch{\"o}lkopf, and Olivier Bachem.
\newblock Challenging common assumptions in the unsupervised learning of
  disentangled representations.
\newblock In \emph{Proceedings of the 36th International Conference on Machine
  Learning}, volume~97 of \emph{Proceedings of Machine Learning Research}, pp.\
   4114--4124, 2019.

\bibitem[Marks \& Tegmark(2024)Marks and Tegmark]{marks2024geometry}
Samuel Marks and Max Tegmark.
\newblock The geometry of truth: Emergent linear structure in large language
  model representations of true/false datasets.
\newblock In \emph{First Conference on Language Modeling}, 2024.

\bibitem[Massey(1990)]{Mass90a}
J.~L. Massey.
\newblock Causality, feedback and directed information.
\newblock In \emph{Proc. 1990 Intl. Symp. on Info. Th. and its Applications,
  Waikiki, Hawaii, Nov. 27-30}, pp.\  1--6, 1990.

\bibitem[Meng et~al.(2022)Meng, Bau, Andonian, and Belinkov]{meng2022locating}
Kevin Meng, David Bau, Alex Andonian, and Yonatan Belinkov.
\newblock Locating and editing factual associations in {GPT}.
\newblock In \emph{Advances in Neural Information Processing Systems},
  volume~35, pp.\  17359--17372, 2022.

\bibitem[Minsky(1986)]{minsky1986society}
Marvin Minsky.
\newblock \emph{The Society of Mind}.
\newblock Simon and Schuster, 1986.

\bibitem[Nanda et~al.(2023)Nanda, Lee, and Wattenberg]{nanda2023emergent}
Neel Nanda, Andrew Lee, and Martin Wattenberg.
\newblock Emergent linear representations in world models of self-supervised
  sequence models.
\newblock In \emph{Proceedings of the 6th BlackboxNLP Workshop: Analyzing and
  Interpreting Neural Networks for NLP}, pp.\  16--30, 2023.

\bibitem[Ni et~al.(2024)Ni, Eysenbach, Seyedsalehi, Ma, Gehring, Mahajan, and
  Bacon]{ni2024bridging}
Tianwei Ni, Benjamin Eysenbach, Erfan Seyedsalehi, Michel Ma, Clement Gehring,
  Aditya Mahajan, and Pierre-Luc Bacon.
\newblock Bridging state and history representations: Understanding
  self-predictive {RL}.
\newblock In \emph{International Conference on Learning Representations}, 2024.

\bibitem[Olah(2022)]{olah2022mech}
Chris Olah.
\newblock Mechanistic interpretability, variables, and the importance of
  interpretable bases.
\newblock Transformer Circuits Thread, 2022.

\bibitem[Olsson et~al.(2022)Olsson, Elhage, Nanda, Joseph, DasSarma, Henighan,
  Mann, Askell, Bai, Chen, et~al.]{olsson2022context}
Catherine Olsson, Nelson Elhage, Neel Nanda, Nicholas Joseph, Nova DasSarma,
  Tom Henighan, Ben Mann, Amanda Askell, Yuntao Bai, Anna Chen, et~al.
\newblock In-context learning and induction heads.
\newblock \emph{arXiv preprint arXiv:2209.11895}, 2022.

\bibitem[Park et~al.(2023)Park, O'Brien, Cai, Morris, Liang, and
  Bernstein]{park2023generative}
Joon~Sung Park, Joseph O'Brien, Carrie~J. Cai, Meredith~Ringel Morris, Percy
  Liang, and Michael~S. Bernstein.
\newblock Generative agents: Interactive simulacra of human behavior.
\newblock In \emph{Proceedings of the 36th Annual ACM Symposium on User
  Interface Software and Technology}, 2023.

\bibitem[Pereira \& Zuidema(2026)Pereira and Zuidema]{pereira2026struggle}
Devin Pereira and Willem Zuidema.
\newblock Transformers struggle to use their emergent world models: Revisiting
  the tower of hanoi, and the illusion of thinking.
\newblock \emph{arXiv preprint arXiv:2608.07077}, 2026.

\bibitem[Piotrowski et~al.(2025)Piotrowski, Riechers, Filan, and
  Shai]{piotrowski2025constrained}
Mateusz Piotrowski, Paul~M Riechers, Daniel Filan, and Adam~S Shai.
\newblock Constrained belief updates explain geometric structures in
  transformer representations.
\newblock \emph{arXiv preprint arXiv:2502.01954}, 2025.

\bibitem[Rao \& Ballard(1999)Rao and Ballard]{rao1999predictive}
Rajesh P.~N. Rao and Dana~H. Ballard.
\newblock Predictive coding in the visual cortex: a functional interpretation
  of some extra-classical receptive-field effects.
\newblock \emph{Nature Neuroscience}, 2\penalty0 (1):\penalty0 79--87, 1999.

\bibitem[Ravindran \& Barto(2003)Ravindran and Barto]{ravindran2003smdp}
Balaraman Ravindran and Andrew~G. Barto.
\newblock {SMDP} homomorphisms: An algebraic approach to abstraction in
  semi-markov decision processes.
\newblock In \emph{Proceedings of the Eighteenth International Joint Conference
  on Artificial Intelligence}, pp.\  1011--1016, 2003.

\bibitem[Richens et~al.(2025)Richens, Everitt, and Abel]{richens2025general}
Jonathan Richens, Tom Everitt, and David Abel.
\newblock General agents need world models.
\newblock In Aarti Singh, Maryam Fazel, Daniel Hsu, Simon Lacoste-Julien, Felix
  Berkenkamp, Tegan Maharaj, Kiri Wagstaff, and Jerry Zhu (eds.),
  \emph{Proceedings of the 42nd International Conference on Machine Learning},
  volume 267 of \emph{Proceedings of Machine Learning Research}, pp.\
  51659--51687. PMLR, 13--19 Jul 2025.

\bibitem[Riechers et~al.(2025)Riechers, Elliott, and Shai]{riechers2025neural}
Paul~M Riechers, Thomas~J Elliott, and Adam~S Shai.
\newblock Neural networks leverage nominally quantum and post-quantum
  representations.
\newblock \emph{arXiv preprint arXiv:2507.07432}, 2025.

\bibitem[Rimsky et~al.(2024)Rimsky, Gabrieli, Schulz, Tong, Hubinger, and
  Turner]{rimsky2024steering}
Nina Rimsky, Nick Gabrieli, Julian Schulz, Meg Tong, Evan Hubinger, and
  Alexander Turner.
\newblock Steering {Llama} 2 via contrastive activation addition.
\newblock In \emph{Proceedings of the 62nd Annual Meeting of the Association
  for Computational Linguistics (Volume 1: Long Papers)}, pp.\  15504--15522,
  2024.

\bibitem[Rosas et~al.(2025)Rosas, Boyd, and Baltieri]{rosas2025ai}
Fernando Rosas, Alexander Boyd, and Manuel Baltieri.
\newblock {AI} in a vat: Fundamental limits of efficient world modelling for
  agent sandboxing and interpretability.
\newblock \emph{arXiv preprint arXiv:2504.04608}, 2025.

\bibitem[Rosas et~al.(2020)Rosas, Mediano, Jensen, Seth, Barrett,
  Carhart-Harris, and Bor]{rosas2020reconciling}
Fernando~E. Rosas, Pedro A.~M. Mediano, Henrik~J. Jensen, Anil~K. Seth, Adam~B.
  Barrett, Robin~L. Carhart-Harris, and Daniel Bor.
\newblock Reconciling emergences: An information-theoretic approach to identify
  causal emergence in multivariate data.
\newblock \emph{PLoS Computational Biology}, 16\penalty0 (12):\penalty0
  e1008289, 2020.

\bibitem[Ruebeck et~al.(2018)Ruebeck, James, Mahoney, and Crutchfield]{Rueb17a}
J.~Ruebeck, R.~G. James, John~R. Mahoney, and J.~P. Crutchfield.
\newblock Prediction and generation of binary {Markov} processes: Can a
  finite-state fox catch a markov mouse?
\newblock \emph{CHAOS}, 28:\penalty0 013109, 2018.
\newblock SFI Working Paper 2017-08-027; arxiv.org:1708.00113
  [cond-mat.stat-mech].

\bibitem[Sap et~al.(2022)Sap, Le~Bras, Fried, and Choi]{sap2022neural}
Maarten Sap, Ronan Le~Bras, Daniel Fried, and Yejin Choi.
\newblock Neural theory-of-mind? {On} the limits of social intelligence in
  large {LMs}.
\newblock In \emph{Proceedings of the 2022 Conference on Empirical Methods in
  Natural Language Processing}, pp.\  3762--3780, 2022.

\bibitem[S{\"a}rkk{\"a} \& Svensson(2023)S{\"a}rkk{\"a} and
  Svensson]{sarkka2023bayesian}
Simo S{\"a}rkk{\"a} and Lennart Svensson.
\newblock \emph{Bayesian filtering and smoothing}, volume~17.
\newblock Cambridge university press, 2023.

\bibitem[Sch{\"o}lkopf et~al.(2021)Sch{\"o}lkopf, Locatello, Bauer, Ke,
  Kalchbrenner, Goyal, and Bengio]{scholkopf2021toward}
Bernhard Sch{\"o}lkopf, Francesco Locatello, Stefan Bauer, Nan~Rosemary Ke, Nal
  Kalchbrenner, Anirudh Goyal, and Yoshua Bengio.
\newblock Toward causal representation learning.
\newblock \emph{Proceedings of the IEEE}, 109\penalty0 (5):\penalty0 612--634,
  2021.

\bibitem[Shai et~al.(2026)Shai, Amdahl-Culleton, Christensen, Bigelow, Rosas,
  Boyd, Alt, Ray, and Riechers]{shai2026transformers}
Adam Shai, Loren Amdahl-Culleton, Casper~L Christensen, Henry~R Bigelow,
  Fernando~E Rosas, Alexander~B Boyd, Eric~A Alt, Kyle~J Ray, and Paul~M
  Riechers.
\newblock Transformers learn factored representations.
\newblock \emph{arXiv preprint arXiv:2602.02385}, 2026.

\bibitem[Shai et~al.(2024)Shai, Marzen, Teixeira, Oldenziel, and
  Riechers]{shai2024transformers}
Adam~S Shai, Sarah~E Marzen, Lucas Teixeira, Alexander~G Oldenziel, and Paul~M
  Riechers.
\newblock Transformers represent belief state geometry in their residual
  stream.
\newblock \emph{Advances in Neural Information Processing Systems},
  37:\penalty0 75012--75034, 2024.

\bibitem[Shalizi \& Crutchfield(2001)Shalizi and Crutchfield]{Shal98a}
C.~R. Shalizi and J.~P. Crutchfield.
\newblock Computational mechanics: Pattern and prediction, structure and
  simplicity.
\newblock \emph{J. Stat. Phys.}, 104:\penalty0 817--879, 2001.

\bibitem[Shazeer et~al.(2017)Shazeer, Mirhoseini, Maziarz, Davis, Le, Hinton,
  and Dean]{shazeer2017outrageously}
Noam Shazeer, Azalia Mirhoseini, Krzysztof Maziarz, Andy Davis, Quoc Le,
  Geoffrey Hinton, and Jeff Dean.
\newblock Outrageously large neural networks: The sparsely-gated
  mixture-of-experts layer.
\newblock In \emph{International Conference on Learning Representations}, 2017.

\bibitem[Singh et~al.(2021)Singh, Peri, Kim, Kim, and Ahn]{singh2021structured}
Gautam Singh, Skand~Vishwanath Peri, Junghyun Kim, Hyunseok Kim, and Sungjin
  Ahn.
\newblock Structured world belief for reinforcement learning in {POMDP}.
\newblock In \emph{Proceedings of the 38th International Conference on Machine
  Learning}, volume 139 of \emph{Proceedings of Machine Learning Research},
  pp.\  9744--9755. PMLR, 2021.

\bibitem[Singh et~al.(2004)Singh, James, and Rudary]{singh2004psr}
Satinder Singh, Michael~R. James, and Matthew~R. Rudary.
\newblock Predictive state representations: A new theory for modeling dynamical
  systems.
\newblock In \emph{Proceedings of the 20th Conference on Uncertainty in
  Artificial Intelligence}, pp.\  512--519, 2004.

\bibitem[Still(2009)]{still2009information}
Susanne Still.
\newblock Information-theoretic approach to interactive learning.
\newblock \emph{EPL (Europhysics Letters)}, 85\penalty0 (2):\penalty0 28005,
  2009.

\bibitem[Subramanian et~al.(2022)Subramanian, Sinha, Seraj, and
  Mahajan]{subramanian2022approximate}
Jayakumar Subramanian, Amit Sinha, Raihan Seraj, and Aditya Mahajan.
\newblock Approximate information state for approximate planning and
  reinforcement learning in partially observed systems.
\newblock \emph{Journal of Machine Learning Research}, 23\penalty0
  (12):\penalty0 1--83, 2022.

\bibitem[Tatikonda \& Mitter(2009)Tatikonda and Mitter]{tatikonda2008capacity}
Sekhar Tatikonda and Sanjoy Mitter.
\newblock The capacity of channels with feedback.
\newblock \emph{IEEE Transactions on Information Theory}, 55\penalty0
  (1):\penalty0 323--349, 2009.

\bibitem[Templeton et~al.(2024)Templeton, Conerly, Marcus, Lindsey, Bricken,
  Chen, Pearce, Citro, Ameisen, Jones, Cunningham, Turner, McDougall,
  MacDiarmid, Tamkin, Durmus, Hume, Mosconi, Freeman, Sumers, Rees, Batson,
  Jermyn, Carter, Olah, and Henighan]{templeton2024scaling}
Adly Templeton, Tom Conerly, Jonathan Marcus, Jack Lindsey, Trenton Bricken,
  Brian Chen, Adam Pearce, Craig Citro, Emmanuel Ameisen, Andy Jones, Hoagy
  Cunningham, Nicholas~L. Turner, Callum McDougall, Monte MacDiarmid, Alex
  Tamkin, Esin Durmus, Tristan Hume, Francesco Mosconi, C.~Daniel Freeman,
  Theodore~R. Sumers, Edward Rees, Joshua Batson, Adam Jermyn, Shan Carter,
  Chris Olah, and Tom Henighan.
\newblock Scaling monosemanticity: Extracting interpretable features from
  {Claude} 3 {Sonnet}.
\newblock \emph{Transformer Circuits Thread}, 2024.

\bibitem[Teoh et~al.(2025)Teoh, Tomar, Ahn, Hu, Pearce, Sharma, Krishnamurthy,
  Islam, Lamb, and Langford]{teoh2025next}
Jayden Teoh, Manan Tomar, Kwangjun Ahn, Edward~S. Hu, Tim Pearce, Pratyusha
  Sharma, Akshay Krishnamurthy, Riashat Islam, Alex Lamb, and John Langford.
\newblock Next-latent prediction transformers learn compact world models.
\newblock \emph{arXiv preprint arXiv:2511.05963}, 2025.

\bibitem[Thomas et~al.(2018)Thomas, Bengio, Fedus, Pondard, Beaudoin,
  Larochelle, Pineau, Precup, and Bengio]{thomas2018disentangling}
Valentin Thomas, Emmanuel Bengio, William Fedus, Jules Pondard, Philippe
  Beaudoin, Hugo Larochelle, Joelle Pineau, Doina Precup, and Yoshua Bengio.
\newblock Disentangling the independently controllable factors of variation by
  interacting with the world.
\newblock \emph{arXiv preprint arXiv:1802.09484}, 2018.

\bibitem[Tishby \& Polani(2011)Tishby and Polani]{tishby2011information}
Naftali Tishby and Daniel Polani.
\newblock Information theory of decisions and actions.
\newblock In \emph{Perception-Action Cycle: Models, Architectures, and
  Hardware}, pp.\  601--636. Springer, 2011.

\bibitem[Torresan \& Baltieri(2024)Torresan and
  Baltieri]{torresan2024disentangled}
Filippo Torresan and Manuel Baltieri.
\newblock Disentangled representations for causal cognition.
\newblock \emph{Physics of Life Reviews}, 51:\penalty0 343--381, 2024.

\bibitem[Turner et~al.(2023)Turner, Thiergart, Leech, Udell, Vazquez, Mini, and
  MacDiarmid]{turner2023steering}
Alexander~Matt Turner, Lisa Thiergart, Gavin Leech, David Udell, Juan~J.
  Vazquez, Ulisse Mini, and Monte MacDiarmid.
\newblock Steering language models with activation engineering.
\newblock \emph{arXiv preprint arXiv:2308.10248}, 2023.

\bibitem[Upper(1997)]{upper1997theory}
Daniel~Ray Upper.
\newblock \emph{Theory and algorithms for hidden Markov models and generalized
  hidden Markov models}.
\newblock University of California, Berkeley, 1997.

\bibitem[Vafa et~al.(2024)Vafa, Chen, Rambachan, Kleinberg, and
  Mullainathan]{vafa2024evaluating}
Keyon Vafa, Justin~Y. Chen, Ashesh Rambachan, Jon Kleinberg, and Sendhil
  Mullainathan.
\newblock Evaluating the world model implicit in a generative model.
\newblock In \emph{Advances in Neural Information Processing Systems},
  volume~37, 2024.

\bibitem[Vig et~al.(2020)Vig, Gehrmann, Belinkov, Qian, Nevo, Singer, and
  Shieber]{vig2020investigating}
Jesse Vig, Sebastian Gehrmann, Yonatan Belinkov, Sharon Qian, Daniel Nevo,
  Yaron Singer, and Stuart Shieber.
\newblock Investigating gender bias in language models using causal mediation
  analysis.
\newblock In \emph{Advances in Neural Information Processing Systems},
  volume~33, pp.\  12388--12401, 2020.

\bibitem[Virgo et~al.(2025)Virgo, Biehl, Baltieri, and Capucci]{virgo2025good}
Nathaniel Virgo, Martin Biehl, Manuel Baltieri, and Matteo Capucci.
\newblock A “good regulator theorem” for embodied agents.
\newblock In \emph{Artificial Life Conference Proceedings 37 (ALIFE 2025)},
  pp.\ ~46. MIT Press, 2025.
\newblock arXiv:2508.06326.

\bibitem[Wang et~al.(2022)Wang, Variengien, Conmy, Shlegeris, and
  Steinhardt]{wang2022interpretability}
Kevin Wang, Alexandre Variengien, Arthur Conmy, Buck Shlegeris, and Jacob
  Steinhardt.
\newblock Interpretability in the wild: a circuit for indirect object
  identification in gpt-2 small.
\newblock \emph{arXiv preprint arXiv:2211.00593}, 2022.

\bibitem[Zhang et~al.(2019)Zhang, Lipton, Pineda, Azizzadenesheli, Anandkumar,
  Itti, Pineau, and Furlanello]{zhang2019learning}
Amy Zhang, Zachary~C. Lipton, Luis Pineda, Kamyar Azizzadenesheli, Anima
  Anandkumar, Laurent Itti, Joelle Pineau, and Tommaso Furlanello.
\newblock Learning causal state representations of partially observable
  environments.
\newblock \emph{arXiv preprint arXiv:1906.10437}, 2019.

\bibitem[Zhang \& Nanda(2024)Zhang and Nanda]{zhang2024towards}
Fred Zhang and Neel Nanda.
\newblock Towards best practices of activation patching in language models:
  Metrics and methods.
\newblock In \emph{The Twelfth International Conference on Learning
  Representations}, 2024.

\bibitem[Zhang et~al.(2024)Zhang, Wei, and Sun]{zhang2024automata}
Yihao Zhang, Zeming Wei, and Meng Sun.
\newblock Automata extraction from transformers.
\newblock \emph{arXiv preprint arXiv:2406.05564}, 2024.

\bibitem[Zhang et~al.(2026)Zhang, Luo, and Baltieri]{zhang2026compositional}
Yivan Zhang, Ziyan Luo, and Manuel Baltieri.
\newblock Compositional behavioral semantics for state abstraction in
  reinforcement learning.
\newblock In \emph{Proceedings of the 43rd International Conference on Machine
  Learning}, 2026.
\newblock Also available as arXiv:2606.25357.

\bibitem[Zhang et~al.(2022)Zhang, Yang, Liu, Tokekar, and
  Huang]{zhang2021reinforcement}
Zhi Zhang, Zhuoran Yang, Han Liu, Pratap Tokekar, and Furong Huang.
\newblock Reinforcement learning under a multi-agent predictive state
  representation model: Method and theory.
\newblock In \emph{International Conference on Learning Representations}, 2022.

\bibitem[Zhu et~al.(2024)Zhu, Zhang, and Wang]{zhu2024language}
Wentao Zhu, Zhining Zhang, and Yizhou Wang.
\newblock Language models represent beliefs of self and others.
\newblock In \emph{Forty-first International Conference on Machine Learning},
  2024.

\bibitem[Zollman(2007)]{zollman2007communication}
Kevin~JS Zollman.
\newblock The communication structure of epistemic communities.
\newblock \emph{Philosophy of science}, 74\penalty0 (5):\penalty0 574--587,
  2007.

\end{thebibliography}

\clearpage

\appendix
\crefalias{section}{appendix}
\crefalias{subsection}{subappendix}
\crefalias{subsubsection}{subsubappendix}
\crefname{section}{Appendix}{Appendices}
\Crefname{section}{Appendix}{Appendices}
\crefname{subsection}{Appendix}{Appendices}
\Crefname{subsection}{Appendix}{Appendices}
\crefname{subsubsection}{Appendix}{Appendices}
\Crefname{subsubsection}{Appendix}{Appendices}

\section{Extended Related Work}
\label{apx:related_work}

\paragraph{Predictive states and belief representations}
The object we track---the minimal sufficient statistic of the past for predicting the future---has been arrived at independently several times. Computational mechanics identifies it with the causal states of the $\epsilon$-machine \citep{Crut88a,Shal98a,Crut12a}, extended to input-driven systems as $\epsilon$-transducers \citep{Barn13a} and to generalized HMMs by \citet{upper1997theory}. Reinforcement learning arrived at the same place via belief-MDPs \citep{astrom1965optimal,kaelbling1998planning} and predictive state representations \citep{littman2001predictive,singh2004psr}, which define state by what it predicts rather than by a postulated latent. Learned analogs appear in neural predictive belief representations \citep{guo2018neural} and in world-model agents \citep{ha2018world,hafner2023dreamerv3}. We take the computational mechanics perspective here, in which the belief simplex, its dimension, and the update operator are all available in closed form, making representational claims quantifiable. \citet{rosas2025ai} use this framing to derive fundamental limits on world models for agent sandboxing and interpretability.

\paragraph{Belief geometry in trained networks}
\citet{shai2024transformers} first showed that the residual stream of a small transformer trained on next-token prediction contains a linear image of the process's mixed-state presentation, including its fractal structure; \citet{piotrowski2025constrained} and \citet{riechers2025neural} sharpen when and where this occurs. \citet{shai2026transformers} propose and test the factored world hypothesis: for a process built from independent sub-processes, or from a dependency chain of sub-processes that remain conditionally independent given the emitted tokens, the network represents the direct sum of per-factor beliefs rather than the tensor product, an exponential saving in embedding dimension. Our contribution is to extend their framework to the more general case of feedback processes. Concurrently, \citet{brenner2026predictive} find that transformers trained on constrained random walks factor their computation into a sufficient-statistic stage and a predictive-distribution stage, with the intermediate world-state representation shared across task variants.

\paragraph{Emergent world models and state tracking in sequence models}
A parallel literature asks whether sequence models trained on surface tokens build models of the generating world. \citet{li2023emergent} recover board state from an Othello-playing transformer and \citet{nanda2023emergent} show that a closely related encoding, relative to the current player, is linear; \citet{liu2023shortcuts} characterize the depth at which transformers can track automaton state. \citet{vafa2024evaluating} caution that models passing standard world-model diagnostics can nonetheless hold incoherent state, using metrics derived from the Myhill--Nerode theorem---i.e.\ from exactly the causal-equivalence relation that defines the causal states above \citep{Shal98a}. \citet{pereira2026struggle} sharpen the caution further: models can encode a geometrically faithful world model yet fail to maintain it through a reasoning trace, with performance partially recovered by injecting the earlier representation at inference time. Our setting is the complementary one: rather than inferring an unknown world model from behavior, we specify the process, compute its belief geometry and dimension exactly, and ask whether the network finds them. The emergence of beliefs we study is orthogonal to approaches to deliberately instill beliefs in transformers by architectural choices \citep{hu2025belief,teoh2025next}.

\paragraph{Probing, and the gap between decodability and use}
Linear probes have been the standard instrument for representational claims since \citet{alain2017understanding}, and their pitfalls are equally well documented: a probe's accuracy conflates what the representation encodes with what the probe itself can learn, motivating control tasks and selectivity \citep{hewitt2019designing} and, more broadly, the methodological cautions surveyed by \citet{belinkov2022probing}. \citet{elazar2021amnesic} sharpen the central objection---high decodability does not establish that the model \emph{uses} the decoded quantity---which is precisely why our probe results (\cref{fig:feedback_msp}) are insufficient on their own and why we report the intervention of \cref{fig:steering}. We adopt both practices: selectivity controls in \cref{fig:apx_probe_controls} and a causal edit measured against null, floor, and unrelated baselines.

\paragraph{Causal interventions and steering}
Our belief-patching experiment inherits the activation-patching paradigm introduced for causal mediation analysis by \citet{vig2020investigating} and developed into causal tracing and targeted weight edits by \citet{meng2022locating}; we follow the metric and baseline recommendations of \citet{zhang2024towards}. Most directly comparable is distributed alignment search \citep{geiger2024finding}, which localizes an interpretable causal variable to a learned subspace and intervenes there---analogous to our write along the encoder matrix, with the difference that our target subspace is given by theory rather than learned. The activation-steering literature \citep{turner2023steering,rimsky2024steering,marks2024geometry} pursues behavioral control through similar edits; our results suggest that where a process factorizes, the corresponding edits are naturally \emph{modular}, addressing one agent's beliefs without disturbing another's.

\paragraph{Factored, disentangled, and modular representations}
The disentanglement literature \citep{higgins2017betavae,higgins2018towards} confronts an identifiability barrier: \citet{locatello2019challenging} show that unsupervised disentanglement is impossible without inductive bias or supervision. Our known factorized generator gives a controlled setting and a theoretically derived target for testing factorization (\cref{cor: Feedback}).  Work seeking modularity without a known ground truth---differentiable weight masks \citep{csordas2021modular}, clusterability \citep{filan2021clusterability}, and structural compositionality probes \citep{lepori2023break}---is the empirical counterpart to our synthetic setup, where the factors are known in advance. Architecturally imposed modularity \citep{shazeer2017outrageously,fedus2022switch} contrasts with the emergent modularity we observe.

\paragraph{Feedback loops, cybernetics, and predictive processing}
That an adaptive system in a loop with its environment must match the variety of that environment is the founding insight of cybernetics \citep{Ashb57a,Ashb60a}, sharpened into an internal-model claim by \citet{conant1970every} and later formalized by e.g.~\citep{richens2025general, baltieri2025bayesian, virgo2025good}: every good regulator of a system must be a model of that system. Our \cref{cor: Feedback} is a formal refinement in the predictive-state setting---not merely that a model is required, but that in a feedback loop it decomposes into the environment's belief and the action-delayed agent's belief. Information-theoretic treatments of the perception--action cycle \citep{klyubin2005all,tishby2011information,still2009information} quantify the channel between agent and world; the thermodynamic costs of running such loops and of modularity itself are treated by \citet{Boyd15a,Boyd17a,fiderer2025work}, and the decomposition of monolithic world models into modular sub-transducers by \citet{boyd2025monoliths}. In neuroscience, predictive processing posits a hierarchy of predictive subunits coupled through prediction error \citep{rao1999predictive,Fris10a,clark2013whatever}, the biological instantiation of the modular-predictor picture that motivates our question about transformers.

\paragraph{Multi-agent systems and models of other agents}
\citet{minsky1986society} is the canonical statement that apparently monolithic cognition may be a society of interacting parts. In language models, \citet{andreas2022language} argues that next-token prediction induces representations of latent agents, and \citet{zhu2024language} find the beliefs of a story's protagonist and of an omniscient observer linearly decodable from attention-head activations; our contribution is to make the corresponding geometry and dimension exact rather than qualitative. Behavioral work probes whether models represent others' mental states \citep{sap2022neural,kosinski2024evaluating} and deploys LLMs as populations of interacting agents \citep{park2023generative,du2024improving}. In multi-agent RL, agents that model each other in a loop are studied as opponent modeling \citep{foerster2018learning,albrecht2018autonomous}.

\paragraph{Abstractions, coarse-graining, and complexification}
State abstractions are a central component of modern model-based RL~\citep{abel2203theory, subramanian2022approximate, ni2024bridging, zhang2026compositional} used to understand the process of coarse-graining concrete states to abstract states while preserving decision-relevant behavioral properties, with connections to bisimulation learning~\citep{givan2003equivalence, ravindran2003smdp, zhang2019learning}. Our approach here differs because we look at coarse-grainings of interfaces, rather than states.

That coarse-graining can \emph{increase} the dimension needed to predict is counterintuitive but has precedent: functions of Markov chains are generally not Markov \citep{burke1958markovian}, and partial observation of a finite HMM can produce uncountably many or fractally distributed predictive states \citep{jurgens2021divergent,jurgens2021ambiguity}. Our contribution is to identify this mechanism in feedback settings, where factored structure is lost by summing over one factor's outputs, and the empirical finding that transformers track this increased complexity (\cref{fig:coarse_graining}d). This connects to the broader question of when macroscopic descriptions are more, rather than less, informative than microscopic ones \citep{Ande72a,hoel2013quantifying,rosas2020reconciling}, and to the problem of where one agent ends and another begins \citep{krakauer2020information,friston2013life,bruineberg2022emperor}, which our framework raises but does not settle: any joint process admits \emph{some} feedback decomposition, and \cref{eq:dimensions} tells us only when a given decomposition pays for itself.

\clearpage

\section{Computational Mechanics Background}
\label{apx:Comp_Mech}

\subsection{Interfaces: Agent Behavior}

\begin{wrapfigure}{R}{0.6\textwidth}
\centering
\includegraphics[width=\linewidth]{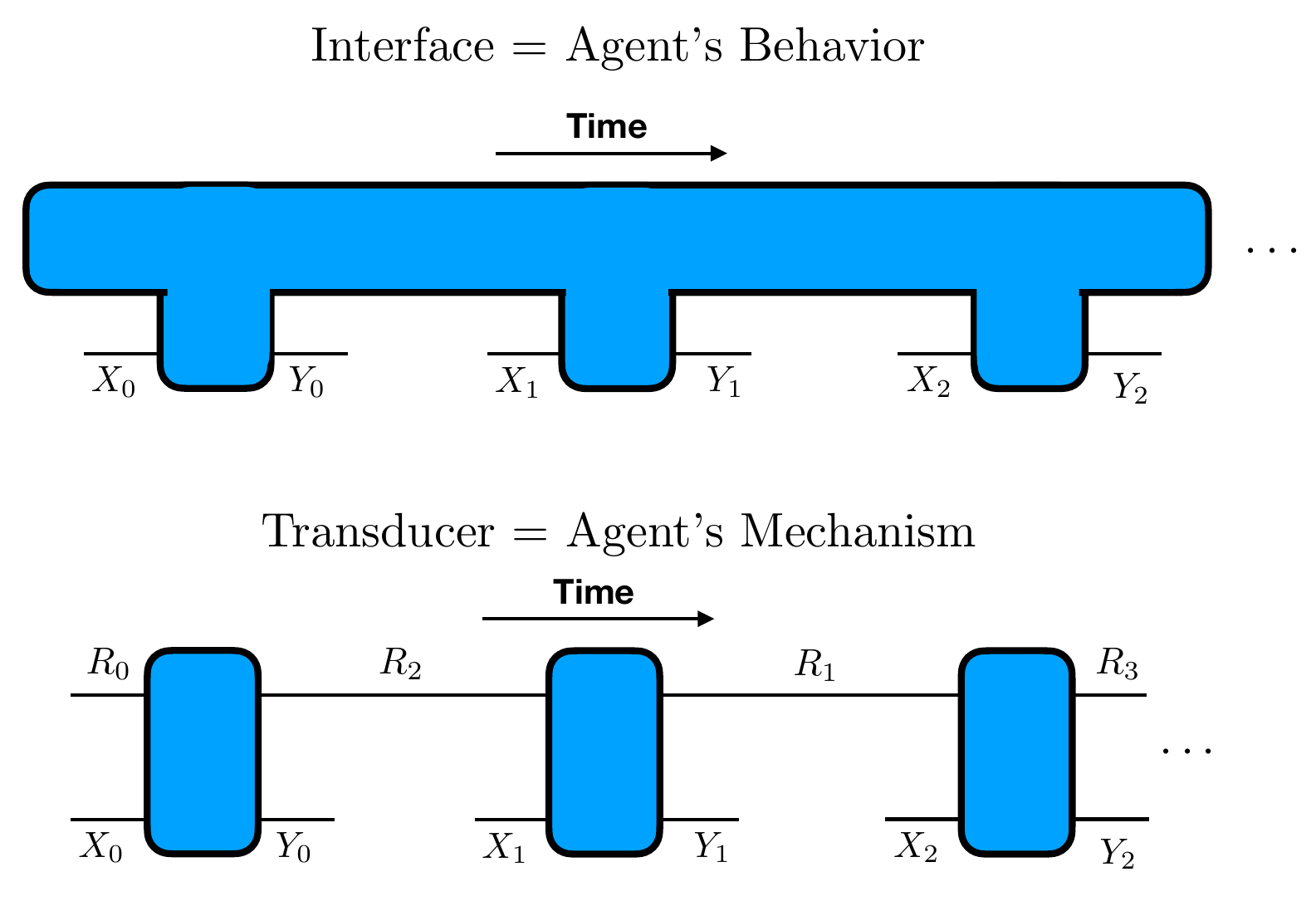}
\caption{ (TOP) An agent's behavior is described by the interface it generates, which can be thought of as a channel whose inputs and outputs are arranged in time. (BOTTOM) An agent's mechanism is described by a transducer, which describes a recursive circuit that uses a memory random variable $R_t$ to carry information between past and future.}
\label{fig:Behavior_Mechanism} 
\end{wrapfigure}

Collective intelligence arises from the interaction of individual \emph{agents}. We define an agent's ``behavior'' as what is observable from the outside: the agent's actions in response to its external stimuli. In other words, what sequence of actions $y_{0:L}\equiv y_0y_1 \cdots y_{L-1}$ will that agent take in response to the sequence of perceptions $x_{0:L}=x_0x_1 \cdots x_{L-1}$?

Formally, we characterize behavior as the stochastic mapping from the agent's sequence of perceptions $x_{0:L}$ to its actions $y_{0:L}$, otherwise known as an \emph{interface}.  Specifically, an interface is a set of conditional probabilities that maps ever-longer perception sequences to action sequences of the same length:
\begin{align*}
    I[Y|X] \equiv \{ p(Y_{0:L}|X_{0:L}) \\
    | L \in \{0,1,2 \cdots \}\},
\end{align*}
where $X_t$ and $Y_t$ are the random variables at time $t$ for the perception and action, respectively, while $x_t$ and $y_t$ are specific realizations. The probability $I[y_{0:L}|x_{0:L}]\equiv p(Y_{0:L}=y_{0:L}|X_{0:L}=x_{0:L})$ reflects the probability of performing the action sequence $y_{0:L}$ if the agent receives the specific perception sequence $x_{0:L}$ in a feed-forward context \citep{Barn13a,boyd2025monoliths}.  This interface construction is sufficiently general to encode \emph{any} stochastic strategy.

We note one constraint on these interfaces to make them realizable: they must be causal.  This means future perceptions cannot affect past actions \citep{boyd2025monoliths}, meaning that
\begin{align}
   p(Y_{0:L}=y_{0:L}|X_{0:\infty}=x_{0:\infty})= p(Y_{0:L}=y_{0:L}|X_{0:L}=x_{0:L}).
\end{align}
This is known as being non-anticipatory or anticipation-free \citep{boyd2025monoliths,Barn13a}.  In essence, the strategy an agent employs must be a function of the past.  If it could use its future perceptions to choose its actions, this would be like writing down tomorrow's lottery numbers today to obtain a vast fortune: it violates our understanding of time's arrow.  \cref{fig:Behavior_Mechanism} shows how the agent's behavior can be conceived of as a circuit in time.  The ordering of elements prevents the input future $\overrightarrow{X}_{t} \equiv X_{t} X_{t+1} \cdots$ from influencing the output past $\overleftarrow{Y}_{t} \equiv Y_0 \cdots Y_{t-1}$.

While the form of the channel remains fixed in a feedback context, the true conditional probability one observes between $X$ and $Y$ can change.  In the base case, where another agent (agent U) takes the action sequence $y_{0:L}$ of our original agent (agent T) as its own perceptions, then returns its actions $x_{1:L+1}$ to agent T, the joint sequence probability of these agents acting in sequence is given by the product of their interfaces \citep{fiderer2025work}:
\begin{align}
\Pr(X_{0:L}=x_{0:L},Y_{0:L}=y_{0:L})=\sum_{x_L} p(x_0)\, I_T[y_{0:L}|x_{0:L}]\, I_U[x_{1:L+1}|y_{0:L}],
\end{align}
where $x_0$ is drawn from an initial distribution $p(x_0)$ and the final action $x_L$ is marginalized as in \cref{eq:joint_process}.
Consider two very simple agents.  Agent T produces a $0$ or $1$ as its action with equal probability at all times, and Agent U is simply the identity channel, meaning that $x_{t+1}=y_t$.  In this case $\Pr(Y_{0:L}=y_{0:L}|X_{0:\infty}=x_{0:\infty})=\delta_{y_{0:L},x_{1:L+1}}$, meaning that the conditional probability from perceptions $X$ to actions $Y$ is \emph{anticipatory}.  It should be noted that this is not because a feedback loop somehow violates causality or time's arrow.  What looks like anticipation of future perceptions by present actions is simply those actions becoming the future perceptions. Therefore, the true conditional probability of the feedback process is not necessarily the same as the conditional probability that defines the feedback loop
\begin{align}
\Pr(Y_{0:L}=y_{0:L}|X_{0:L}=x_{0:L}) \neq I_T[y_{0:L}|x_{0:L}].
\end{align}
Equality only holds for feed-forward networks of transducers.

In summary, to examine the effect an agent has on other agents in its environment, one needs only determine its interface $I[Y|X]$.  While it won't be equivalent to the conditional probability in the case of feedback, it can be used to directly calculate the feedback process when it is combined with the interface of its feedback companion.  An agent's causal interface is its objective statistical behavior.

\subsection{Transducers: Agent Mechanism}

Alongside its perception $X_t$ and action $Y_t$ at time $t$, an agent also has an internal (latent) ``state of mind'' that reflects how it thinks and responds to its environment.  Let us represent this latent random variable as $R_t$. \cref{fig:Behavior_Mechanism} shows how the dynamics of the latent, together with the perceptions and actions, specify the mechanism of this agent.  The present perception $X_t$ and latent $R_t$ specify both the current action $Y_t$ and next latent $R_{t+1}$.  For instance, if someone thinks ``it is nighttime'' and perceives a ray of sunshine on the horizon, their next thought will likely be along the lines ``it is dawn,''  and perhaps they will act by making coffee.

The most general way to capture this mechanism is to specify a conditional probability
\begin{align}
    T_{r \rightarrow r'}^{(y|x)}= \Pr(Y_t=y,R_{t+1}=r'|X_t=x,R_t=r)
\end{align}
of the next state $r'$ and action $y$ given the present state $r$ and perception $x$.  This specifies a \emph{transducer} automaton \citep{Barn13a,boyd2025monoliths,Boyd15a} that operates on the latent space $\mathcal{R}$ with a symbol-labeled linear operator\footnote{We use the bra-ket notation of quantum mechanics in this appendix, where $|r\rangle $ represents a column vector with a $1$ in the $r$th place, and zeros everywhere else.}
\begin{align}
    \hat{T}^{(y|x)} \equiv \sum_{r, r'} T_{r \rightarrow r'}^{(y|x)} |r' \rangle \langle r|.
\end{align}

Partially Observable Markov Decision Processes (POMDPs), a standard model for agentic dynamics, are a type of transducer.  However, these POMDPs obey restrictions that limit the dependencies between perception, mind, and action \citep{rosas2025ai}.  By contrast, transducer models are fully general, because every transducer produces a causal interface through the relation
\begin{align}
I_T[y_{0:L}|x_{0:L}]= \langle 1 | \prod_{t=0}^{L-1}\hat{T}^{(y_t|x_t)} | \rho_0 \rangle_R,
\end{align}
and for every causal interface, there exists a transducer that produces that channel.  Here, the bra and ket on either side of the expression represent row and column vectors in the space of $\mathcal{R}$, i.e.
\begin{align*}
\langle 1| & \equiv \sum_{r} \langle r|
\quad \text{and} \quad | \rho_{0} \rangle_R \equiv \sum_{r}|r \rangle \Pr(R_0=r).
\end{align*}
In other words, the class of transducers fully expresses all possible agentic behaviors.

For every causal behavior, there exists a transducer model that produces that behavior and vice versa, but that doesn't mean the mapping is one-to-one.  In fact, for every interface, there are infinitely many transducer mechanisms which yield that particular perception-to-action mapping.  Each such transducer is known as a \emph{presentation} of an interface.  

\begin{wrapfigure}{R}{0.5\textwidth}
\centering
\includegraphics[width=\linewidth]{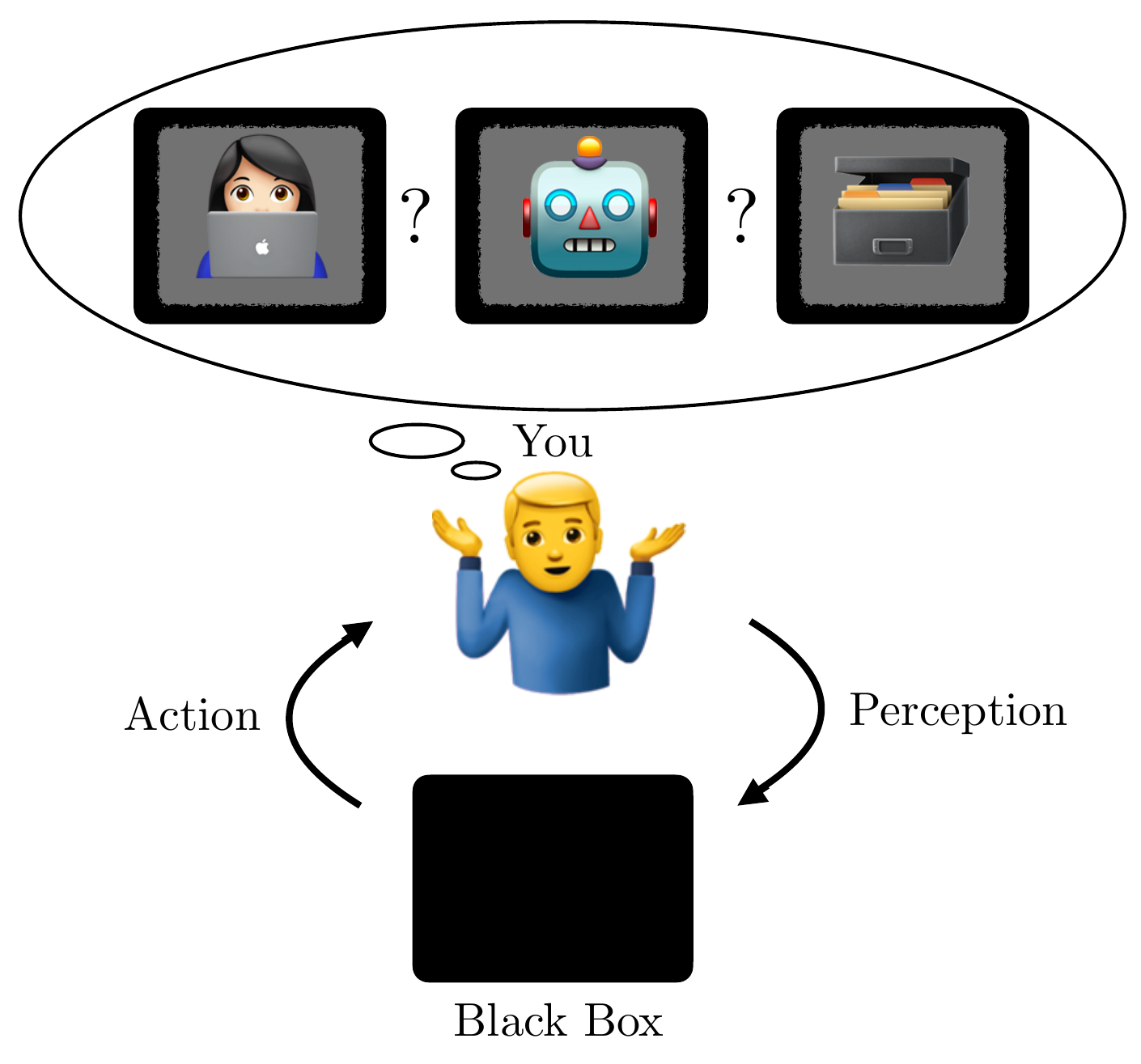}
\caption{An agent's identity may be obscured by a black box.  Even if you can perfectly characterize the way they map their perceptions to actions, you may be uncertain as to how their latent structure leads to those actions.  This is a classic manifestation of the Turing test, where many mechanisms (human, AI, lookup tables) can all produce statistically similar behavior.}
\label{fig:Black_Box} 
\vspace{-1cm}
\end{wrapfigure}

The multiplicity of transducer presentations is a reflection of the fact that there are many mechanisms to generate the same observable statistics.  Consider, for example, a help window on a website.  The user experience of the statistical behavior of that interface might be the same regardless of whether it's a person on the other side, a lookup table for responses, or an AI (\cref{fig:Black_Box}).
 But the latent ``mind'' of the agent driving those responses would be very different.  The presentation captures dynamics of the latent in parallel to the perceptions and actions: the mechanism.  These transducer presentations parallel the role of Hidden Markov Models (HMMs) in modeling a stochastic process \citep{upper1997theory, Crut01a, Shal98a}.

\newpage
\subsection{Mixed State Presentation: Universal Agent Beliefs}

The parallel between HMMs and transducers extends further to their predictive belief states.  Among the many presentations of a process or interface, predictive presentations, which use the past to anticipate future outcomes, have garnered special attention for both their theoretical implications in computational mechanics \citep{Jame11a,Boyd17a,Crut13a,Crut12a}, and their application to transformers at the frontier of AI interpretability \citep{shai2024transformers,shai2026transformers,piotrowski2025constrained}.
Specifically, it is possible to compute the predictive states of a belief-transducer by examining the Mixed State Presentation (MSP) \citep{jurgens2021shannon, jurgens2021divergent, rosas2025ai}, also known as belief states \citep{shai2026transformers,shai2024transformers}:
\begin{align}
    |\rho_{t+1} \rangle_R = \frac{\hat{T}^{(y_t|x_t)}|\rho_t \rangle_R }{\langle 1| \hat{T}^{(y_t|x_t)}|\rho_t \rangle_R}.
\end{align}
This belief state provides the distribution of latent states inferred from the action|perception history $y_{0:t}|x_{0:t}$:
\begin{align*}
    |\rho_t\rangle_R \equiv \sum_{r}|r\rangle \Pr(R_t=r|Y_{0:t}=y_{0:t},\\
    X_{0:t}=x_{0:t}).
\end{align*}

The predictive belief states linearly encode the information from the past that is necessary to predict the token.  Unless there is redundancy in the presentation, the MSP is isomorphic to the $\epsilon$-machine or $\epsilon$-transducer: unique minimal prediction presentations of the process \citep{Barn13a,Crut01a}. These objects provide predictions for the geometry of activations within neural networks trained on next-token prediction \citep{shai2024transformers,piotrowski2025constrained}.  In this way, the belief geometry reflects a sort of universal Platonic representation of the agent's state of mind.  In what follows, we will extend these results to many agents exchanging information with each other, showing that their individual belief states are components of the belief states of the whole.

\section{Beyond Feed-Forward}
\label{apx: Beyond Feed-forward}

Unlike feed-forward networks of transducers \citep{boyd2025monoliths,shai2026transformers}, rather than only using the memory space of the component processes, this construction also includes the perception space of Agent $T$ as memory.  Some additional memory is necessary to carry additional information in the feedback loop forward in time.

\begin{wrapfigure}{R}{0.5\textwidth}
\centering
\includegraphics[width=\linewidth]{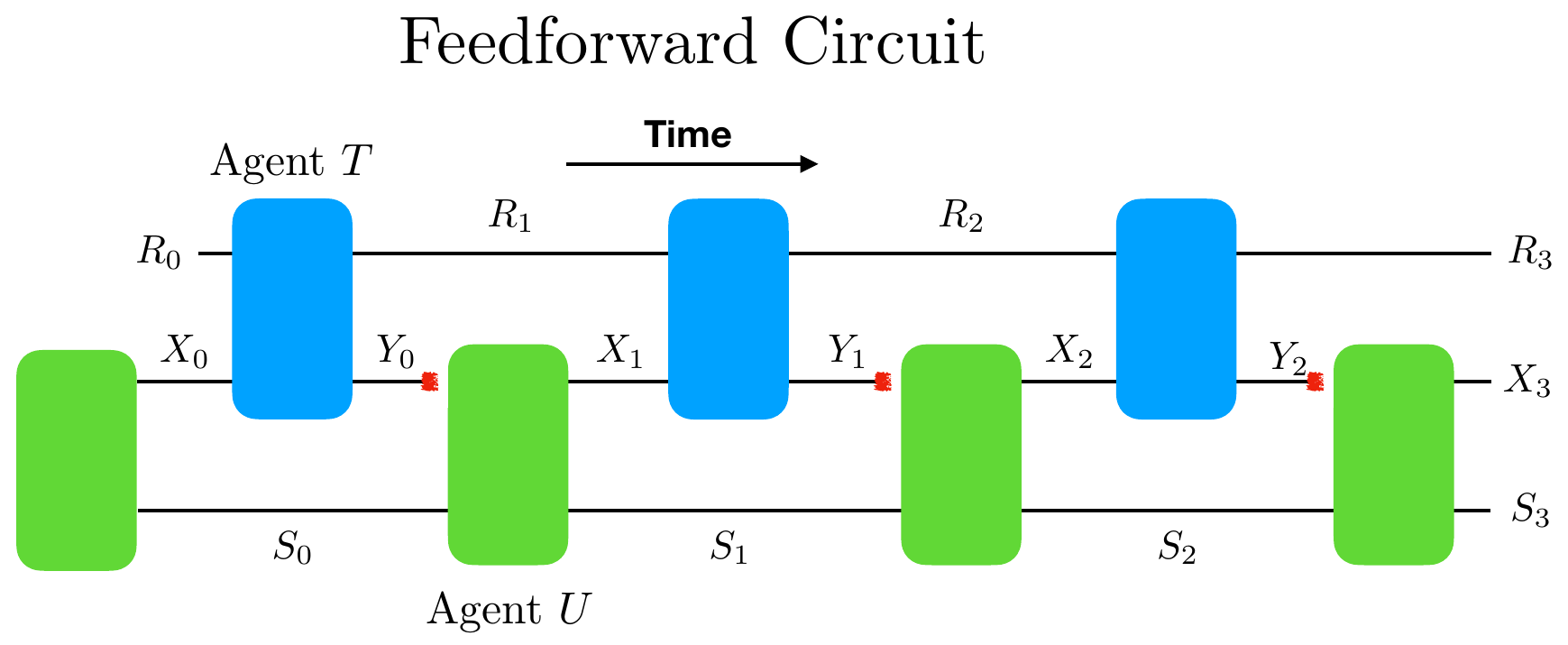}
\caption{A feed-forward circuit composed of multiple agents (bottom) is a special case of a feedback loop.  In this case, that is achieved by eliminating the dependence of Agent $U$ on Agent $T$'s action, indicated by striking the wire for $Y_t$ going to $U$ with a red slash.}
\label{fig:Feedback_Vs_Feedforward} 
\end{wrapfigure}

Consider the simple example of two memoryless agents, which simply stochastically transform their perceptions to actions
\begin{align}
T^{(y|x)} & = \Pr(Y_t=y|X_t=x)
\\ U^{(x|y)} & = \Pr(X_{t+1}=x|Y_t=y).
\end{align}
In the feed-forward case (\cref{fig:Feedback_Vs_Feedforward}), in which we break the feedback $U^{(x|y)}=U^{(x)}$, we simply obtain an IID process, which is also memoryless.  No information is carried between time steps, and the joint process can be expressed as an uncorrelated product
\begin{align}
    \Pr(X_{0:L},Y_{0:L}) = \prod_{t=0}^{L-1}U^{(x_t)}T^{(y_t|x_t)}.
\end{align}

By contrast, if we allow for feedback, then we allow for the Markovian dependency.  In fact, in the marginalized sub-processes $\Pr(X_{0:L})$ or $\Pr(Y_{0:L})$, \emph{any} Markov order-1 process can be realized by simply making $Y_t$ a copy of $X_t$.  Therefore, we are able to introduce memory, temporal correlations, and complexity into a collection of maximally simple memoryless agents by linking them into a feedback process.  In essence, \emph{complexity emerges} through feedback. This is qualitatively different than feed-forward processes, where the complexity of the whole cannot exceed the sum of its parts.  The following sections explore the complexity of feedback processes in greater detail through the dimension of their mixed states.

Feedback processes clearly go beyond the feed-forward processes, but how much further do they go?  Are there some systems that can be naturally understood as separate agents interacting while it is impossible to fully distinguish separate agents in other scenarios?  For instance, it may be fairly clear that we can distinguish each of the people in a country as a distinct agent with their own beliefs, while the country itself is a macromind that operates through the composition of the actions and perceptions of all its members.  On the other hand, can we model every individual cell in our body as an individual agent?

Feedforward processes provide a firm binary, identifying which sort of probabilistic dependencies allow for an interpretation as a feedforward network of agents.  This boils down to whether each agent is anticipation-free from its actions to its perceptions \citep{Barn13a}.  However, no such binary appears to exist in a network of agents linked through feedback.  This is because, as shown in \cref{app:Decomposing Joint Processes}, \emph{any} joint process can be decomposed into a feedback process.  Therefore, the results described in what follows are relevant to \emph{any} joint process.  There will always be a way to interpret the system as if it were two or more agents interacting. Whether such an interpretation is \emph{useful}, however, is a quantitative question which depends on the complexity of the individual factors.
\clearpage

\section{Joint Process for a Feedback Loop}
\label{app:Joint Process for a Feedback Loop}

Here we prove \cref{theo:linear_HMM}, which relates the linear operators of the individual transducers in a feedback loop to the linear operator for an HMM that produces the joint process.

\begin{figure}
\centering
\includegraphics[width=\linewidth]{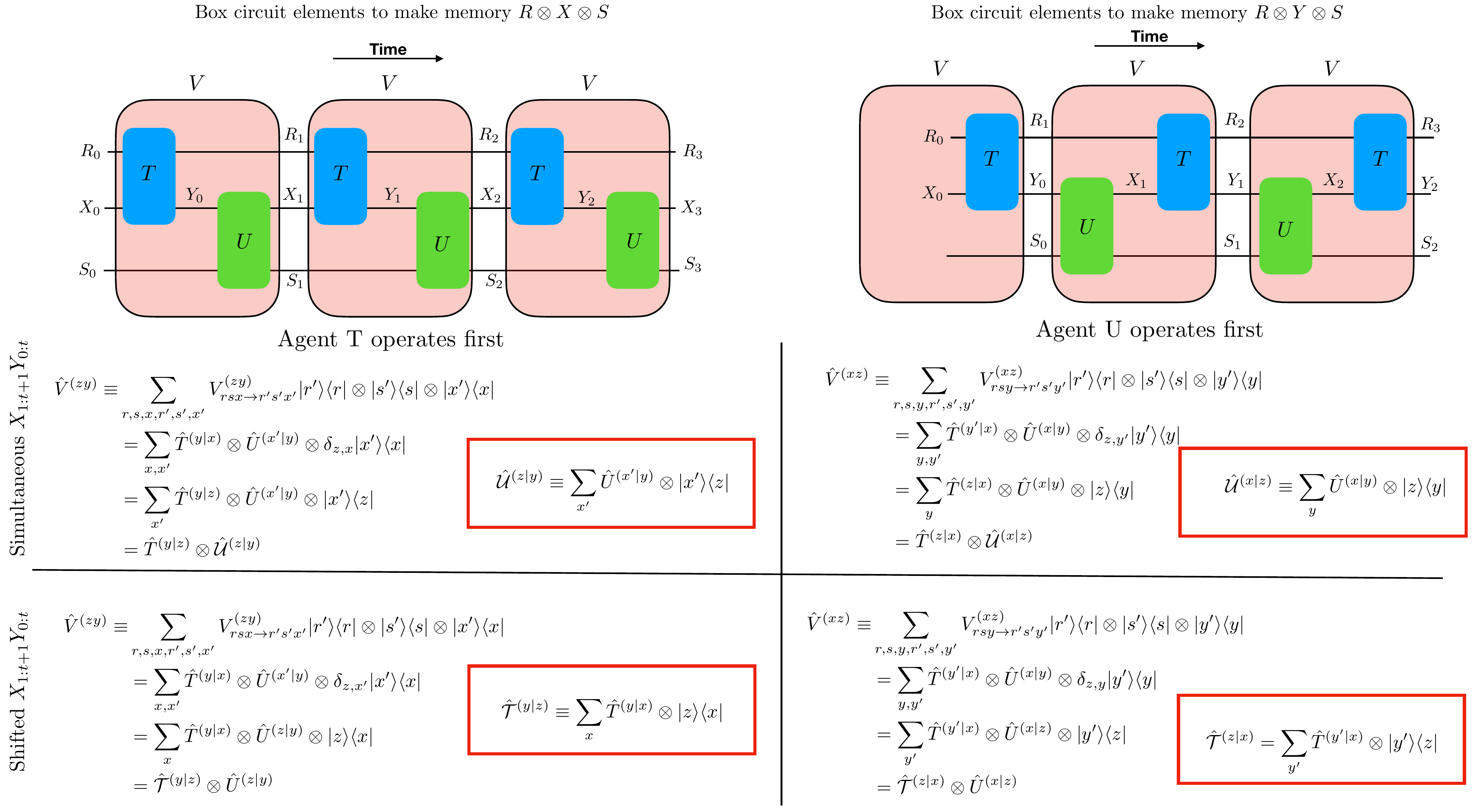}
\caption{There are four different readily accessible presentations for a feedback process between agents $T$ and $U$.  Which one you arrive at depends on $1)$ whether you choose to model the simultaneous process $X_{0:t}Y_{0:t}$ or the shifted process $X_{1:t+1}Y_{0:t}$, and $2)$ whether you box circuit elements so that Agent $U$ or Agent $T$ operates first.  All presentations modify the linear operator of one of the transducers, either delaying the action or the perception.  }
\label{fig:Different_Presentations} 
\end{figure}

\begin{proof}[Proof of \cref{theo:linear_HMM}]
From the circuit shown in \cref{fig:Feedback_Stages}, we can construct the joint process over observables $X$ and $Y$, and latents $R$ and $S$ by applying the transducers for agent $T$ and agent $U$.  They can be thought of as symbol-labeled transition matrices, with elements given by the conditional probabilities:
\begin{align}
    T^{(y|x)}_{r \rightarrow r'} & = \Pr(Y_t=y,R_{t+1}=r'|X_t=x,R_t=r)
    \\ U^{(x|y)}_{s \rightarrow s'} & = \Pr(X_{t+1}=x,S_{t+1}=s'|Y_t=y,S_t=s).
\end{align}

Given the perception-action loop, we must decide how to start.  We might either choose Agent $T$ or Agent $U$ to go first.  Based on our indexing, we choose to start with the initial joint probability $p_0(r,s,x) \equiv \Pr(R_0=r,S_0=s,X_0=x)$ of Agent $T$'s memory $R$, Agent $U$'s memory $S$, and Agent $U$'s action $X$.  By applying agent $T$'s transducer and agent $U$'s transducer repeatedly, we obtain the joint process over all variables 
\begin{gather}
 \Pr(R_{0:L+1}=r_{0:L+1},S_{0:L+1}=s_{0:L+1},X_{0:L}=x_{0:L},Y_{0:L}=y_{0:L})\nonumber
\\= \sum_{x_L} p_0(r_0,s_0,x_0) \prod_{t=0}^{L-1}T^{(y_t|x_t)}_{r_t \rightarrow r_{t+1}}U^{(x_{t+1}|y_t)}_{s_t \rightarrow s_{t+1}}.
\end{gather}
After $L$ rounds of the loop, agent $U$ has produced one more action, $x_L$, than agent $T$ has consumed; it is summed out so that both observables run over the same $L$ time steps. The combination of the transducers $T$ and $U$ provides a Markov kernel for generating the process over both observables ($X$ and $Y$) and latents ($R$ and $S$).  

However, as an external observer of the feedback process, the latents are exactly that, hidden from view.  The external observer only sees the joint process over actions and perceptions $\Pr(X_{0:L},Y_{0:L})$.  

We need only marginalize the latent trajectories in $R$ and $S$ to produce the joint process $\Pr(X_{0:L},Y_{0:L})$
\begin{gather}
\Pr(X_{0:L}=x_{0:L},Y_{0:L}=y_{0:L}) \nonumber \\
 = \sum_{s_{0:L+1},r_{0:L+1}} \Pr(R_{0:L+1}=r_{0:L+1},S_{0:L+1}=s_{0:L+1},X_{0:L}=x_{0:L},Y_{0:L}=y_{0:L}) \nonumber
\\ = \sum_{x_L,s_{0:L+1},r_{0:L+1}}p_0(r_0,s_0,x_0) \prod_{t=0}^{L-1}T^{(y_t|x_t)}_{r_t \rightarrow r_{t+1}}U^{(x_{t+1}|y_t)}_{s_t \rightarrow s_{t+1}},
\end{gather}
This reveals the architecture of an HMM that uses the joint latent space $\mathcal{R} \otimes \mathcal{S} \otimes \mathcal{Z}$, where $\mathcal{Z}$ is a copy of the action space $\mathcal{X}$.  Specifically, we introduce a new HMM for the joint process with elements
\begin{align}
    V^{(xy)}_{rsz \rightarrow r's'z'}= U^{(z'|y)}_{s \rightarrow s'}T^{(y|z)}_{r \rightarrow r'}\delta_{z,x}.
    \label{eq:Element_Product}
\end{align}
This arises by ``boxing'' the $T$ and $U$ circuit elements into a single $V$ element as shown in \cref{fig:Different_Presentations}.  $R_t$, $S_t$, and $X_t$ all carry information between elements, making them the new memory basis.  It's also worth noting that (due to indexing) we copy $X_t$ to the output of $V$ using the delta function $\delta_{z,x}$. 

We can reinterpret the linear operator for $V$ as a tensor product, given the equation above.  We begin by using the standard notation for an HMM linear operator
\begin{align}
    \hat{V}^{(xy)} \equiv \sum_{r,r' \in \mathcal{R},s,s' \in \mathcal{S},z,z' \in \mathcal{X}} |r' \rangle \langle r| \otimes | s' \rangle \langle s| \otimes  |z' \rangle \langle z| V^{(xy)}_{rsz \rightarrow r's'z'} .
\end{align}
Then we apply the decomposition and find
\begin{align}
    \hat{V}^{(xy)} & \equiv \sum_{r,r' \in \mathcal{R},s,s' \in \mathcal{S},z,z' \in \mathcal{X}} |r' \rangle \langle r| \otimes | s' \rangle \langle s| \otimes  |z' \rangle \langle z| U^{(z'|y)}_{s \rightarrow s'}T^{(y|z)}_{r \rightarrow r'}\delta_{z,x} 
    \\ & = \sum_{z,z' \in \mathcal{X}} \left( \sum_{r,r' \in \mathcal{R}} T^{(y|z)}_{r \rightarrow r'} |r' \rangle \langle r| \right) \otimes \left( \sum_{s,s' \in \mathcal{S}} U^{(z'|y)}_{s \rightarrow s'} |s' \rangle \langle s|\right) \otimes \delta_{z,x}|z' \rangle \langle z|
    \\ & = \sum_{z,z' \in \mathcal{X}} \hat{T}^{(y|z)} \otimes \hat{U}^{(z'|y)}\delta_{z,x} \otimes |z' \rangle \langle z|
    \\ & = \sum_{z' \in \mathcal{X}} \hat{T}^{(y|x)} \otimes \hat{U}^{(z'|y)}\otimes |z' \rangle \langle x|
     \\ & =  \hat{T}^{(y|x)} \otimes \sum_{z' \in \mathcal{X}} \hat{U}^{(z'|y)}\otimes |z' \rangle \langle x|.
\end{align}
We introduce a new operator on the joint space of $U$'s actions and memory
\begin{align}
    \hat{\mathcal{U}}^{(x|y)} \equiv \sum_{z' \in \mathcal{X}} \hat{U}^{(z'|y)}\otimes |z' \rangle \langle x|,
\end{align}
which is the action-delayed agent, because it returns the probability of $U$'s last action.  The joint HMM linear operator is the tensor product of agent $T$'s linear operator with action-delayed $U$'s linear operator:
\begin{align}
\hat{V}^{(xy)} = \hat{T}^{(y|x)}\otimes \hat{\mathcal{U}}^{(x|y)}.
\end{align}
\end{proof}

Thus, the two factors are updated separately, each by its own operator. The process generated by $V$ can therefore be presented by this direct sum of factors, with next-token probabilities given by the bilinear equation
\begin{align}
p(xy \,|\, \bm{\eta}_t) =
\langle 1 |_{R}\, \hat{T}^{(y|x)} | \bm{\eta}_t \rangle_{R}\;
\langle 1 |_{S,X}\, \hat{\mathcal{U}}^{(x|y)} | \bm{\eta}_t \rangle_{S,X}.
\end{align}

It should be noted that this particular decomposition is the result of broken symmetries in how we analyze the joint process.  For instance, we might as easily try to predict $X_{t+1}Y_t$ from its past as $X_tY_t$.  This corresponds to predicting the shifted process $\Pr(X_{1:t+1},Y_{0:t})$ rather than the simultaneous process $\Pr(X_{0:t},Y_{0:t})$.  Additionally, in boxing the joint feedback process, we can choose whether Agent $T$ or Agent $U$ operates first.  This leads to four different valid presentations, as shown in \cref{fig:Different_Presentations}.

\section{Belief State Updating Retains the Tensor Product Structure}
\label{apx:proof_tensorproduct}

Here we prove \cref{cor: Feedback}, i.e.\ that the belief state update rule $|\rho_{t+1} \rangle=\frac{\hat{V}^{(x_ty_t)}|\rho_t \rangle}{\langle 1| \hat{V}^{(x_ty_t)}|\rho_t \rangle}$ preserves product structure of the belief states.

\begin{proof}[Proof of \cref{cor: Feedback}]
\begin{align}
&|\rho_{t+1} \rangle_{R,S,X} 
\\& \nonumber = \frac{\hat{V}^{(x_ty_t)}|\rho_t \rangle_{R,S,X}}{\langle 1|_{R,S,X} \hat{V}^{(x_ty_t)}|\rho_t \rangle_{R,S,X}}
\\ & \nonumber = \frac{\left( \hat{T}^{(y|x)} \otimes \hat{\mathcal{U}}^{(x|y)} \right)\left( |\rho_t \rangle_{R} \otimes |\rho_t \rangle_{S,X}  \right)}{\left(\langle 1|_{R} \otimes \langle 1|_{S,X}  \right)\left( \hat{T}^{(y|x)} \otimes \hat{\mathcal{U}}^{(x|y)} \right)\left( |\rho_t \rangle_{R} \otimes |\rho_t \rangle_{S,X}  \right)}
\\ & \nonumber  = \frac{ \hat{T}^{(y|x)}|\rho_t \rangle_{R} \otimes \hat{\mathcal{U}}^{(x|y)}|\rho_t \rangle_{S,X}   }{\langle 1|_{R} \hat{T}^{(y|x)}|\rho_t \rangle_{R} \langle 1|_{S,X} \hat{\mathcal{U}}^{(x|y)}|\rho_t \rangle_{S,X}}
\\ & \nonumber  = \frac{ \hat{T}^{(y|x)}|\rho_t \rangle_{R} }{\langle 1|_{R} \hat{T}^{(y|x)}|\rho_t \rangle_{R}} \otimes \frac{ \hat{\mathcal{U}}^{(x|y)}|\rho_t \rangle_{S,X}   }{ \langle 1|_{S,X} \hat{\mathcal{U}}^{(x|y)}|\rho_t \rangle_{S,X}}
\\ & \nonumber = |\rho_{t+1} \rangle_{R} \otimes |\rho_{t+1} \rangle_{S,X}.
\end{align}
Therefore, the belief state remains in a product state
\begin{align}
    |\rho_{t} \rangle_{R,S,X}= |\rho_{t} \rangle_{R} \otimes |\rho_{t} \rangle_{S,X} \Rightarrow |\rho_{t+1} \rangle_{R,S,X}= |\rho_{t+1} \rangle_{R} \otimes |\rho_{t+1} \rangle_{S,X}.
\end{align}
\end{proof}

\section{Factored Social Network Operator}
\label{app:Factored Social Network}

The linear operator for a social network can be expressed as a tensor product of the network hub and its many agent factors, which yields the factored dimension in \cref{eq:factored_social_network}

\begin{align}
\hat{V}^{(\vec{x}\vec{y})}=\hat{H}^{(\vec{y}|\vec{x})} \otimes \left( \bigotimes_{i=0}^{N-1}\hat{\mathcal{A}}(i)^{(x(i)|y(i))} \right).
\end{align}

\begin{proof}
From Theorem \ref{theo:linear_HMM}, the feedback loop HMM in a social network can be expressed as
\begin{align}
    \hat{V}^{(\vec{x}\vec{y})}=\hat{H}^{(\vec{y}|\vec{x})} \otimes \hat{\vec{\mathcal{A}}}^{(\vec{x}|\vec{y})},
\end{align}
where $\hat{H}^{(\vec{y}|\vec{x})}$ is the social network hub and the action-delayed macro-agent is
\begin{align}
    \hat{\vec{\mathcal{A}}}^{(\vec{x}|\vec{y})} \equiv \sum_{\vec{z} \in \vec{\mathcal{X}}} \hat{\vec{A}}^{(\vec{z}|\vec{y})} \otimes | \vec{z} \rangle \langle \vec{x}|.
\end{align}
The macro-agent is composed of conditionally independent sub-agents, so it can be expressed as a tensor product
\begin{align}
\hat{\vec{A}}^{(\vec{x}|\vec{y})} = \bigotimes_{i=0}^{N-1}\hat{A}(i)^{(x(i)|y(i))}.
\end{align}
This, in turn, yields an expression for the action-delayed macro-agent as a tensor product of the action-delayed sub-agents:
\begin{align}
    \hat{\vec{\mathcal{A}}}^{(\vec{x}|\vec{y})} = \bigotimes_{i=0}^{N-1} \hat{\mathcal{A}}(i)^{(x(i)|y(i))},
\end{align}
where the action-delayed sub-agent is expressed as
\begin{align}
    \hat{\mathcal{A}}(i)^{(x(i)|y(i))}= \sum_{z(i) \in \mathcal{X}(i)} \hat{A}(i)^{(z(i)|y(i))} \otimes | z(i) \rangle \langle x(i)|.
\end{align}
\end{proof}

\section{Decomposing Joint Processes into Feedback Agents}
\label{app:Decomposing Joint Processes}

Any joint process can be decomposed into two transducers interacting in a feedback loop (\cref{theo:feedback_universality}).

\begin{proof}[Proof of \cref{theo:feedback_universality}] To illustrate this, let us consider a joint process
\begin{align}
\Pr(X_{0:L},Y_{0:L}).
\end{align}
We can build this up from past to future using the probability of the present conditioned on the past:
\begin{align}
    \Pr(X_{0:L},Y_{0:L})= \prod_{t=0}^{L-1} \Pr(X_t,Y_t|X_{0:t},Y_{0:t}).
\end{align}
Each of these conditional probabilities can be decomposed using Bayes' rule:
\begin{align}
    \Pr(X_t,Y_t|X_{0:t},Y_{0:t})& = \Pr(X_t|Y_t,X_{0:t},Y_{0:t}) \Pr(Y_t|X_{0:t},Y_{0:t}).
\end{align}
Here we have indicated, through conditioning, that $Y_t$ comes before $X_t$ in the feedback loop.  This is an arbitrary choice, and this proof can proceed as easily if the alternate time-indexing is used, where $X_t$ precedes $Y_t$.

The decomposition of this process into an update of next $Y$ then next $X$ allows us to identify kernels that build the process step-by-step
\begin{align}
    k(X_t|X_{0:t},Y_{0:t+1}) & \equiv \Pr(X_t|Y_t,X_{0:t},Y_{0:t})
    \\ m(Y_t|X_{0:t},Y_{0:t}) & \equiv \Pr(Y_t|X_{0:t},Y_{0:t}). 
\end{align}
Each of these kernels can be used to generate an interface:
\begin{align}
k(X_{0:L}|Y_{0:L}) & \equiv \prod_{t=0}^{L-1} k(X_t|X_{0:t},Y_{0:t+1})
\\ m(Y_{0:L}|X_{0:L}) & \equiv \prod_{t=0}^{L-1}m(Y_t|X_{0:t},Y_{0:t}),
\end{align}
Note that while the interfaces describe a stochastic channel from $X$ to $Y$ or vice versa, they are not necessarily equal to the true conditional probability between the two sequences:
\begin{align}
k(X_{0:L}|Y_{0:L})& \neq \Pr(X_{0:L}|Y_{0:L})
\\ m(Y_{0:L}|X_{0:L}) & \neq \Pr(Y_{0:L}|X_{0:L}).
\end{align}  
Instead, their product is the joint process:
\begin{align}
\Pr(X_{0:L},Y_{0:L})=k(X_{0:L}|Y_{0:L})m(Y_{0:L}|X_{0:L}).
\end{align}
This factored form has been examined before in the context of feedback \citep{Mass90a,tatikonda2008capacity}. Our contribution stems from recognizing that \emph{both} interfaces can be realized by transducers linked into this feedback loop.

If these interfaces are causal (non-anticipatory) \citep{boyd2025monoliths}, then this is precisely the form of the joint probability when two agents interact in a feedback loop, otherwise known as a perception-action loop in \citet{fiderer2025work}.  In this case, there exist transducer presentations for both interfaces, and they can produce the observed joint process if linked into a feedback loop.

To complete the proof, we need simply verify that these interfaces are non-anticipatory.  This means that future inputs cannot influence past outputs
\begin{align}
k(X_\text{past}|Y_\text{past},Y_\text{future}) & =k(X_\text{past}|Y_\text{past})
\\ m(Y_\text{past}|X_\text{past},X_\text{future}) & =m(Y_\text{past}|X_\text{past}).
\end{align}
The past/future partitioning is asynchronous in a feedback loop.  In this case, the past inputs for $k(X_{0:t}|\cdot)$ are $Y_{0:t}$, because $Y_{t-1}$ immediately precedes $X_{t-1}$, while the past inputs for $m(Y_{0:t}|\cdot)$ are $X_{0:t-1}$, because $X_{t-2}$ immediately precedes $Y_{t-1}$.  Therefore, causality in this case reduces to two relations for all $t$:
\begin{align}
k(X_{0:t}|Y_{0:L})& =k(X_{0:t}|Y_{0:t})
\\ m(Y_{0:t}|X_{0:L})& =m(Y_{0:t}|X_{0:t-1}).
\end{align}  
Finally, we can employ marginalization to validate the above terms:
\begin{align}
k(X_{0:t}|Y_{0:L}) &  \equiv \sum_{X_{t:L}} k(X_{0:L}|Y_{0:L}) 
\\ & = \sum_{X_{t:L}}  \prod_{i=0}^{L-1} k(X_i|X_{0:i},Y_{0:i+1})
\\ & = \left(\prod_{i=t}^{L-1} \sum_{X_i} k(X_i|X_{0:i},Y_{0:i+1}) \right)\left(\prod_{i=0}^{t-1} k(X_i|X_{0:i},Y_{0:i+1}) \right)
\\ & = k(X_{0:t}|Y_{0:t})
\\ m(Y_{0:t}|X_{0:L}) & \equiv \sum_{Y_{t:L}}m(Y_{0:L}|X_{0:L})
\\ & = \sum_{Y_{t:L}}  \prod_{i=0}^{L-1}m(Y_i|X_{0:i},Y_{0:i})
\\ & = \left( \prod_{i=t}^{L-1} \sum_{Y_i} m(Y_i|X_{0:i},Y_{0:i}) \right) \left( \prod_{i=0}^{t-1}m(Y_i|X_{0:i},Y_{0:i}) \right)
\\  & = m(Y_{0:t}|X_{0:t-1}).
\end{align}
Thus, the interfaces $k$ and $m$ are non-anticipatory, and can each be generated by a transducer.  Therefore, the joint process can be generated by those two respective transducers in a feedback loop.
\end{proof}

\section{Additional Evidence for Belief Geometry}
\label{apx:role_swap}

\begin{figure}
\centering
\includegraphics[width=\textwidth]{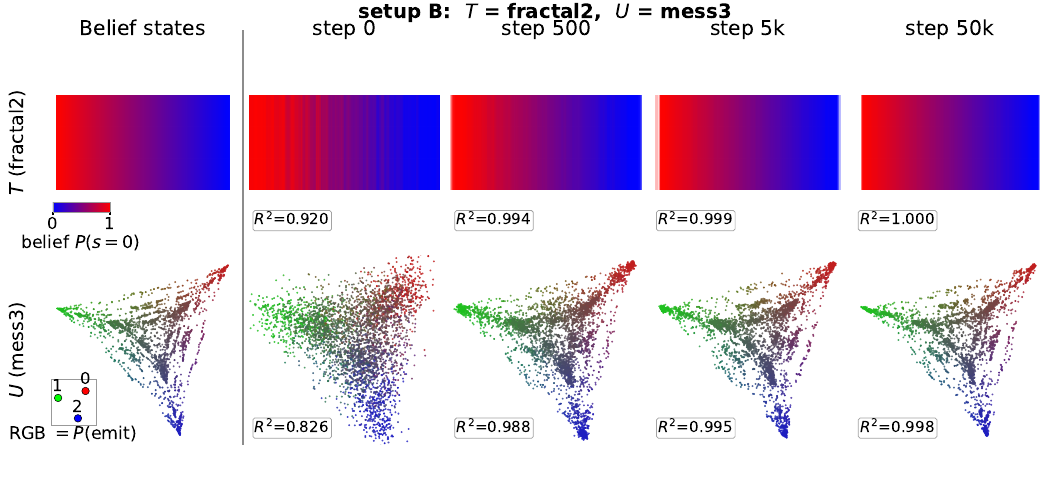}
\caption{\textbf{Belief-state geometry of the role-swapped setup likewise refines over training.} Rows: both factors of setup B (agent $T$ fractal2, agent $U$ mess3); setup A is shown in \cref{fig:feedback_msp}. Leftmost column: ground-truth MSP. Remaining columns: linear reconstruction from the final-layer residual stream at checkpoints $0$, $500$, $5$k, and $50$k (probe refit per checkpoint; same PCA basis and color conventions as \cref{fig:feedback_msp}; held-out $R^2$ per panel). Both factors are already reconstructed to $R^2 \geq 0.83$ at initialization and are refined by training to $R^2 \geq 0.998$ by step $50$k.}
\label{fig:feedback_msp_timeline_B}
\end{figure}

\begin{figure}
\centering
\includegraphics[width=\textwidth]{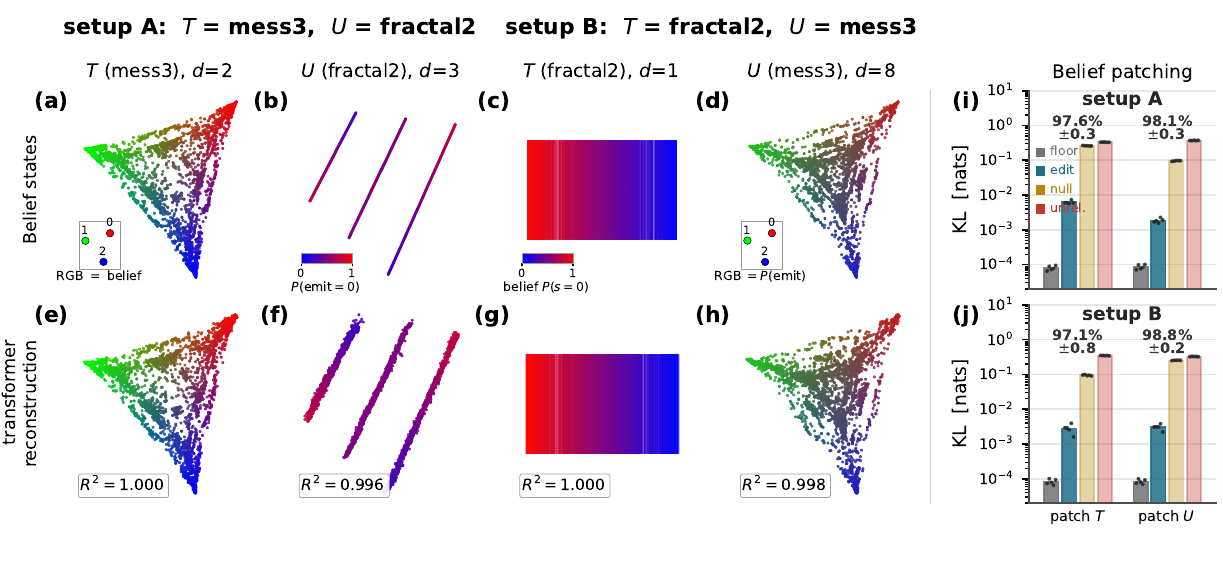}
\caption{\textbf{Both mess3--fractal2 setups are represented by factorized, causally load-bearing beliefs in the residual stream.} The final-checkpoint
reconstruction of \cref{fig:feedback_msp} with the role-swapped arrangement
alongside it: mess3 as agent $T$ and fractal2 as agent $U$ (setup A, panels a,b,e,f---
the setup reproduced in the main text) or vice-versa (setup B, panels
c,d,g,h). Top row: Bayes-optimal belief states; bottom row: reconstruction from the
final-layer residual stream at step $50$k. Panel and color conventions and the
probe protocol are as in \cref{fig:feedback_msp}; the belief patching in
\textbf{(i,j)} is as in \cref{fig:steering}.}
\label{fig:apx_role_swap}
\end{figure}

\cref{fig:feedback_msp} shows a feedback loop between a mess3 transducer as agent $T$ and a fractal2 transducer as $U$. To form a predictive state of this process, the latent state of the succeeding factor $U$ is augmented by the additional time-latched action $X$.
What happens when the two factors are swapped? The token sequence that this process describes is still the same as before, with the crucial difference that the tuples of tokens which are predicted together for the same timestep are now offset by one. For this reason, now the action of the mess3 process needs to be latched for the prediction of the next token tuple, which extends the latent space of the mess3 factor. In both the original setup (setup A) and the role-swapped setup (setup B), the beliefs in the residual stream are refined over training (see \cref{fig:feedback_msp,fig:feedback_msp_timeline_B}) to $R^2 \geq 0.996$, and patching either factor closes $97$--$99\%$ of the null-to-target distance (see \cref{fig:apx_role_swap}).

\begin{figure}
\centering
\includegraphics[width=\textwidth]{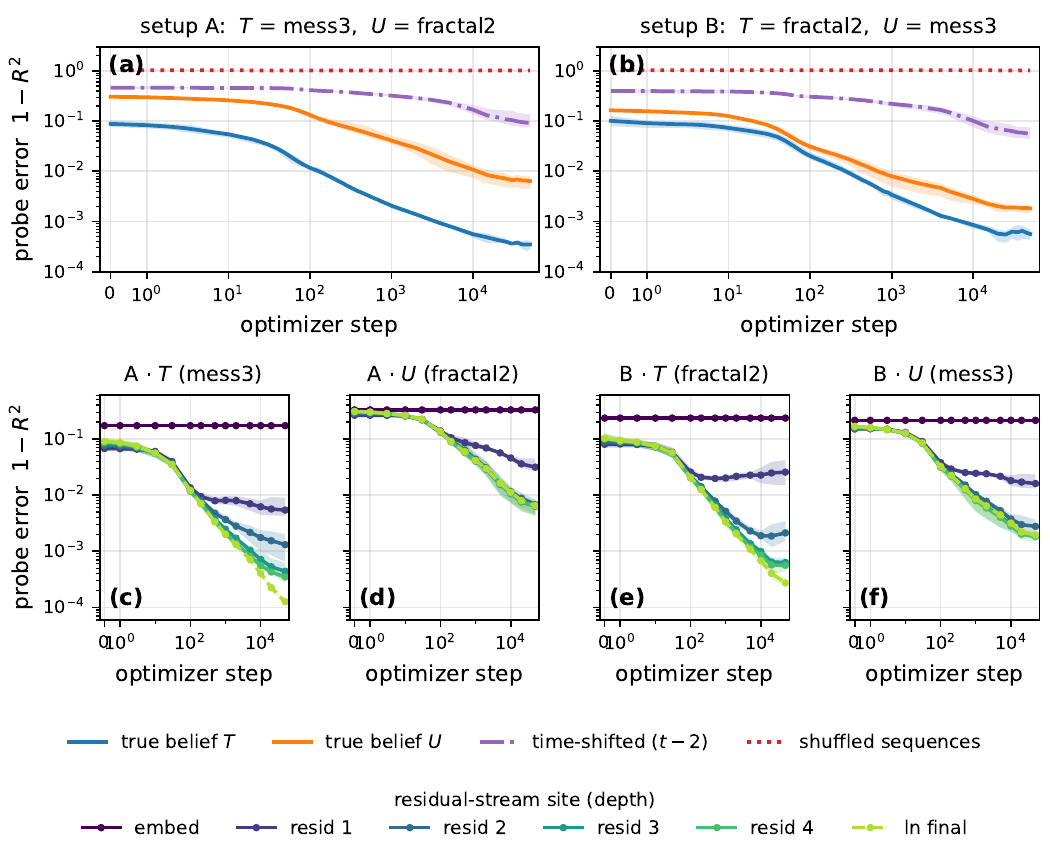}
\caption{\textbf{Probe selectivity controls and layer-wise emergence.} All
panels show the held-out linear-probe error $1-R^2$ on the residual stream
over training (mess3--fractal2 feedback loops). Solid curves: mean over
five seeds per setup---the canonical run plus four reruns
(\cref{tab:transformer_params}); shaded bands: min--max across seeds. \textbf{(a,b)} Final-layer probes
fitted to the true per-factor beliefs and to two controls: the same
sequence's belief two positions earlier (time-shifted) and beliefs of
shuffled sequences (a derangement of the batch). The shuffled floor stays
at $1-R^2 \approx 1$ throughout; the true factors improve by
$40$--$280\times$ over training.
\textbf{(c--f)} The true-belief probe refit at each residual-stream depth
(embedding, blocks $1$--$4$, final LayerNorm). Blocks $1$--$2$ acquire
essentially all of the structure; the action-delayed factor $U$ is complete
by block~2, while factor $T$ sharpens to the last layer. The embedding row
is training-invariant by construction (a linear probe on it can realize any
function of token and position).}
\label{fig:apx_probe_controls}
\end{figure}

We performed two additional controls to verify these results. First, we show that the probe error of the true belief is significantly lower than a shuffled or a time-shifted control over all steps of training (see \cref{fig:apx_probe_controls}a,b). In addition, we tested probing from different layers to find that later network layers have lower probe errors in our experiments (see \cref{fig:apx_probe_controls}c-f), justifying our choice of analyzing the last layer in our main figures.

\section{The fractal2 Telephone Process}
\label{apx:tel_cev_f2}

\begin{figure}
\centering
\includegraphics[width=0.4\textwidth]{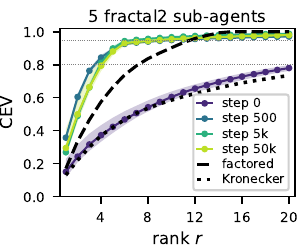}
\caption{\textbf{Cumulative explained variance for the five-agent fractal2
telephone process.} Same protocol and conventions as
\cref{fig:telephone_cev}b: CEV of the final-layer residual stream by rank at
several training steps; solid curves the mean over five seeds, shaded bands
the min--max across them; dashed the CEV of the factored concatenation
of the Bayes-optimal belief states, dotted that of their Kronecker product.}
\label{fig:apx_tel_cev_f2}
\end{figure}

\cref{fig:telephone_cev}b reports the cumulative explained variance for the three-agent mess3 telephone process. \cref{fig:apx_tel_cev_f2} is the same
measurement for five fractal2 sub-agents. Here the residual stream is \emph{lower}-dimensional than even the factored reference: fractal2's mixed-state presentation occupies lower-dimensional sub-manifolds of its probability simplex (\cref{fig:apx_role_swap}b), so the factored concatenation is
itself a loose bound on what the network has to represent.

\section{Effective Rank Across Model Widths}
\label{apx:width_sweep}

\begin{figure}
\centering
\includegraphics[width=0.72\textwidth]{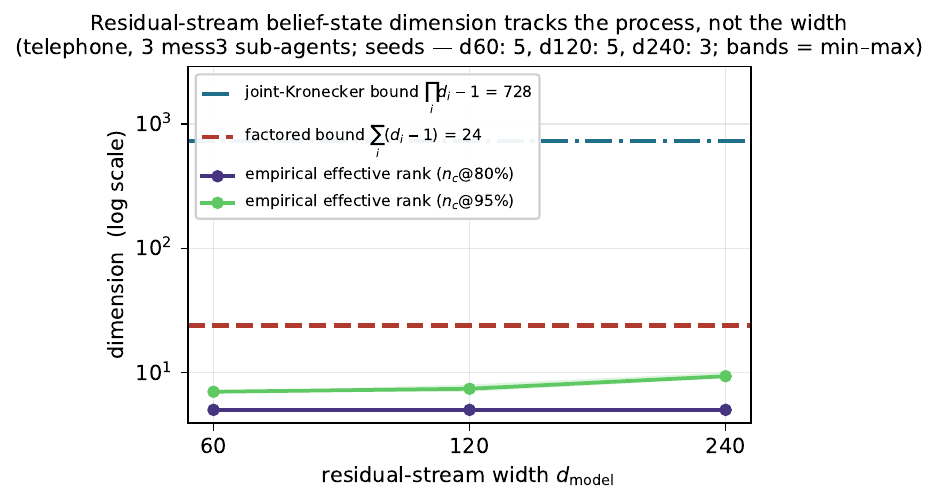}
\caption{\textbf{Effective rank is mostly independent of model width.}
Effective rank of the final-layer residual stream at step $50000$ (BOS
position excluded, $1024$ sequences, same protocol as
\cref{fig:telephone_cev}c) for the three-agent mess3 telephone process,
against residual-stream width $d_\mathrm{model} \in \{60, 120, 240\}$:
$n_c$ at the $80\%$ and $95\%$ CEV thresholds.
Points are means over five seeds for widths 60 and 120, and three seeds for width 240. Bands are min--max across seeds and are degenerate for
$n_c@80\%$. Horizontal lines: the factored
bound $\sum_i (d_i - 1) = 24$ (dashed) and the joint-Kronecker bound
$\prod_i d_i - 1 = 728$ (dash-dotted).}
\label{fig:apx_width_sweep}
\end{figure}

To ensure our results are robust to changes in the transformer architecture, we ran additional variations in which we changed the residual stream width. Sweeping the residual stream width from $60$ to $240$ while keeping the process fixed, we find that the effective rank of the residual stream is largely independent of the architecture (see \cref{fig:apx_width_sweep}). The number of components at the $80\%$ and $95\%$ CEV thresholds remains consistent across widths, indicating that the effective rank is determined by the process rather than the model's architecture. This is further supported by the fact that all runs at different widths yield an $n_c@80\%$ of $5$, which is below the factored bound of $24$ and well below the joint-Kronecker bound of $728$.

\section{Simulated Processes and Parameters}
\label{apx:process_parameters}

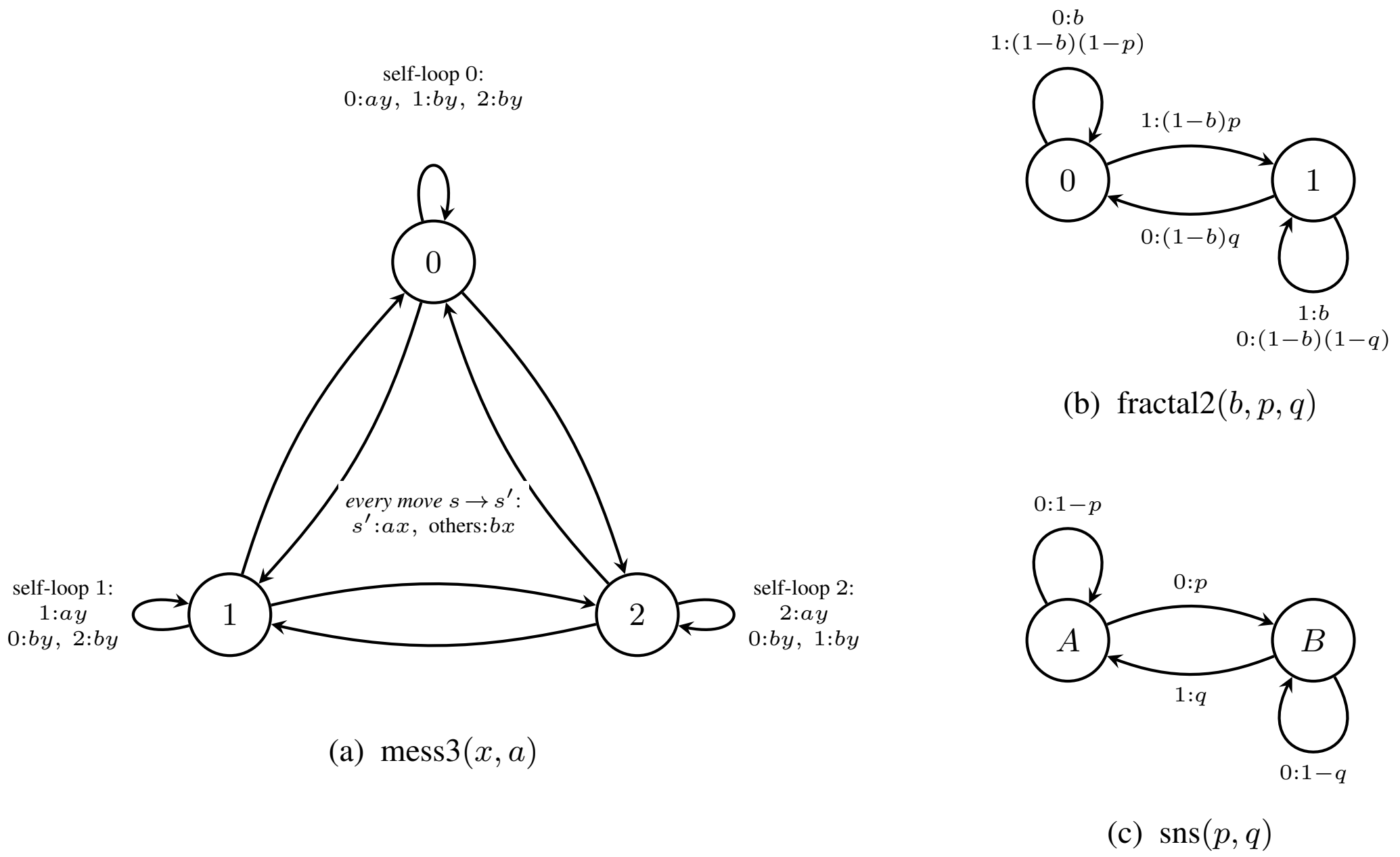
\begin{figure}
\centering
\begin{tikzpicture}[>=stealth, thick, baseline=(current bounding box.center),
  st/.style={draw, circle, minimum size=0.8cm, inner sep=1pt, font=\small},
  lab/.style={font=\tiny, align=center},
  ttl/.style={font=\small},
  pl/.style={font=\normalsize\bfseries}]
  \def\R{2.3}
  \node[st] (m0) at (90:\R)  {$0$};
  \node[st] (m1) at (210:\R) {$1$};
  \node[st] (m2) at (330:\R) {$2$};
  \foreach \a/\b in {m0/m1, m1/m0, m1/m2, m2/m1, m2/m0, m0/m2}{
    \draw[->] (\a) to[bend left=13] (\b);
  }
  \draw[->] (m0) edge[loop above, min distance=7mm, looseness=9] (m0);
  \draw[->] (m1) edge[loop left,  min distance=7mm, looseness=9] (m1);
  \draw[->] (m2) edge[loop right, min distance=7mm, looseness=9] (m2);
  \node[lab] at (90:\R+1.7) {self-loop $0$:\\ $0{:}ay,\ 1{:}by,\ 2{:}by$};
  \node[lab, anchor=east]  at ($(m1)+(-0.95,0)$) {self-loop $1$:\\ $1{:}ay$\\ $0{:}by,\ 2{:}by$};
  \node[lab, anchor=west]  at ($(m2)+( 0.95,0)$) {self-loop $2$:\\ $2{:}ay$\\ $0{:}by,\ 1{:}by$};
  \node[lab, fill=white, inner sep=2pt] at (0,-0.15)
    {\emph{every move} $s\!\to\!s'$:\\ $s'{:}ax,\ \text{others}{:}bx$};
  \node[ttl] at (0,-\R-0.2) {(a)\ \ mess3$(x,a)$};
  \begin{scope}[xshift=7.4cm, yshift=3.1cm]
  \node[st] (f0) at (-1.2,0) {$0$};
  \node[st] (f1) at (1.2,0)  {$1$};
  \draw[->] (f0) to[bend left=22] node[lab, above]{$1{:}(1{-}b)p$} (f1);
  \draw[->] (f1) to[bend left=22] node[lab, below]{$0{:}(1{-}b)q$} (f0);
  \draw[->] (f0) to[out=120, in=60, looseness=7] node[lab, above]{$0{:}b$\\ $1{:}(1{-}b)(1{-}p)$} (f0);
  \draw[->] (f1) to[out=-60, in=-120, looseness=7] node[lab, below]{$1{:}b$\\ $0{:}(1{-}b)(1{-}q)$} (f1);
  \node[ttl] at (0,-2.2) {(b)\ \ fractal2$(b,p,q)$};
  \end{scope}
  \begin{scope}[xshift=7.4cm, yshift=-1.4cm]
  \node[st] (s0) at (-1.2,0) {$A$};
  \node[st] (s1) at (1.2,0)  {$B$};
  \draw[->] (s0) to[bend left=22] node[lab, above]{$0{:}p$} (s1);
  \draw[->] (s1) to[bend left=22] node[lab, below]{$1{:}q$} (s0);
  \draw[->] (s0) to[out=120, in=60, looseness=7] node[lab, above]{$0{:}1{-}p$} (s0);
  \draw[->] (s1) to[out=-60, in=-120, looseness=7] node[lab, below]{$0{:}1{-}q$} (s1);
  \node[ttl] at (0,-1.9) {(c)\ \ sns$(p,q)$};
  \end{scope}
\end{tikzpicture}
\caption{\textbf{The base HMMs} (edge labels read ``emission\,{:}\,probability''). \textbf{(a)} mess3$(x,a)$: three states, three symbols; with $b = (1-a)/2$ and $y = 1-2x$. A transition into state $s'$ emits symbol $s'$ with probability $ax \equiv a x$ and each other symbol with probability $bx$ (independent of the source state, so both incoming arrows share the label shown by each vertex); a self-loop at $s$ emits $s$ with probability $ay$ and each other symbol with probability $by$. Its mixed-state presentation is the fractal simplex of \cref{fig:feedback_msp} (top row). \textbf{(b)} fractal2$(b,p,q)$: two states, two symbols; state $s$ re-emits its own symbol with probability $b$ and stays, otherwise emits the other symbol and moves with probability $p$ (resp.\ $q$). Its two-state belief simplex is the barcode of \cref{fig:apx_role_swap}c. \textbf{(c)} sns$(p,q)$: the simple nonunifilar source; symbol $1$ is emitted only on the resetting $B \to A$ transition, giving a Cantor-like 1-D belief geometry, drawn as a barcode like the fractal2 factor of \cref{fig:apx_role_swap}c.}
\label{fig:apx_hmms}
\end{figure}

The transducers in our simulations are built from three base HMMs, mess3, fractal2 and sns (see \Cref{fig:apx_hmms}), with the input tokens selecting one of multiple parametrizations of the respective HMM (see \cref{tab:process_params}).

\begin{table}
\caption{\textbf{Generative-process parameters.} Sequences have length 8 plus a BOS token. Each agent's kernels are stacked along its input axis: the incoming symbol selects the kernel. Agent $U$ is the action-delayed factor, whose state is augmented by the latched action ($\times A$). Setup A is the mess3--fractal2 loop shown in \cref{fig:feedback_msp}, and the pair (A, B) is shown in \cref{fig:apx_role_swap}. The primed setups (A$'$, B$'$) replace fractal2 by sns; they are the mess3--sns pair of \cref{apx:intervention}. In both pairs, setup A has mess3 as agent $T$. The mess3$\leftrightarrow$mess3 loop appears in \cref{fig:apx_orth_comparison}. The fractal2 kernels use $p = q$: fractal2's state-transition matrix is doubly stochastic exactly then, so the uniform initial belief used by the belief machinery is the true stationary distribution.}
\label{tab:process_params}
\begin{ruledtabular}
\footnotesize
\setlength{\tabcolsep}{3.5pt}
\begin{tabular}{lll}
\toprule
setup & factor & kernels (parameters) \\
\colrule
feedback A & agent $T$ $=$ mess3 ($S{=}3$) & $(x,a) \in \{(0.05, 0.85), (0.15, 0.60)\}$ \\
 (vocab $6$) & agent $U$ $=$ fractal2 ($S{=}2$, $A{=}2$) & $(b; p{=}q) \in \{(0.25; 0.25), (0.50; 0.35),$ \\
 & & $(0.75; 0.65)\}$ \\
\colrule
feedback B & agent $T$ $=$ fractal2 ($S{=}2$) & $(b; p{=}q) \in \{(0.25; 0.25), (0.50; 0.35),$ \\
 (vocab $2{\cdot}3{=}6$) & & $(0.75; 0.65)\}$ \\
 & agent $U$ $=$ mess3 ($S{=}3$, $\times A{=}3$) & $(x,a) \in \{(0.05, 0.85), (0.15, 0.60)\}$ \\
\colrule
feedback A$'$ & agent $T$ $=$ mess3 ($S{=}3$) & $(x,a) \in \{(0.05, 0.85), (0.15, 0.60)\}$ \\
 (vocab $6$) & agent $U$ $=$ sns ($S{=}2$, $\times A{=}2$) & $(p,q) \in \{(0.30, 0.50), (0.50, 0.50), (0.70, 0.50)\}$ \\
\colrule
feedback B$'$ & agent $T$ $=$ sns ($S{=}2$) & $(p,q) \in \{(0.30, 0.50), (0.50, 0.50), (0.70, 0.50)\}$ \\
 (vocab $2{\cdot}3{=}6$) & agent $U$ $=$ mess3 ($S{=}3$, $\times A{=}3$) & $(x,a) \in \{(0.05, 0.85), (0.15, 0.60)\}$ \\
\colrule
loop mess3$\leftrightarrow$mess3 & agent $T$ $=$ mess3 ($S{=}3$) & $(x,a) \in \{(0.05, 0.60), (0.10, 0.50), (0.15, 0.70)\}$ \\
 (vocab $3{\cdot}3{=}9$) & agent $U$ $=$ mess3 ($S{=}3$, $\times A{=}3$) & $(x,a) \in \{(0.05, 0.60), (0.10, 0.50), (0.15, 0.70)\}$ \\
\colrule
telephone, fractal2 & agents $1$--$5$ & $(b; p{=}q)$ pairs, kernel $0$ / kernel $1$: \\
 ($N{=}5$, $A{=}2$, & & $A_1$: $(0.25; 0.25)$ / $(0.75; 0.75)$ \\
 vocab $4^5{=}1024$) & & $A_2$: $(0.30; 0.30)$ / $(0.70; 0.70)$ \\
 & & $A_3$: $(0.20; 0.30)$ / $(0.80; 0.70)$ \\
 & & $A_4$: $(0.35; 0.325)$ / $(0.65; 0.675)$ \\
 & & $A_5$: $(0.40; 0.30)$ / $(0.60; 0.70)$ \\
\colrule
telephone, mess3 & agent $i$, kernels & $(x,a)$ triples (percept digit $0/1/2$): \\
 ($N \in \{2,\dots,5\}$, & & $A_1$: $(0.05,0.85)\,(0.10,0.50)\,(0.15,0.70)$ \\
 $A{=}3$, & & $A_2$: $(0.05,0.60)\,(0.10,0.50)\,(0.15,0.70)$ \\
 vocab $9^N$) & & $A_3$: $(0.03,0.80)\,(0.12,0.55)\,(0.15,0.65)$ \\
 & & $A_4$: $(0.06,0.75)\,(0.09,0.45)\,(0.14,0.60)$ \\
 & & $A_5$: $(0.04,0.70)\,(0.11,0.65)\,(0.13,0.55)$ \\
\bottomrule
\end{tabular}
\end{ruledtabular}
\end{table}

\begin{table}
\caption{\textbf{Transformer and optimization hyperparameters}, shared by all runs (exceptions noted).}
\label{tab:transformer_params}
\begin{ruledtabular}
\footnotesize
\setlength{\tabcolsep}{3.5pt}
\begin{tabular}{ll}
\toprule
parameter & value \\
\colrule
layers / $d_\mathrm{model}$ / heads / $d_\mathrm{head}$ / $d_\mathrm{mlp}$ & $4$ / $120$ / $3$ / $40$ / $480$ \\
activation / normalization / positions & ReLU / LayerNorm / learned absolute \\
context $n_\mathrm{ctx}$ & $9$ (BOS $+$ 8 tokens) \\
vocabulary & joint process vocab $+$ 1 (BOS) \\
optimizer & AdamW, lr $5{\times}10^{-4}$, $\beta = (0.9, 0.999)$, \\
 & weight decay $0$ \\
LR schedule & windowed reduce-on-plateau (factor $0.7$, \\
 & patience $10^4$ steps, cooldown $2{\times}10^4$ steps, min lr $10^{-7}$) \\
batch size & $25000$ sequences, sampled fresh each step (online); \\
 & $8192$ for telephone mess3 $N{=}5$ (memory) \\
training steps & $50000$ \\
seeds (process / weight init) & canonical runs: $77$ / $7$ (canonical A: $1102$ / $1152$, \\
 & B: $1101$ / $1151$); four additional seeds per setup: \\
 & feedback A $1301$--$1304$, feedback B $1001$--$1004$, \\
 & feedback A$'$ $201$--$204$, feedback B$'$ $101$--$104$, \\
 & loop mess3$\leftrightarrow$mess3 $901$--$904$, \\
 & telephone fractal2 $801$--$804$, telephone mess3 \\
 & $N{=}2$: $501$--$504$, $N{=}3$: $401$--$404$, $N{=}4$: $601$--$604$, \\
 & $N{=}5$: $701$--$704$ (weight-init seed $=$ process seed $+\,50$; \\
 & data streams of the reruns and of canonical A/B are \\
 & seeded per run, the seed-$77$ canonical runs share the \\
 & legacy step-indexed stream) \\
width sweep (\cref{apx:width_sweep}) & $d_\mathrm{model} \in \{60, 240\}$ with $3$ heads, $d_\mathrm{head} = d_\mathrm{model}/3$, \\
 & $d_\mathrm{mlp} = 4\,d_\mathrm{model}$; seeds $2001$--$2005$ resp.\ $2101$--$2103$ \\
 & ($d_\mathrm{model} = 120$ is the canonical telephone mess3 family) \\
evaluation & every $10^3$ steps on fresh sequences; exact Bayes floor \\
 & from $32768$ MC samples \\
checkpoints & log-spaced schedule ($\sim$78 per run) \\
\bottomrule
\end{tabular}
\end{ruledtabular}
\end{table}

If not specified otherwise, the same four-layer transformer with a residual stream width of $120$ and three heads with a head dimension of $40$ has been used (see \cref{tab:transformer_params}).

\begin{figure}
\centering
\includegraphics[width=\textwidth]{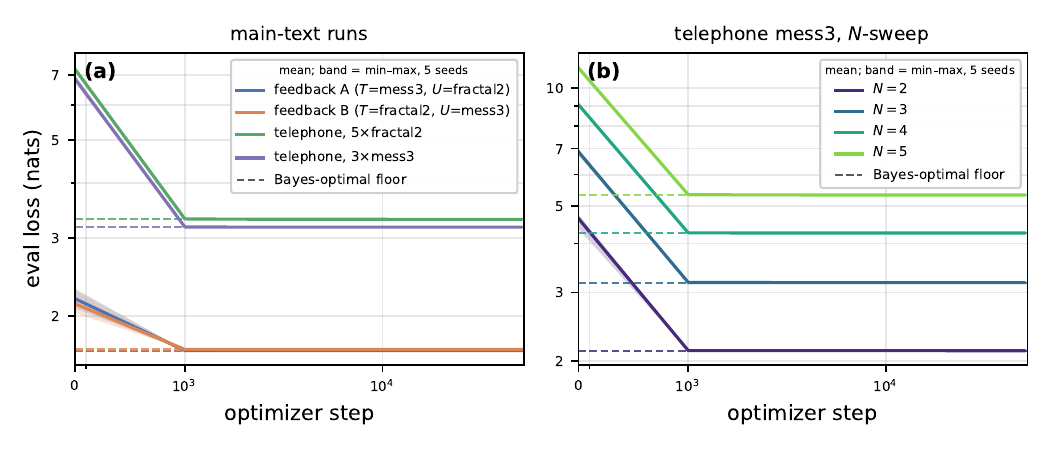}
\caption{\textbf{All runs converge to their Bayes-optimal floors.} Evaluation loss over training (log scale); solid curves are means over five seeds per setup, shaded bands the min--max across seeds (mostly narrower than the line). Dashed lines: each setup's Bayes-optimal floor, position-averaged and estimated with $32768$ MC samples (per-seed floor estimates differ by ${\sim}10^{-3}$ nats; the line shows the family mean). (a) The four main-text setups: the mess3--fractal2 feedback loops A and B of \cref{fig:feedback_msp,fig:apx_role_swap} and the two telephone processes. (b) The telephone-mess3 $N$-sweep of \cref{fig:telephone_cev}c. Every setup reaches its floor within ${\sim}10^{3}$ steps; final losses lie within $1$--$5 \times 10^{-3}$ nats of the floor, i.e.\ at the precision of the MC estimate for all runs except two of the five $N{=}4$ seeds, which plateau $3$--$5 \times 10^{-3}$ nats above it. The $x$-axis is symlog (linear below $10^3$ to show the step-0 evaluation, logarithmic above).}
\label{fig:apx_convergence}
\end{figure}

The models are trained online for $50,000$ training steps and converge close to the Bayes-optimal floor: across all five seeds per setup, runs converge to within a few $10^{-3}$ nats of their floors in all cases except two of the five $N{=}4$ seeds, which settle $3$--$5\times 10^{-3}$ nats above it (see \cref{fig:apx_convergence}).

For each figure, a fixed batch of $1024$ sequences is sampled from the generative process together with the Bayes-optimal per-factor belief states (the MSP coordinates) along each sequence. \emph{MSP panels}: the ground-truth belief states of a factor are projected onto their own top-two PCA directions; the model reconstruction is an ordinary-least-squares linear probe from the final-layer residual stream (\texttt{resid\_post}, BOS position excluded) to the belief coordinates, fit on $80\%$ of sequences, evaluated on the held-out $20\%$ (reported $R^2$), and projected onto the \emph{same} PCA basis as the ground truth. For two-state factors the belief simplex is one-dimensional and PCA direction two is degenerate; those panels are drawn as barcode plots, with one vertical line per sample at its belief coordinate. \emph{CEV panels}: cumulative explained variance by rank of the centered final-layer residual stream (SVD), per checkpoint, compared against the CEV of the concatenated per-factor beliefs (factored reference) and of their Kronecker product (joint reference) on the same batch.

The code to reproduce all figures is based on the implementation used by \citet{shai2026transformers} and is \ifarxiv publicly available at \coderepo\fi.

\section{Interventional Test of Belief Factorization}
\label{apx:intervention}
\label{apx:steering}

Linear decodability of per-factor beliefs (\cref{fig:feedback_msp}) is correlational. Here we test the \emph{causal} counterpart: does the trained network actually use the observed factored representations to predict the next token? To test this question, we patch counterfactual beliefs into a single factor's residual-stream subspace and check whether the output distribution shifts exactly as the Bayes-optimal predictor would if only that factor's belief had changed.

For a fixed batch of $1024$ sequences with exact per-factor Bayes-optimal belief states $b_f$, we collect the final-layer residual stream $h$ (\texttt{resid\_post}, before the final LayerNorm) at all non-BOS positions and fit, by ordinary least squares, a single \emph{joint encoder}
\begin{align}
h \;\approx\; c + \textstyle\sum_f E_f\, b_f,
\end{align}
whose blocks $E_f$ serve as per-factor \emph{write} matrices, together with per-factor \emph{read} probes $b_f \approx W_f^{\top} h + a_f$. (For the telephone process, the trivial hub factor is omitted.) Fitting this joint encoder instead of inverting the belief probes of the correlational results has been crucial for successful patching, as probes appear to sometimes fit smaller variance directions, which mirror the belief state but have little causal power.

\begin{figure}
\centering
\includegraphics[width=\textwidth]{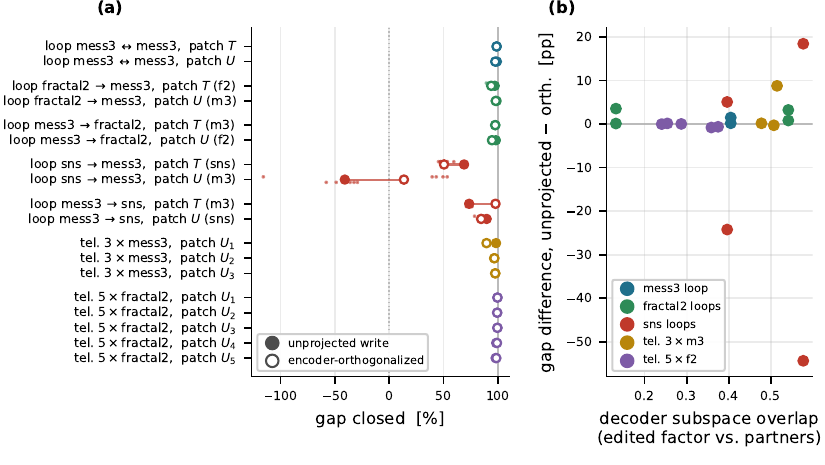}
\caption{\textbf{Belief patching under the unprojected and the encoder-orthogonalized write.} \textbf{(a)} Fraction of the null-to-target distance closed by the edit, per setup and patched factor (large markers: ratio of means over seeds; small dots: individual seeds; filled: unprojected write $\Delta$, open: $\Delta_{\perp}$ of \cref{eq:steering_write}). Every setup without an sns factor closes $89.5$--$99.6\%$ of the gap under either construction. The two sns loops fail under both, and are the only setups the construction changes qualitatively: under the unprojected write, editing the mess3 partner in the sns$\,\to\,$mess3 loop leaves the output \emph{farther} from the target than no edit at all ($-41\%$), which the projection turns into $+14\%$---the raw write was corrupting the sns encoding directly. \textbf{(b)} The between-factor encoder subspace overlap (share of the edited factor's encoder span lying in its partners') against the gap difference between the constructions. The overlap is substantial everywhere ($0.13$--$0.58$) and does not separate fragile from robust writes---the sns loops are the outliers in both directions---so the deviations are not explained by overlapping belief encodings.}
\label{fig:apx_orth_comparison}
\end{figure}

Counterfactual beliefs come from a random donor permutation $\pi$ of the batch, $b_f' = b_{f,\pi(i)}$, taken at the same sequence position (the spliced tuple $(b_f', b_{-f})$ is in general \emph{not} jointly realizable). At the last trained position we add
\begin{align}
\Delta \;=\; E_f\,(b_f' - b_f), \qquad
\Delta_{\perp} \;=\; \Delta - \big(\Delta\, E_{-f}\big)\, E_{-f}^{+},
\label{eq:steering_write}
\end{align}
where $E_{-f}$ stacks the write matrices of all other factors and $(\cdot)^+$ is the pseudoinverse: the edit is orthogonalized against the subspace in which the residual stream \emph{encodes} the other factors' beliefs, so the write cannot directly alter any partner's belief encoding. Projecting out the partners' read probes $W_{-f}$ instead would be nearly vacuous: the raw write already lies almost entirely in the probes' null space (the projection removes $0.1$--$0.3\%$ of its norm and changes no reported number relative to using no projection at all), because a factor's read-probe and encoder subspaces are themselves nearly orthogonal (mean overlap $0.05$ across setups). The encoder projection removes $23$--$66\%$ of the write. Neither construction is uniformly conservative: the encoder projection also removes whatever part of the \emph{intended} write falls in the overlap of the encoder spans, deflating steering rates where that overlap is large, while the unprojected write can corrupt the partners' encodings directly, deflating shielding for reasons that are artifacts of the write rather than properties of the network. The main text reports the unprojected write $\Delta$; \cref{fig:apx_orth_comparison} compares both constructions across every setup, together with the between-factor encoder subspace overlap that determines how much the projection removes. In every setup that steers, the two constructions agree to within a few percentage points. The model is then run forward from the edited activation, and its next-token distribution (renormalized over the process vocabulary, excluding BOS) is compared against \emph{exact} Bayes references computed from the generative process by evaluating the observation distribution on spliced belief tuples.

The edited output is compared, in KL divergence on the joint next-token distribution, against \emph{exact} Bayes references evaluated on patched belief tuples: $H_0 = P(\cdot \,|\, b_T, b_U)$ (edit inert), $H_1 = P(\cdot \,|\, b_f', b_{-f})$ (the factorized single-factor prediction---the patching target), and $H_2 = P(\cdot \,|\, b_T', b_U')$ (a wholly unrelated sequence). The headline quantity is the \emph{closed gap} $1 - \mathrm{KL}(p_{\mathrm{edited}} \,\|\, H_1) / \mathrm{KL}(p_{\mathrm{clean}} \,\|\, H_1)$---the fraction of the distance between an inert write and the patching target that the edit actually covers---reported per patched factor as a mean over evaluable interventions. Interventions whose references are closer than $0.05$ total variation are excluded as unidentifiable ($\sim\!70$--$78\%$ of the batch remains).

\Cref{fig:apx_orth_comparison}a summarizes belief patching across every setup. All setups without an sns factor close $96.5$--$99.6\%$ of the null-to-target gap under the unprojected write ($89.5$--$99.6\%$ under the orthogonalized one; the largest single change is the first sub-agent of the mess3 telephone, $98.3\% \to 89.5\%$, which is also where the projection removes the most write mass, $52\%$). The two sns-containing loops, however, deviate from this finding: the closed gaps drop to $74\%$/$90\%$ (setup A$'$, patching mess3 resp.\ sns) and $69\%$/$-41\%$ (setup B$'$, patching sns resp.\ its mess3 partner), while no other setup falls below $96\%$. Why this failure occurs remains a subject of future research.

\begin{figure}
\centering
\includegraphics[width=\textwidth]{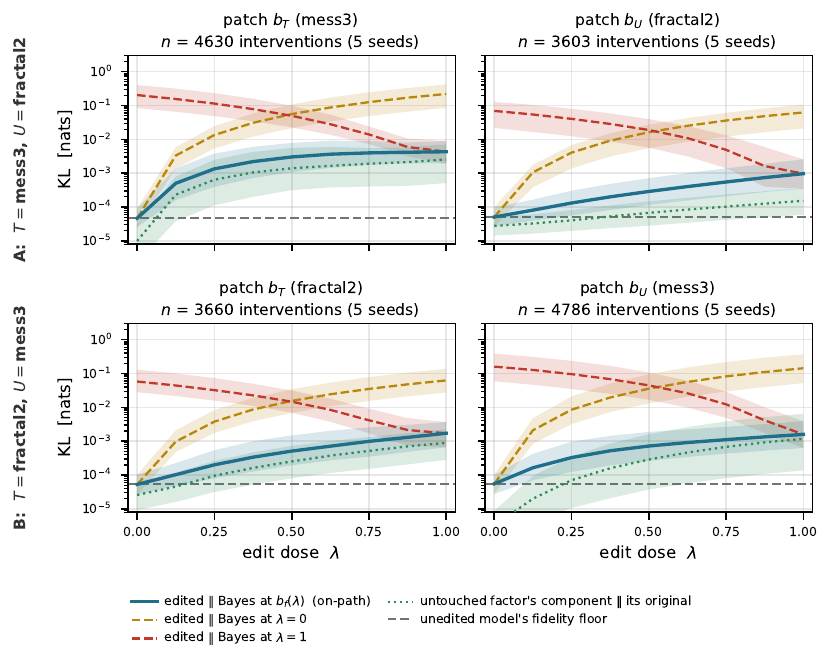}
\caption{\textbf{Belief edits track the Bayes prediction continuously along
the interpolation path.} KL divergence (median over interventions; bands:
interquartile range) between the edited next-token distribution and three
references, as a function of the edit dose $\lambda$: the Bayes prediction
at the interpolated tuple $(b_f(\lambda), b_{-f})$ (on-path, solid), and
the Bayes predictions at the two endpoints ($\lambda = 0$: original;
$\lambda = 1$: full counterfactual; dashed). Rows: the two feedback setups;
columns: the patched factor; $n$: evaluable interventions pooled over five
seeds per setup. The dashed horizontal rule is the unedited model's own
fidelity floor; the dotted curve is the untouched factor's component
against its original prediction (leakage). At $\lambda = 0.5$ the nearest
endpoint reference is $16$--$56\times$ farther from the edited output than
the interpolated target. Panel counts differ only through the
identifiability gate ($0.05$ total variation between the endpoint
references): the candidate pool is identical across panels, but the
fractal2 belief distribution concentrates on a near-degenerate
one-dimensional set, so swapping a fractal2 belief more often produces
indistinguishable references.}
\label{fig:apx_dose_response}
\end{figure}

\paragraph{Dose--response} The endpoint edit ($b_f \to b_f'$) leaves open whether the write works only between realizable belief states. We therefore
interpolate the edit, splicing $b_f(\lambda) = (1-\lambda)\,b_f + \lambda\, b_f'$ for $\lambda \in \{0, 0.125, \dots, 1\}$---a valid simplex point,
but in general \emph{not} a reachable mixed state of the process---and compare the edited output against the Bayes prediction at the interpolated tuple $(b_f(\lambda), b_{-f})$. \Cref{fig:apx_dose_response} shows the result for the two mess3--fractal2 loops (five seeds each, unprojected
write): at intermediate dose, the edited output lies one to two orders of magnitude closer to the interpolated target than
to either endpoint reference---the endpoint references cross near
$\lambda = 0.5$ while the on-path curve passes beneath both---and the
edit closes $81$--$99\%$ of the inert-write-to-target gap at every
$\lambda$. These results suggest that the transformer truly learns the functional (linear) relationship between the belief state and the next-token probabilities, instead of memorizing.

\section{Complexification Through Coarse-Graining: Construction and Dimensions}
\label{apx:coarse_graining}

This appendix constructs an explicit two-agent feedback loop for which
coarse-graining the observed interface by means of hiding either agent's emissions produces a measurable and predictable
increase in the belief-state dimension an optimal predictor must represent.
The increase in dimension happens because while the fully observed feedback loop factorizes, the coarse-grained process cannot make use of the same compression, as it does not fully observe all exchanged tokens between the two factors.

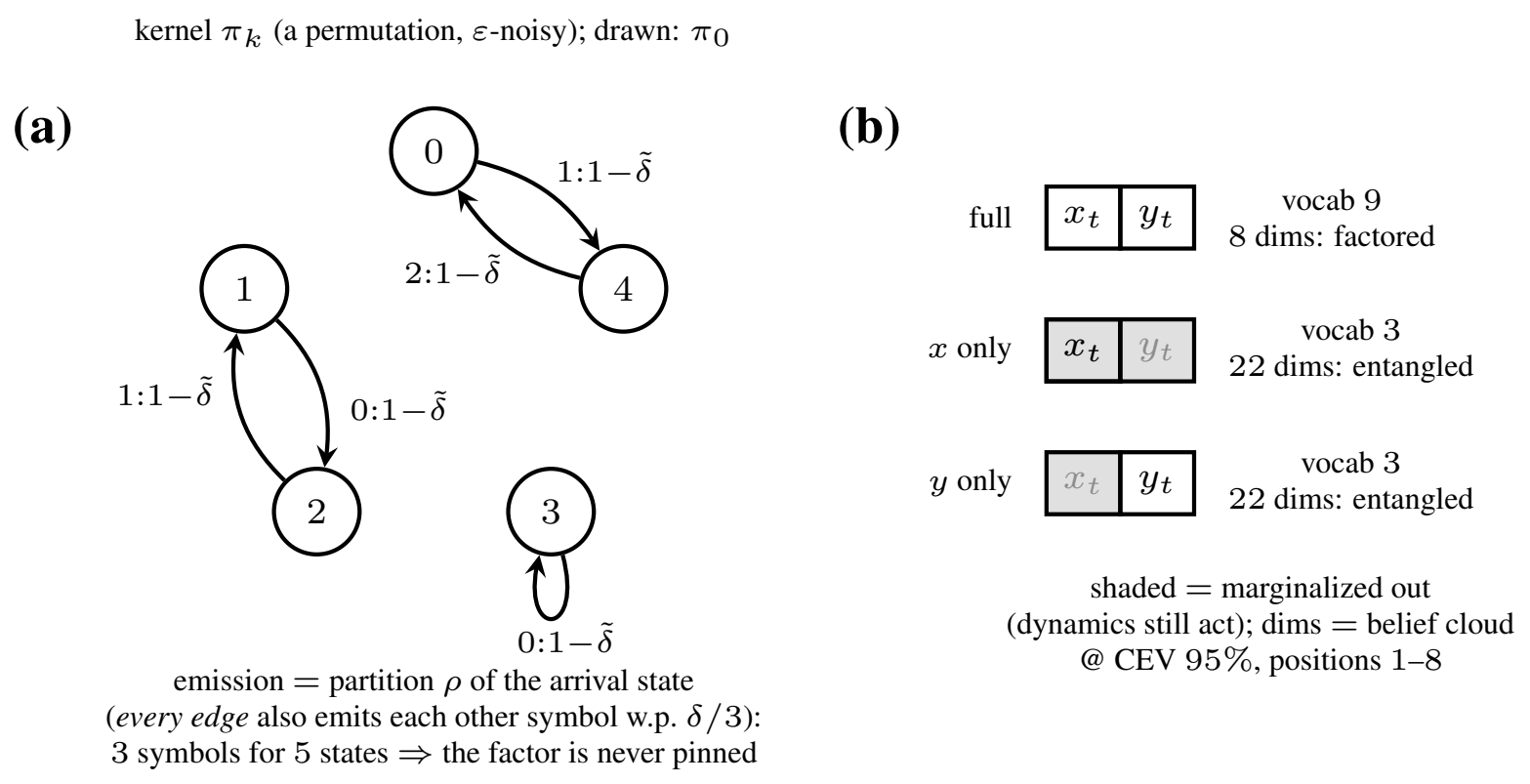
\begin{figure}
\centering
\begin{tikzpicture}[>=stealth, thick,
  st/.style={draw, circle, minimum size=0.58cm, inner sep=0pt, font=\scriptsize},
  cell/.style={draw, minimum width=0.5cm, minimum height=0.42cm,
               inner sep=0pt, font=\scriptsize},
  hid/.style={cell, fill=black!12, text=black!45},
  lab/.style={font=\tiny, align=center},
  pl/.style={font=\normalsize\bfseries}]

\begin{scope}[shift={(1.9,0.25)}]
\def\Rc{1.35}
\node[pl] at (-2.65, 1.5) {(a)};
\node[st] (c0) at (90:\Rc)  {$0$};
\node[st] (c1) at (162:\Rc) {$1$};
\node[st] (c2) at (234:\Rc) {$2$};
\node[st] (c3) at (306:\Rc) {$3$};
\node[st] (c4) at (18:\Rc)  {$4$};
\draw[->, bend left=22] (c0) to node[lab, above right, inner sep=1pt]
  {$1{:}1{-}\tilde{\delta}$} (c4);
\draw[->, bend left=22] (c4) to node[lab, below left, inner sep=1pt]
  {$2{:}1{-}\tilde{\delta}$} (c0);
\draw[->, bend left=28] (c1) to (c2);
\draw[->, bend left=28] (c2) to (c1);
\node[lab, anchor=east] at ($($(c1)!0.5!(c2)$)+(-0.32,0.05)$)
  {$1{:}1{-}\tilde{\delta}$};
\node[lab, anchor=west] at ($($(c1)!0.5!(c2)$)+(0.34,-0.05)$)
  {$0{:}1{-}\tilde{\delta}$};
\draw[->] (c3) edge[loop below, min distance=6mm, looseness=8] (c3);
\node[lab, anchor=north] at ($(c3)+(0.1,-0.62)$) {$0{:}1{-}\tilde{\delta}$};
\node[lab] at (0, 2.15) {kernel $\pi_k$ (a permutation, $\varepsilon$-noisy);
  drawn: $\pi_0$};
\node[lab] at (0, -2.5) {emission $=$ partition $\rho$ of the arrival state\\
  (\emph{every edge} also emits each other symbol w.p.\ $\delta/3$):\\
  $3$ symbols for $5$ states $\Rightarrow$ the factor is never pinned};
\end{scope}

\begin{scope}[xshift=5.4cm]
\node[pl] at (-0.55, 1.75) {(b)};
\node[cell] (fx) at (0.9, 1.15) {$x_t$};
\node[cell] (fy) at (1.4, 1.15) {$y_t$};
\node[lab, anchor=east] at (0.55, 1.15) {full};
\node[lab, anchor=west] at (1.75, 1.15) {vocab $9$\\ $8$ dims: factored};

\node[hid] (xx) at (0.9, 0.25) {$x_t$};
\node[hid] (xy) at (1.4, 0.25) {$y_t$};
\node[cell] at (0.9, 0.25) {$x_t$};
\node[lab, anchor=east] at (0.55, 0.25) {$x$ only};
\node[lab, anchor=west] at (1.75, 0.25) {vocab $3$\\ $22$ dims: entangled};

\node[hid] (yx) at (0.9, -0.65) {$x_t$};
\node[cell] at (1.4, -0.65) {$y_t$};
\node[lab, anchor=east] at (0.55, -0.65) {$y$ only};
\node[lab, anchor=west] at (1.75, -0.65) {vocab $3$\\ $22$ dims: entangled};

\node[lab, anchor=north] at (2.1, -1.15) {shaded $=$ marginalized out\\ (dynamics still act); dims $=$
  belief cloud\\ @ CEV $95\%$, positions $1$--$8$};
\end{scope}
\end{tikzpicture}
\caption{\textbf{The coarse-graining demonstrator process:} two five-state
transducers of this family in the canonical feedback loop, each one's emitted
symbol the other's next input symbol $k$ (agent action-delayed).
\textbf{(a)} Structure of one factor (edge labels read
``emission\,{:}\,probability'' as in \cref{fig:apx_hmms}, with
$\tilde{\delta} \equiv 2\delta/3$). Each input symbol $k$ selects a
near-deterministic permutation
kernel of the five states (drawn: $\pi_0 = (4,2,1,3,0)$, two transpositions
and a fixed point), so different hidden-action histories imply visibly
different dynamics; each transition emits only a three-way \emph{partition}
$\rho$ of its arrival state (edges into the same state carry the same
dominant token), so an observer never pins the state exactly and
cannot invert the partner's symbol. \textbf{(b)} The three observed
interfaces. Observing the pair $(x_t, y_t)$ makes the joint belief state factorize
into two per-factor belief states ($8$ dimensions); marginalizing either symbol
destroys the factorization and forces an entangled belief state over the $75$-state
joint latent ($22$ dimensions), even though the coarse observer receives
strictly less information.}
\label{fig:apx_cg_process}
\end{figure}

To demonstrate this complexification, we analyzed a feedback loop between two transducers implementing a noisy, input-dependent permutation of their five internal states, with a noisy mapping to three emission tokens for each transition. \cref{fig:apx_cg_process} illustrates the
construction and the three observed interfaces. For a factor with states
$\mathcal{S} = \{0,\dots,4\}$, input symbol $k \in \{0,1,2\}$ (the partner's
current symbol) and output alphabet $\{0,1,2\}$:
\begin{align}
T^{(k)} &= (1-\varepsilon)\, P_{\pi_k} + \tfrac{\varepsilon}{5}\,\mathbf{1}\mathbf{1}^{\!\top},
&
\Pr(v \,|\, s') &= (1-\delta)\,[\rho(s') = v] + \tfrac{\delta}{3},
\end{align}
where $P_{\pi_k}$ is the permutation matrix of $\pi_k$, the emission is read
from the \emph{arrival} state $s'$, and $\rho: \mathcal{S} \to \{0,1,2\}$ is a
fixed 5-to-3 partition map. The permutations $\pi_k$, the partition maps
$\rho$ and the noise levels ($\varepsilon \in [0.02, 0.1]$,
$\delta \in [0, 0.15]$) of both agents are drawn independently for each
replicate of \cref{apx:cg_trained}; \cref{fig:apx_cg_process}a shows one
kernel of the draw analyzed in \cref{apx:cg_dims}.
The joint token is the pair $(x, y)$ of both emissions (vocabulary $9$); the
two coarse-grainings observe only $x$ or only $y$ (vocabulary $3$),
marginalizing the other symbol while its dynamics continue to act.

\subsection{Theoretical and Measured Dimensions}
\label{apx:cg_dims}

The dimensions of the factored simplices can be computed directly: agent $T$ has a dimension of $d_1 = 5$, while the action-delayed agent $U$
has dimension $d_2 = 5 \cdot 3 = 15$ (mind $\times$ latched action). In contrast, the Kronecker product of the two simplices lives on a $74$-dimensional simplex ($75$ states), revealing a large dimensional compression through factorization in this process:
\begin{align}
\underbrace{\textstyle\sum_f (d_f - 1) = 18}_{\text{factored budget}}
\;\ll\;
\underbrace{\textstyle\prod_f d_f - 1 = 74}_{\text{joint simplex (hull bound)}},
\qquad
\text{full-obs.\ product cloud} \;\le\; 74 .
\end{align}
Exact filtering on the $75$-state joint machine (labeled transition operators
extracted from the factored implementation by delta-probing; $1536$
sequences) gives, within the model context window (positions $1$--$8$),
cumulative-explained-variance dimensions:

\begin{center}
\footnotesize
\begin{tabular}{lccc}
observation & representation & dims @$80\%$ & dims @$95\%$ \\
\hline
full $(x,y)$ & factored concatenation $(b^{(1)}, b^{(2)})$ & $7$ & $8$ \\
full $(x,y)$ & joint product cloud (containment bound) & $19$ & $23$ \\
only $x$ (agent hidden) & entangled joint belief state & $16$ & $22$ \\
only $y$ (env.\ hidden) & entangled joint belief state & $17$ & $22$ \\
\end{tabular}
\end{center}

Both coarse-grainings force the observer from the $8$-dimensional factored
representation onto a $22$-dimensional entangled cloud---a
$2.75\times$ increase at the $95\%$ threshold, obtained by \emph{removing}
observation symbols---while respecting containment in the fully-observed product cloud
($22 \le 23$),
confirming that the mechanism is the loss of factored compression rather
than the creation of new belief geometry. (Pooled over longer horizons the
coarse clouds saturate the hull, dims $19/23$ at $80/95\%$.) The prediction
for trained networks is the corresponding effective-rank inversion between
models trained on the full ($9$-token) and coarse ($3$-token) streams of the
same process.

\subsection{Trained Networks: The Effective-Rank Inversion}
\label{apx:cg_trained}

We trained transformers, identical in architecture and optimization to
all other runs (\cref{tab:transformer_params}), on the three streams of
five independent draws of this process family (replicates r0--r4; r0 is
the draw analyzed in \cref{apx:cg_dims}, and each draw has its own
permutation kernels, partition maps and noise levels $(\varepsilon, \delta)$
for both agents). For each draw we
trained one model per interface---the full joint token (vocabulary $9$) and
the two coarse-grainings (vocabulary $3$)---with its own training seed,
$15$ models in all. All $15$ converge to their own Bayes-optimal floors
(final loss within $\pm 4.4 \times 10^{-3}$ nats of the floor, i.e.\ at the
Monte-Carlo precision of the floor estimates), so the comparison is between
models that have solved their respective prediction tasks. The exact belief-state
dimensions are essentially draw-independent ($8$ under full observation in
every draw; $20$--$22$ resp.\ $18$--$22$ at $95\%$ CEV under the two
coarse-grainings), whereas the trained models' ranks vary considerably
between draws, so the within-draw comparison of the three interfaces is the
relevant statistic.

\begin{table}
\centering
\caption{\textbf{Effective rank of the final-layer residual stream per replicate}, the smallest number of principal components whose CEV reaches $95\%$ (resp.\ $99\%$), on the three observed interfaces of the coarse-graining demonstrator process ($1024$ sequences, BOS position excluded). Replicates are independent draws of the process family (r0 is the draw analyzed in \cref{apx:cg_dims}). For reference, the exact Bayes-optimal belief states on the same sequences reach $95\%$ at $8$ dimensions under full observation in every draw and at $18$--$22$ dimensions under either coarse-graining.}
\label{tab:cg_replicates}
\footnotesize
\begin{tabular}{lcccccc}
\toprule
 & \multicolumn{2}{c}{full $(x,y)$} & \multicolumn{2}{c}{only $x$ ($U$ hidden)} & \multicolumn{2}{c}{only $y$ ($T$ hidden)} \\
replicate & $95\%$ & $99\%$ & $95\%$ & $99\%$ & $95\%$ & $99\%$ \\
\midrule
r0 & $16$ & $48$ & $32$ & $69$ & $24$ & $56$ \\
r1 & $21$ & $56$ & $42$ & $77$ & $27$ & $57$ \\
r2 & $13$ & $38$ & $27$ & $62$ & $25$ & $55$ \\
r3 & $10$ & $29$ & $12$ & $38$ & $23$ & $54$ \\
r4 & $20$ & $53$ & $27$ & $62$ & $34$ & $67$ \\
\midrule
mean & $16.0$ & $44.8$ & $28.0$ & $61.6$ & $26.6$ & $57.8$ \\
\bottomrule
\end{tabular}
\end{table}

\Cref{fig:coarse_graining}d and \cref{tab:cg_replicates} confirm the predicted
inversion. At the $95\%$ threshold the final-layer residual stream has a mean
effective rank of $16$ (range $10$--$21$) for the fully observing models but
$28$ (range $12$--$42$) and $27$ (range $23$--$34$) for the models observing
only $x$ resp.\ only $y$ (at $99\%$: $45$ against $62$ and $58$). The
absolute ranks vary considerably between draws, but the within-draw
difference does not change sign: in all ten (draw, coarse interface)
pairs the coarse model's rank exceeds that of the fully observing model
of the same draw, by $2$--$21$ components at $95\%$ and
$1$--$25$ at $99\%$. The networks given strictly less information thus
carry the higher-dimensional representation, as they cannot use the factored compression.

Three limitations should be stated. First, the inversion is visible at the
$95\%$ and $99\%$ thresholds but not at $80\%$ (mean ranks $5.8$, $6.4$ and
$6.2$; the within-draw difference changes sign across draws). Second, the
effect is weakest in draw r3, where the $x$-only model exceeds the
full-observation model by only two components at $95\%$. Third, model ranks exceed the
corresponding belief-state ranks everywhere (e.g.\ $16$ vs $8$ under full
observation), so the raw effective rank is a loose upper proxy for the
dimension of the belief representation itself; isolating the
belief-predictive subspace before measuring its rank would sharpen the
comparison.

\end{document}